\pdfoutput=1

\documentclass[11pt]{article}

\usepackage[final]{acl}

\usepackage{times}
\usepackage{latexsym}
\usepackage{enumitem}
\usepackage{amsmath}
\usepackage{amssymb}
\usepackage{graphicx}
\usepackage{amsfonts}
\usepackage{bm}
\usepackage{algorithm}
\usepackage{algpseudocode}
\usepackage{subcaption}
\usepackage{tcolorbox}
\usepackage{xcolor}
\usepackage{hyperref}
\usepackage{inconsolata}
\usepackage[T1]{fontenc}

\usepackage[utf8]{inputenc}

\usepackage{microtype}

\usepackage{inconsolata}

\usepackage{graphicx}
\usepackage{adjustbox}
\usepackage{xcolor}

\definecolor{mplgreen}{HTML}{4DAF4A}
\definecolor{mplblue}{HTML}{377eb8}
\definecolor{mplred}{HTML}{E41A1C}
\definecolor{mplpurple}{HTML}{984EA3}
\definecolor{mplorange}{HTML}{FF7F00}
\definecolor{mplpurplesae}{HTML}{7570b3}

\newcommand\blfootnote[1]{%
  \begingroup
  \renewcommand\thefootnote{}\footnote{#1}%
  \addtocounter{footnote}{-1}%
  \endgroup
}

\newtcolorbox{keytakeaway}[1][]{
  colframe=blue!30!black,
  colback=blue!3!white,
  boxrule=0.4mm,
  arc=1.5mm,
  title=\textbf{Key Takeaway},
  fonttitle=\bfseries\small, 
  toptitle=0.5mm,            
  bottomtitle=0.5mm,         
  fontupper=\small,          
  boxsep=0.5mm,              
  left=1mm, right=1mm,       
  top=0.5mm, bottom=0.5mm,   
  before skip=4pt,           
  after skip=4pt,
  #1
}

\newtcolorbox[auto counter]{takeaway}[1][]{
  colframe=blue!30!black, 
  colback=blue!3!white,   
  boxrule=0.4mm,             
  arc=1.5mm,                 
  title=Takeaway~\thetcbcounter,
  fonttitle=\bfseries\small, 
  toptitle=0.5mm,            
  bottomtitle=0.5mm,         
  fontupper=\small,          
  boxsep=0.5mm,              
  left=1mm,                  
  right=1mm,                 
  top=0.5mm,                 
  bottom=0.5mm,              
  before skip=4pt,           
  after skip=4pt,
  #1
}

\title{Multilingual Language Models Encode Script Over Linguistic Structure}
\title{Multilingual Language Models \\ Encode Script Over Linguistic Structure}

\author{
    Aastha A K Verma$^{\bm{\dagger},1}$~\;~
    Anwoy Chatterjee$^{\bm{\dagger},1}$~\;~ 
    Mehak Gupta${^1}$~\;~
    Tanmoy Chakraborty${^{1,2}}$\\
    ${^1}$Indian Institute of Technology Delhi, New Delhi, India\\ 
    ${^2}$Indian Institute of Technology Delhi, Abu Dhabi, UAE\\
\begin{tabular}{ll}
        \href{mailto:aastha.v1411@gmail.com}{\texttt{aastha.v1411@gmail.com}} & \href{mailto:anwoychatterjee@gmail.com}{\texttt{anwoychatterjee@gmail.com}} \\
        \href{mailto:mehak.gupta.tech@gmail.com}{\texttt{mehak.gupta.tech@gmail.com}} & \href{mailto:tanchak@iitd.ac.in}{\texttt{tanchak@iitd.ac.in}}
    \end{tabular}
}
\usepackage{color, colortbl}
\definecolor{Gray}{gray}{0.9}
\usepackage{multirow,adjustbox}
\usepackage{diagbox}
\newcolumntype{M}[1]{>{\centering\arraybackslash}m{#1}}

\usepackage{amsmath}
\usepackage{amsfonts}
\usepackage{booktabs}

\begin{document}
\maketitle
\begingroup
\renewcommand\thefootnote{{$\bm{\dagger}$}}
\footnotetext{These two authors contributed equally to this work.}
\endgroup
\begin{abstract}
Multilingual language models (LMs) organize representations for typologically and orthographically diverse languages into a shared parameter space, yet the nature of this internal organization remains elusive. In this work, we investigate which linguistic properties -- abstract language identity or surface-form cues -- shape multilingual representations. To do so, we analyze language-associated units across different model families and scales using the Language Activation Probability Entropy (LAPE) metric, and further decompose activations with Sparse Autoencoders. We find that these units are strongly conditioned on orthography: romanization induces near-disjoint representations that align with neither native-script inputs nor English, while word-order shuffling has limited effect on unit identity. Probing shows that typological structure becomes increasingly accessible in deeper layers, while causal interventions indicate that generation is most sensitive to units that are invariant to surface-form perturbations rather than to units identified by typological alignment alone. Overall, our results suggest that multilingual LMs organize representations around surface form, with linguistic abstraction emerging gradually without collapsing into a unified interlingua.\blfootnote{\rule{0pt}{1.8em}\raisebox{-0.2em}{\includegraphics[width=1em,height=1em]{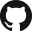}}\hspace{.5em}\parbox{\dimexpr\linewidth}{\url{https://github.com/loadthecode0/multilingual-interpretability}}}
\end{abstract}

\section{Introduction}

Language is an amalgamation of historical accidents, cognitive constraints, and cultural evolution. It is rarely a monolith; rather, it emerges as a layered outcome of interactions among peoples, geographies, and time \citep{12cea4fd-05b0-3201-bb45-108663b558ed, Toscano_Perry_Mueller_Bean_Galle_Samuelson_2008, 10.1098/rstb.2008.0145, beckner_language_cas, Evans_Levinson_2009, Michaud1898938}. Modern English illustrates this clearly: while it is taxonomically a West Germanic language, sharing core syntactic and phonological structure with German and Dutch, its lexicon is heavily shaped by Romance influence through Latin and French \citep{baugh2002history, crystal2003english, wardhaugh2014introduction}. When a sentence such as \emph{``the magnitude of liberty''} is processed, Latinate vocabulary is embedded within a Germanic grammatical frame \citep{Reppucci2017SpeakingDE}. This raises a fundamental question for modern auto-regressive language models (LMs): do they internally preserve such linguistic distinctions, or do they abstract away surface variations into a shared, language-agnostic representation?

This question becomes especially crucial in multilingual settings. When a model processes typologically distant languages such as English, Hindi, and Chinese, does it rely on distinct internal representations for each language, or does it converge toward a shared interlingual latent space? 
Insights from bilingual cognition show that shared semantic representations can coexist with segregated surface-form processing \citep{Costa2014-jq, MARIAN200370, Buchweitz2011-hs, Miozzo01022010}. However, within NLP, this distinction remains underexplored in modern auto-regressive multilingual models. Investigating these models across different parameter scales allows us to determine whether the trade-offs between surface-form processing and linguistic abstraction are mere artifacts of limited capacity or fundamental properties of multilingual architectures.

Recent work has begun to probe this question \citep{lape, kojima-etal-2024-multilingual, deng-etal-2025-unveiling, sae_lape}. Specifically, \citet{lape} introduced the Language Activation Probability Entropy (LAPE) metric to identify neurons that preferentially activate for specific languages in multilingual LMs. They showed that a relatively small subset of neurons concentrated primarily in early and late layers has a strong influence on language selection and can be causally manipulated to steer the output language. Subsequent work extended this approach using Sparse Autoencoders (SAEs), the method being referred to as \textsc{SAE-LAPE} ~\citep{sae_lape}, which decomposes dense activations into sparse latent features and performs selection of language-associated features in the latent space using LAPE. Related intervention-based analyses similarly suggest that language control can be induced by targeting carefully selected units 
\citep{gurgurov2025languagearithmeticssystematiclanguage, rahmanisa2025unveilinginfluenceamplifyinglanguagespecific}. These studies show that language-associated units exist and can be causally manipulated, but they leave open a key question: \emph{what linguistic properties do these language-associated units encode?} 

In this work, we systematically investigate this question by analyzing language-associated units at two complementary levels: raw model neurons in the MLP sublayers that directly affect generation, and sparse latent features extracted with SAEs for interpretability. Rather than assuming these units encode abstract language identity, we test their sensitivity to orthography, word order, and deeper linguistic structure. We study these representations across different model families and scales -- specifically in Llama-3.2-1B, Llama-3-8B, Gemma-2-2B, and Gemma-2-9B -- analyzing languages that span Latin, Cyrillic, Devanagari, Perso-Arabic, and logographic scripts. This diverse selection ensures our observations reflect broad architectural traits rather than scale-specific bottlenecks.

Our analysis is guided by four research questions: (i) \textbf{Language vs.\ script}: do language-associated units encode abstract language identity, or are they primarily tied to orthographic form? Furthermore, does semantic competence in a given script guarantee representational alignment? In particular, does romanizing a language (e.g., Hindi or Chinese written in Latin script) activate the same neurons as its native script? (ii) \textbf{Robustness to structural perturbation}: how stable are these units when word order is disrupted? (iii) \textbf{Typological alignment}: do language-associated units correlate with known typological properties, such as genealogy, phonology, or syntax, as captured by \texttt{lang2vec}~\citep{lang2vec}? (iv) \textbf{Layer-wise organization}: how does the accessibility of these properties vary across network depth, and how are they organized in deeper layers?

To answer these questions, we combine sparse feature extraction with a series of controlled experiments. We analyze the behaviour of language-associated units under script romanization, structural perturbations, typological probing, and causal intervention. Across these analyses, several consistent patterns emerge:
\begin{itemize}[nosep, wide, labelwidth=!, labelindent=0pt]
\item \textbf{Language-associated units are largely script-bound}: native and romanized variants of non-Latin languages activate almost disjoint sets of language-associated units, whereas shared scripts exhibit significant overlap. Notably, units associated with romanized non-Latin inputs align with neither their native counterparts nor English, indicating \textit{fragmented representations} within the LMs, even when the models exhibit high semantic competence on the romanized text (c.f. Sections~\ref{sec:romanization} and \ref{sec:discussion}).

\item \textbf{Disrupting word order has only a minor effect on unit identity}, suggesting reliance on lexical statistics or orthographic cues rather than syntactic structure (c.f. Section~\ref{sec:shuffling}).

\item \textbf{Units in deeper layers show stronger typological alignment}, indicating increased representational accessibility with depth (c.f. Section~\ref{sec:probing}). Causal interventions further show that functional importance during generation is more closely associated with invariance to surface perturbations than with typological alignment alone (c.f. Section~\ref{sec:causal}).

\end{itemize}
 
Together, these findings distinguish representational accessibility from functional necessity in multilingual LMs: language-associated units are closely tied to surface form, while deeper linguistic regularities become accessible with depth, and causal importance aligns more with invariance to surface perturbations than with representational alignment alone.

\begin{keytakeaway}
Language-associated units primarily encode surface form, and units invariant to surface perturbations play a central role in generation.
\end{keytakeaway}

\section{Related Work}

Prior work has shown that multilingual language models do not form a fully language-agnostic interlingua, but instead organize representations in a partially shared space structured by language identity and similarity \citep{johnson-etal-2017-googles, pires-etal-2019-multilingual, libovicky-etal-2020-language}. Neuron-level analyses further demonstrated that language control can be localized to specific internal units. In particular, \citet{lape} introduced the LAPE metric to identify language-selective neurons and showed that manipulating a small subset, often in early and late layers, can steer output language. Subsequent work confirmed that targeted interventions on such units enable controlled language switching \citep{kojima-etal-2024-multilingual, gurgurov2025languagearithmeticssystematiclanguage, rahmanisa2025unveilinginfluenceamplifyinglanguagespecific}. While these studies establish the functional relevance of language-associated units, they leave open what linguistic properties these units encode.

In parallel, SAEs have been proposed to decompose dense transformer activations into more interpretable sparse features \citep{DBLP:conf/cvpr/BauZKO017, shi2025routesparseautoencoderinterpret}, and have recently been applied to identify language-associated features in multilingual models \citep{sae_lape, deng-etal-2025-unveiling}. Separately, work on typology and script effects shows that orthography and transliteration can strongly shape multilingual representations and cross-lingual alignment \citep{lang2vec, artetxe-etal-2020-cross, jauhiainen-etal-2019-language}. 

Our work connects these threads by moving from identification to interpretation: we test whether language-associated units -- both raw neurons and sparse features -- encode abstract linguistic structure or are primarily driven by surface-form cues. In doing so, we contextualize recent literature surrounding the ``interlingua'' hypothesis, which often highlights semantic alignment and shared grammatical concepts across typologically diverse languages \citep{wendler2024do, schut2025do, brinkmann2025large, fierro2025how}. Our findings complement these works by demonstrating that while semantic alignment is achievable, it does not necessitate the topological collapse of representations into a single manifold. Instead, language-neutral components coexist with a persistent set of script-specific neurons. By identifying script as a primary barrier to global unification, our work reveals that what appears as a unified space is actually deeply fragmented when orthography varies. For a more detailed discussion of prior works, we refer the reader to Appendix~\ref{app:related-work}.

\section{Analysis Framework}
\label{sec:method}

\paragraph{Terminology.} 
We adopt the term \textit{unit} as a unifying abstraction for the atomic elements of representation. Specifically, a unit refers to either a raw neuron -- an individual element of the MLP's hidden activation vector -- or an SAE feature representing a single direction within the latent space of the SAE. Accordingly, we define a \textit{language-associated unit} as any unit that exhibits high selectivity for a specific target language, as quantified by the LAPE metric.

\paragraph{Identifying Language-Associated Units.}
Our analysis builds on the LAPE framework \citep{lape} and its sparse extension \textsc{SAE-LAPE} \citep{sae_lape} to identify language-associated structure in multilingual LMs. For each transformer layer $\ell$, we analyze both raw feed-forward (MLP) activations $h_\ell(x)$ and sparse latent representations obtained via pre-trained SAEs. Language association is quantified using LAPE: for each neuron or SAE feature $f$, we estimate its activation probability across languages and compute the entropy of this distribution. Units with low entropy and a dominant language are selected as \emph{language-associated}, yielding a set $\mathcal{N}_{\ell,L}$ for each layer and language. Details of the LAPE and SAE-LAPE procedures along with the hyperparameters used are provided in Appendix~\ref{app:lape}.

\paragraph{Models and Representations.}
We conduct experiments across multiple model families and scales, specifically Llama-3.2-1B, Llama-3-8B \citep{grattafiori2024llama3herdmodels}, Gemma-2-2B, and Gemma-2-9B \citep{gemmateam2024gemma2improvingopen}. Prior work has applied LAPE and SAE-based analyses to Llama-family and Gemma-family models \citep{lape, sae_lape, deng-etal-2025-unveiling}, motivating our choice of architectures and sparse decompositions. Following this line of work, we use open-sourced \textit{Top-K} SAEs\footnote{\url{https://huggingface.co/EleutherAI/sae-Llama-3.2-1B-131k}, \url{https://huggingface.co/EleutherAI/sae-llama-3-8b-32x}} for the Llama models and \textit{JumpReLU} SAEs for the Gemma models \citep{lieberum2024gemmascopeopensparse}, focusing on MLP sublayers. \textbf{For clarity of exposition, we primarily present results for Llama-3.2-1B in the core analysis sections of the main paper; corresponding analyses for Gemma-2-2B are provided in the Appendix, and validations on the larger 8B and 9B architectures are detailed in Section \ref{sec:discussion} and Appendix~\ref{app:scaling}.}

\paragraph{Experimental Design.}
We design a set of targeted experiments to probe what linguistic properties language-associated units encode, including (i) controlled script perturbations via romanization, (ii) robustness tests under word-order shuffling, (iii) typological probing against \texttt{lang2vec} features, and (iv) targeted causal interventions. As each experiment involves distinct language sets, perturbations, and evaluation protocols, we describe the detailed setups in the corresponding sections.

\section{Orthography as a Barrier to Latent Language Abstraction}
\label{sec:romanization}

A central question in multilingual representation learning is whether neurons or features identified as \emph{language-associated} encode abstract linguistic identity or merely respond to orthographic surface form. To disentangle these factors, we conduct a controlled romanization experiment that isolates script variation while holding lexical content and sentence structure fixed.

\paragraph{Experimental Setup.}
We use sentence-aligned data from the \texttt{dev} split of FLORES+\footnote{\url{https://huggingface.co/datasets/openlanguagedata/flores_plus}}, an extension of the FLORES-200 dataset \citep{nllb-24}, covering a typologically and orthographically diverse set of languages spanning Abugida, Abjad, Cyrillic, Logographic, and Syllabic scripts. For each non-Latin language, we construct a parallel Romanized corpus using the ICU Transliterator \citep{icu_transliterator}. Where applicable, we generate two Romanized variants: one preserving diacritics and one ASCII-only version with diacritics removed. Language-associated units are identified independently for native and Romanized inputs using the LAPE criterion for raw neurons and \textsc{SAE-LAPE} for sparse features, and overlap is quantified using Jaccard similarity. More detailed experimental details are provided in Appendix~\ref{app:romanization-setup}.

\begin{figure}[!t]
    \centering
    \begin{subfigure}[b]{0.49\linewidth}
        \centering
        \includegraphics[width=\linewidth]{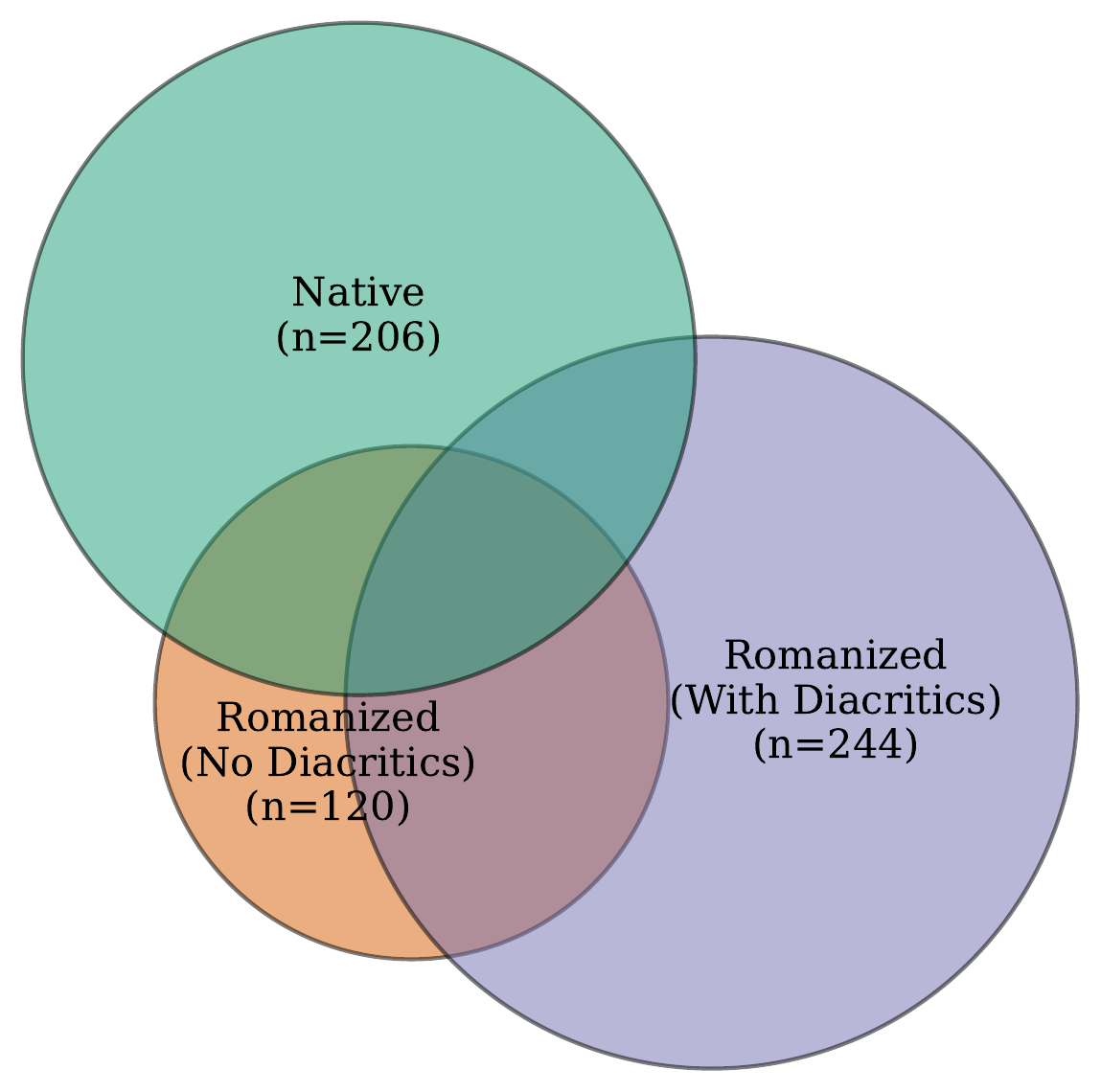}
        \caption{For Raw Neurons}
    \end{subfigure}
    \hfill
    \begin{subfigure}[b]{0.48\linewidth}
        \centering
        \includegraphics[width=\linewidth]{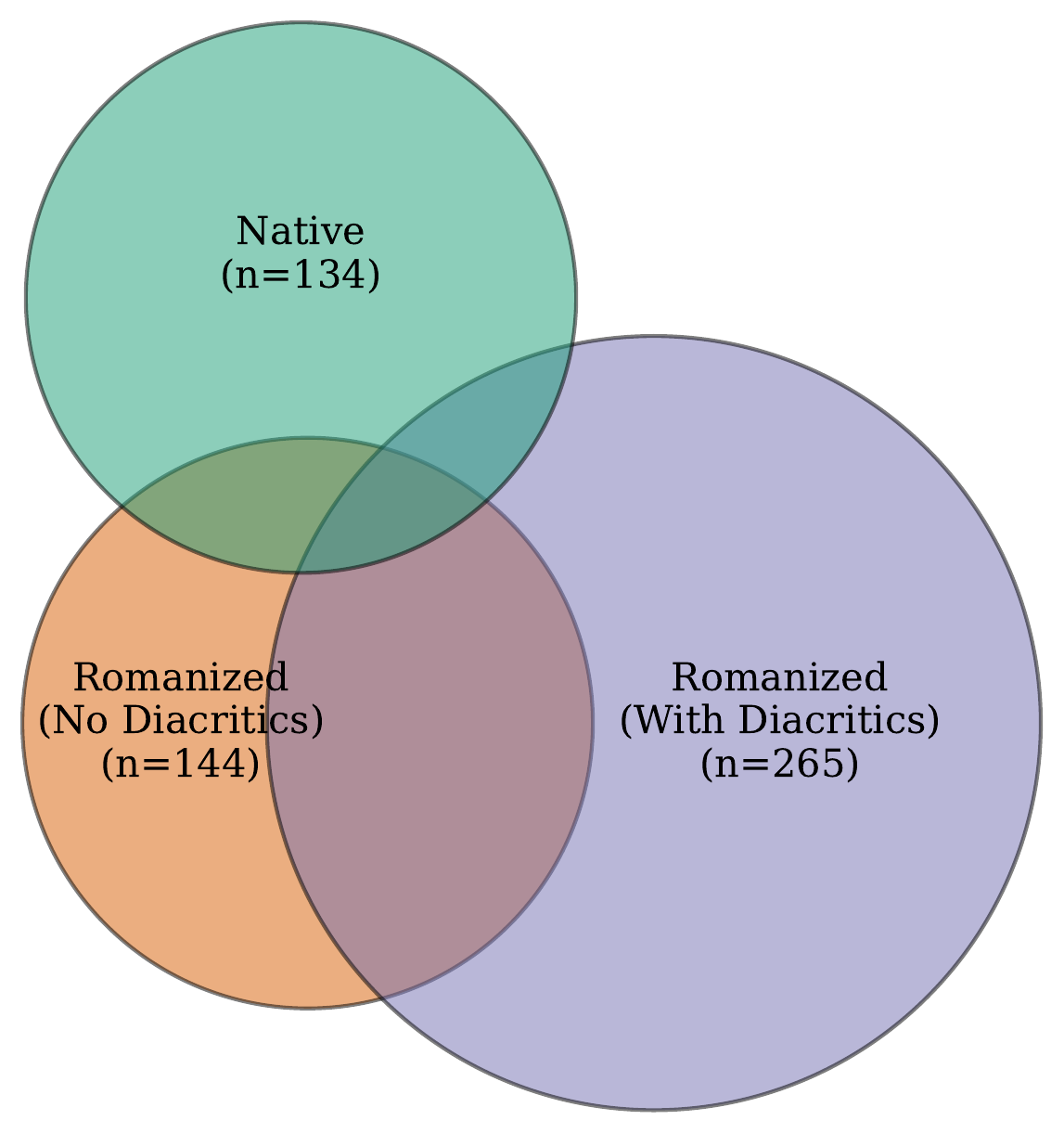}
        \caption{For SAE Features}
    \end{subfigure}
    \caption{Overlap of language-associated units for Hindi under script variation in Llama-3.2-1B. Euler diagrams show units shared among up to three languages for (a) raw neurons and (b) SAE features. Native, Romanized (with diacritics), and Romanized (without diacritics) inputs activate largely disjoint sets in both representations. Corresponding results for all languages, for both raw neurons and SAE features, and for Gemma-2-2B are shown in Figures~\ref{fig:romanization_venn_llama} and \ref{fig:romanization_venn_gemma} in Appendix~\ref{app:romanization_supplementary}.
}
    \label{fig:identity_fracture}
\end{figure}

\paragraph{Orthography Acts as a Barrier to Language Identity.}
If language-associated units encoded abstract linguistic identity, they would remain stable under changes in script.
Instead, Figure~\ref{fig:identity_fracture} shows near-complete fragmentation under romanization for Hindi (similar observations are also made for other languages, as shown in the Figures~\ref{fig:romanization_venn_llama} and \ref{fig:romanization_venn_gemma}).
Across both raw neurons and SAE features, native-script Hindi, Romanized Hindi with diacritics, and its ASCII-only variant activate largely disjoint sets of language-associated units, even when allowing overlap across multiple languages.
This fragmentation persists despite identical lexical content, indicating that language association in these models is strongly conditioned on orthographic form rather than abstract language identity.

\begin{takeaway}
\textbf{Language-associated units are tightly bound to orthography.}
Even minimal script changes induce near-disjoint unit sets in both raw neurons and sparse features.
\end{takeaway}

\begin{figure}[!t]
    \centering
    \includegraphics[width=\linewidth]{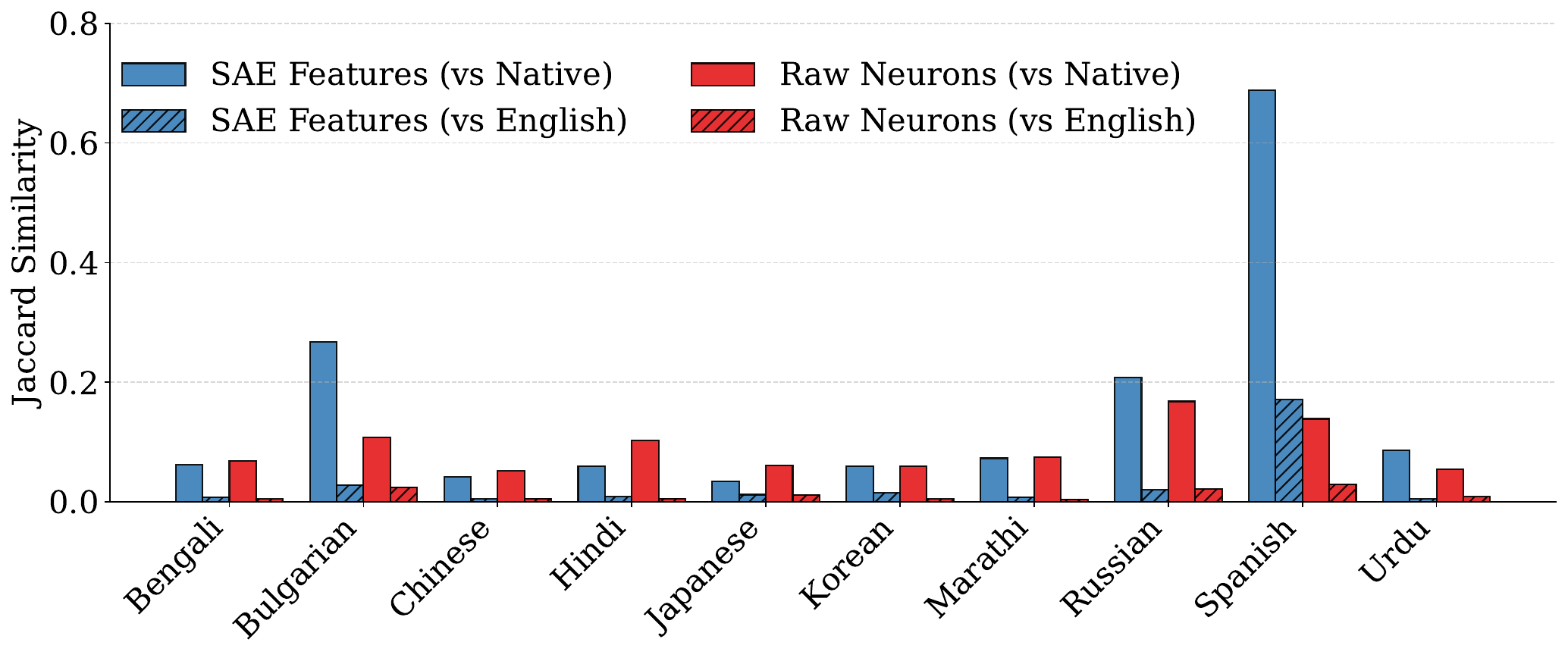}
    \caption{
    Jaccard similarity between Romanized and native-script or English language-associated units (\textbf{\textcolor{mplred}{raw neurons}} and \textbf{\textcolor{mplblue}{SAE features}}) in Llama-3.2-1B (see Figure~\ref{fig:jaccard_sae_raw_native_english_gemma} for Gemma-2-2B). Romanized inputs exhibit low overlap with their native-script counterparts and near-zero overlap with English in both representations, indicating limited cross-script alignment without convergence to English.}
    \label{fig:shared_isolation}
\end{figure}

\paragraph{Romanization Induces an Isolated Latent Subspace.}
Figure~\ref{fig:shared_isolation} examines whether Romanized inputs align with native-script or English representations when considering all language-associated units.
Across languages, overlap between Romanized and native-script representations remains consistently low (typically below $0.3$) for both raw neurons and SAE features, with higher overlap only for Spanish, which already uses the Latin script.
Crucially, overlap with English is near zero in all cases.
Together, these results show that Romanization neither recovers native-script representations nor induces convergence toward English.
Instead, Romanized text occupies a distinct, script-conditioned subspace that remains isolated even when considering shared language-associated units, effectively forming a third latent configuration that is neither native nor English.

\begin{takeaway}
\textbf{Romanization does not lead to Anglicization}. Romanized inputs form a distinct, script-conditioned latent subspace, separate from both native-script and English representations.
\end{takeaway}

\begin{figure}[!t]
    \centering
    \includegraphics[width=\columnwidth]{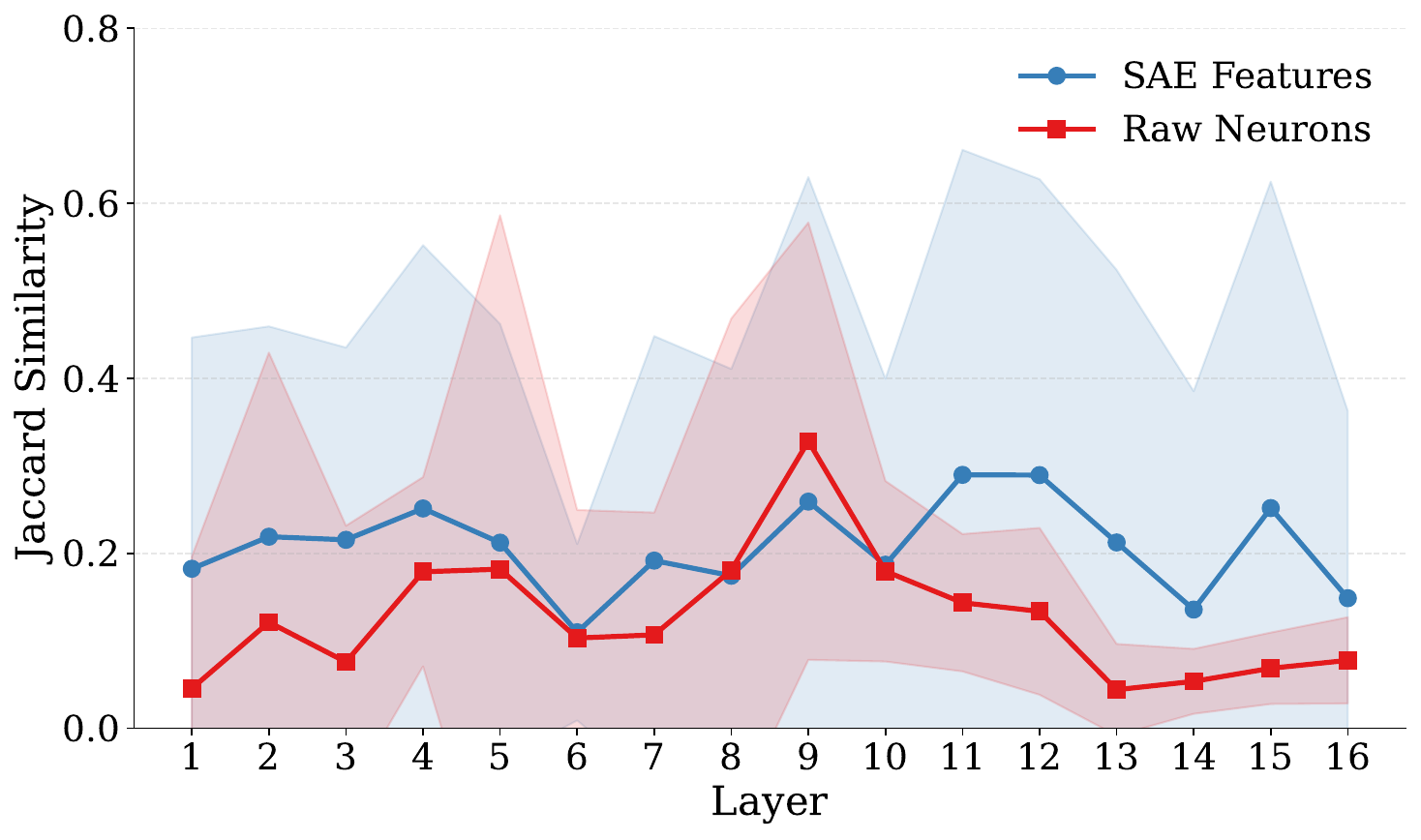}
    \caption{Layer-wise alignment between language-associated units for Native and Romanized inputs in Llama-3.2-1B (see Figure~\ref{fig:layerwise_trend_gemma} for Gemma-2-2B). The \textbf{\textcolor{mplred}{red line}} denotes average Jaccard similarity for \textbf{\textcolor{mplred}{raw neurons}}, and the \textbf{\textcolor{mplblue}{blue line}} for \textbf{\textcolor{mplblue}{SAE features}}; shaded regions indicate standard deviation across languages. Raw neurons show a modest mid-layer increase in overlap, while SAE features remain uniformly low across depth. In all cases, alignment remains far from convergence, indicating that representational separation persists beyond input tokenization.
}
    \label{fig:layerwise_trend}
\end{figure}

\paragraph{Limited Intermediate Alignment and Persistent Separation.}
Figure~\ref{fig:layerwise_trend} shows how language-associated units for Native and Romanized inputs align across layers in Llama-3.2-1B. 
While low overlap in early layers is expected due to disjoint token embeddings, this separation persists well beyond the input stage. 
\textbf{\textcolor{mplred}{Raw neurons}} exhibit a modest mid-layer increase in overlap, peaking around layer~$9$, but the alignment remains limited (Jaccard $\approx 0.3$) and never approaches convergence. 
In contrast, \textbf{\textcolor{mplblue}{SAE features}} show consistently low and flat overlap across all layers, indicating that sparse language-associated features remain strongly script-conditioned throughout the model. 
Together, these trends indicate that although dense activations briefly align surface-level statistics, the model ultimately maintains parallel, script-specific subspaces, revealing a limitation in abstraction rather than a trivial consequence of tokenization.

\paragraph{Implications for Model Capacity.}
The emergence of disjoint feature sets for native, Romanized, and even minor orthographic variants (e.g., diacritic vs.\ ASCII) points to a fundamental fragmentation of representational capacity. This aligns with recent observations that orthographic variations, such as the presence of diacritics, cause severe subword fragmentation and representational shifts in modern tokenizers and LMs \citep{inoue-etal-2026-diacritics}. We refer to this latent phenomenon as \emph{capacity fragmentation}: the model allocates separate internal features to encode superficially different realizations of the same language. Even highly shared features fail to fully unify these variants, suggesting that many purportedly language-agnostic representations remain implicitly conditioned on script.

\paragraph{Scaling and Semantic Competence.}
Crucially, this representational fragmentation is not merely an artifact of data sparsity or limited model capacity. As we discuss in Section \ref{sec:discussion}, this topological disjointness persists even in larger architectures (e.g., Llama-3-8B and Gemma-2-9B) that achieve higher semantic competence on romanized inputs.

\section{Robustness of Language-Associated Features to Structural Perturbations}
\label{sec:shuffling}

Section~\ref{sec:romanization} illustrates that language-associated features are highly sensitive to script, with minor orthographic changes inducing substantial reorganization. We complement this with a perturbation that preserves surface form but disrupts structure by applying controlled word-level shuffling. Unlike romanization, shuffling preserves token identity, frequency, and script while breaking local word order, allowing us to test whether language-associated features depend on syntactic structure or primarily reflect token-level and distributional cues.

\paragraph{Setup.}
For each language, we construct a shuffled version of the evaluation corpus by randomly permuting word order within sentences. Language-associated units are re-identified using the same \textsc{SAE-LAPE} procedure applied in earlier sections. Stability is measured via Jaccard similarity between the unit sets obtained from original and shuffled text. Additional experimental details and analyses are reported in Appendix~\ref{app:shuffling}.

\begin{figure}[!t]
    \centering
    \includegraphics[width=\columnwidth]{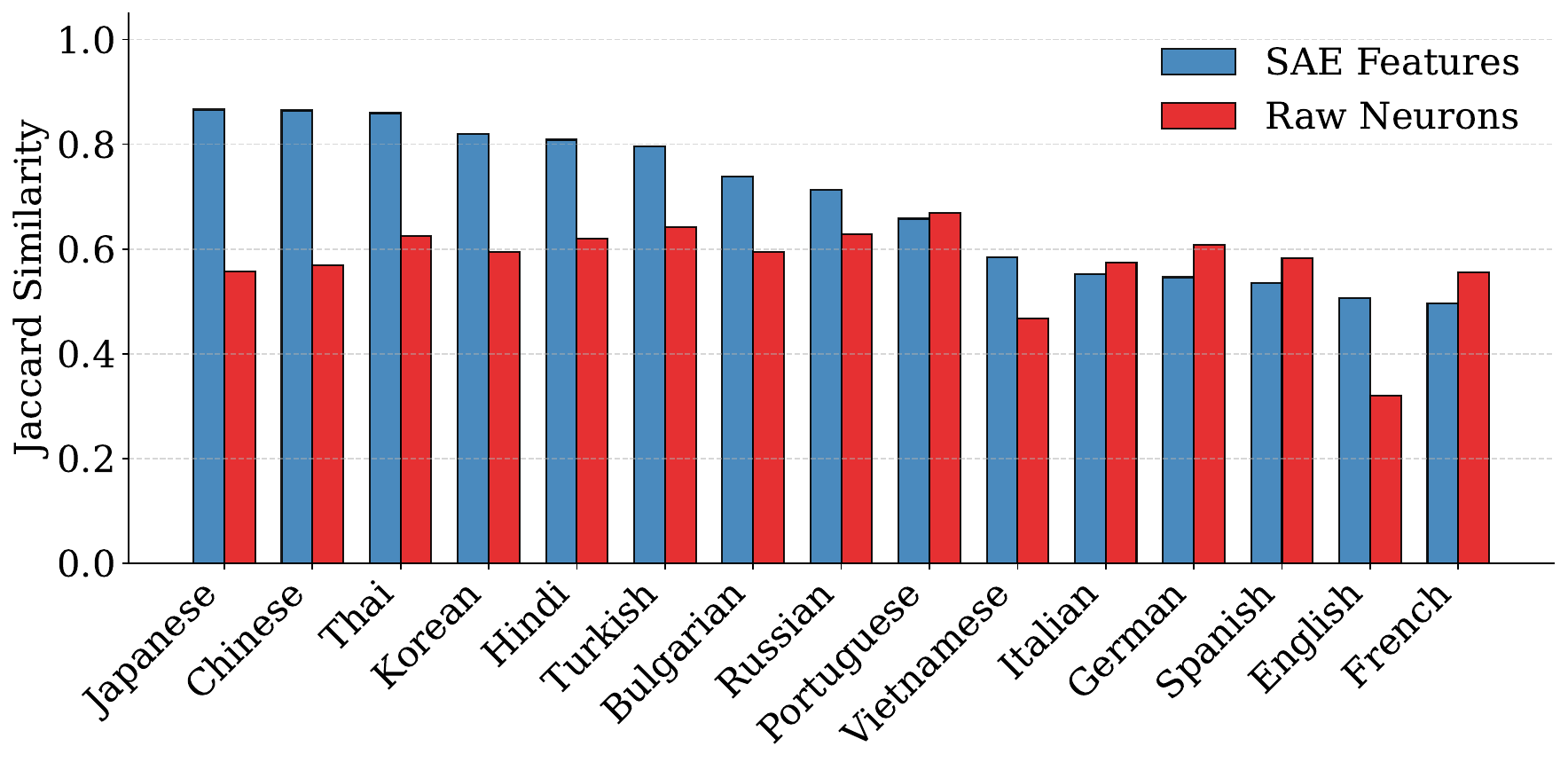}
    \caption{
Jaccard similarity between language-associated units identified from original and word-shuffled text in Llama-3.2-1B (see Figure~\ref{fig:shuffle_jaccard_gemma} for Gemma-2-2B).
\textbf{\textcolor{mplred}{Raw neurons}} exhibit consistently moderate-to-high overlap across languages, indicating robustness to word-order perturbation.
In contrast, \textbf{\textcolor{mplblue}{SAE features}} show only \emph{selective instability}: languages with distinctive scripts (e.g., Chinese, Japanese, Thai) remain highly stable, whereas several Latin-script languages exhibit somewhat reduced overlap, revealing sensitivity of sparse features to local distributional patterns disrupted by shuffling for these languages.
}
    \label{fig:shuffle_jaccard}
\end{figure}

\paragraph{Shuffling Reveals Selective Instability in Sparse Features.}
Figure~\ref{fig:shuffle_jaccard} shows that many languages retain a substantial fraction of their language-associated units under shuffling, indicating limited dependence on word order. However, this robustness varies across languages and representations. Languages with distinctive scripts such as Chinese, Japanese, Thai, Korean, and Cyrillic languages remain highly stable, with overlap often exceeding $0.7$, suggesting dominance of token identity and orthographic cues. In contrast, several Latin-script languages exhibit relative reductions in overlap specifically in \textbf{\textcolor{mplblue}{SAE features}}, indicating sensitivity of a subset of sparse features to local distributional or sequence-level statistics disrupted by shuffling. This selective instability is largely absent in \textbf{\textcolor{mplred}{raw neurons}}, which maintain stable overlap across languages, highlighting that dense representations encode language information redundantly, while sparse decompositions expose heterogeneity that is otherwise masked.

\paragraph{Activation Statistics Remain Stable.}
Although shuffling alters feature identity for some languages, it induces negligible changes in activation entropy or probability. Both language-level means and full distributions remain nearly identical before and after shuffling, indicating that shuffling affects \emph{which} features are selected rather than overall activation behavior (see Figures~\ref{fig:shuffling_dist_llama}, ~\ref{fig:shuffling_dist_gemma} and ~\ref{fig:shuffling_means} in Appendix~\ref{app:shuffling} for full distributional analyses and language-level means in case of both Llama and Gemma).

\paragraph{Implications.}
In contrast to the fragmentation induced by script changes (Section~\ref{sec:romanization}), word-order disruption leaves most language-associated representations intact. The limited instability that does occur is selective, appearing mainly in sparse features for languages that share script and subword statistics, and not in raw neurons.

\paragraph{Robustness at Scale.}
Furthermore, as detailed in Section \ref{sec:discussion}, this robustness to structural perturbation consistently holds across larger parameter scales (Llama-3-8B and Gemma-2-9B), reinforcing that language-associated units fundamentally prioritize surface form over syntactic structure regardless of model capacity.

\begin{takeaway}
\textbf{Language-associated units are largely insensitive to word order}, 
while sparse features expose limited, language-dependent reliance on local distributional cues.
\end{takeaway}

\section{Typological Structure Revealed by Probing}
\label{sec:probing}

Sections~\ref{sec:romanization} and~\ref{sec:shuffling} show that language-associated units are strongly shaped by surface form: script changes induce near-complete reorganization, while word-order perturbations leave many units intact. We now ask whether, despite this surface sensitivity, model representations encode deeper linguistic structure in a linearly accessible form. Specifically, we use probing to characterize \emph{where} typological information is concentrated and \emph{when} it emerges across model depth.

\paragraph{Setup.}
We probe both raw MLP activations and SAE-based representations against typological features from \texttt{lang2vec}~\citep{lang2vec}. For each layer, linear probes are trained with cross-validation over languages, and performance is summarized using the average of family-wise maximum $R^2$ scores. We report results across different neuron subsets induced by romanization and shuffling (e.g., condition-specific vs.\ overlap sets). Full probing details are provided in Appendix~\ref{app:probing}.

\begin{figure}[t]
    \centering
    \includegraphics[width=0.9\linewidth]{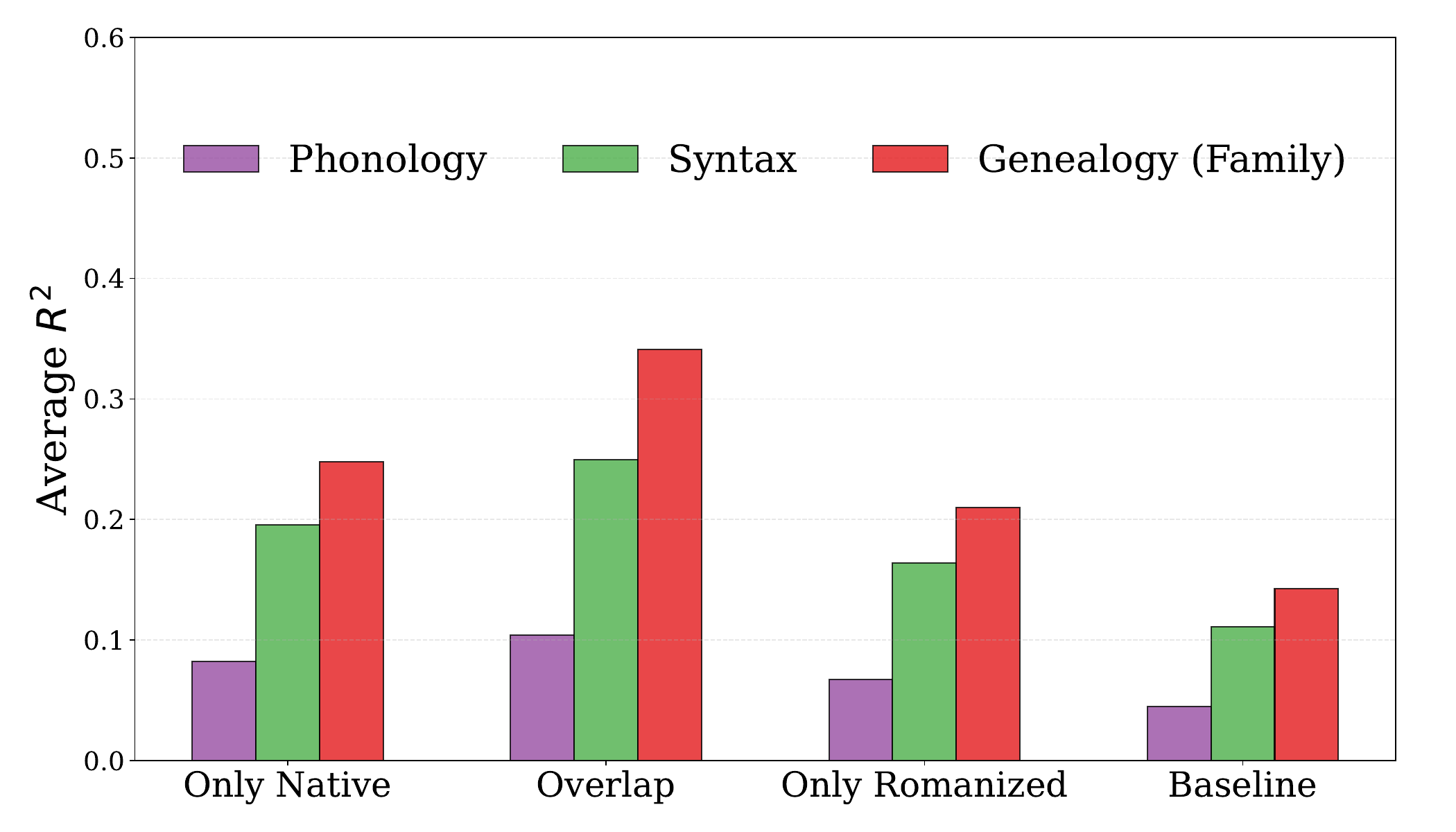}
    \caption{Average family-wise probing $R^2$ scores across neuron subsets induced by \textit{romanization} in Llama-3.2-1B (raw neurons).
Neurons overlapping between native and romanized inputs exhibit the strongest typological alignment, while script-specific subsets encode weaker signal.
\textit{Baseline} denotes probing over the pooled set of all neurons that were selected for either native or romanized inputs (across all layers), serving as a non-selective reference.
Corresponding results for Llama-3.2-1B (SAE features) and for Gemma-2-2B using both raw neurons and SAE features are shown in Figures~\ref{fig:romanization_probing_llama_sae}, \ref{fig:romanization_probing_gemma_raw}, and~\ref{fig:romanization_probing_gemma_sae} in Appendix~\ref{app:romanization_probing_correlation}.}

    \label{fig:romanization_probing_llama_raw}
\end{figure}

\paragraph{Typological Structure Aligns with Invariance to Script.}
Figure~\ref{fig:romanization_probing_llama_raw} shows probing results across neuron subsets, in Llama-3.2-1B, induced by romanization (see Figure~\ref{fig:romanization_probing_llama_sae} for SAE-features in Llama, and Figures~\ref{fig:romanization_probing_gemma_raw} and \ref{fig:romanization_probing_gemma_sae} for Gemma-2-2B).
Across both raw neurons and SAE features, a consistent pattern emerges: \emph{neurons preserved across native and romanized inputs exhibit the strongest typological alignment}.
Overlap subsets dominate across genealogical, syntactic, and phonological families, while script-specific subsets (native-only or romanized-only) encode substantially weaker typological signal.
This directly connects to Section~\ref{sec:romanization}: the same units that are invariant to orthographic change are those that preferentially encode deeper linguistic structure.
Together, these results indicate that typological abstraction is not tied to language-specific or script-specific units, but instead concentrates in representations that are robust to script variation.

\begin{figure}[t]
    \centering
    \includegraphics[width=0.9\linewidth]{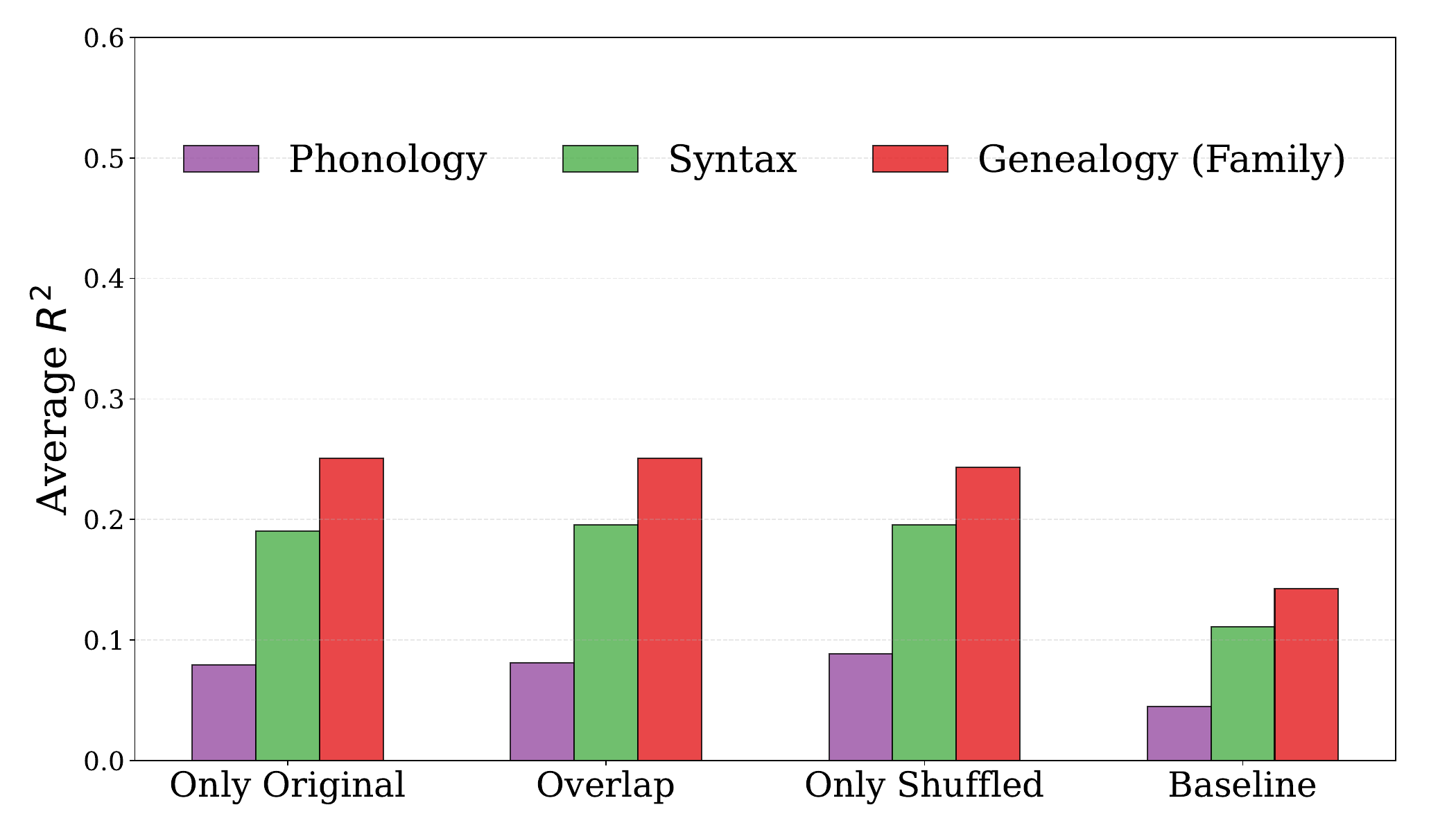}
    \caption{Average family-wise probing $R^2$ scores across neuron subsets induced by \textit{word-order shuffling} in Llama-3.2-1B (raw neurons).
Neurons specific to original text, shuffled text, and their overlap exhibit comparable typological alignment, indicating that sensitivity to word order is largely decoupled from typological information.
\textit{Baseline} denotes probing over the pooled set of all neurons selected for either condition, serving as a non-selective reference.
Corresponding results for Llama-3.2-1B (SAE features) and for Gemma-2-2B using both raw neurons and SAE features are shown in Figures~\ref{fig:shuffling_probing_llama_sae}, \ref{fig:shuffling_probing_gemma_raw}, and~\ref{fig:shuffling_probing_gemma_sae} in Appendix~\ref{app:shuffling_probing_correlation}.}
    \label{fig:shuffling_probing_llama_raw}
\end{figure}

\paragraph{Typological Structure Does Not Prefer Order-Invariant Units.}
In contrast, probing under word-order shuffling reveals a qualitatively different pattern.
Figure~\ref{fig:shuffling_probing_llama_raw} shows that typological alignment is \emph{comparable across normal-only, shuffled-only, and overlap subsets}.
This holds for both raw and sparse representations, although overall scores are lower for SAE features.
Unlike romanization, invariance to word order does not preferentially select typologically informative units.
This observation aligns with Section~\ref{sec:shuffling}: while shuffling leaves many language-associated units intact, this robustness does not correspond to a privileged locus of linguistic abstraction.

\begin{figure}[!t]
    \centering
    \includegraphics[width=0.9\linewidth]{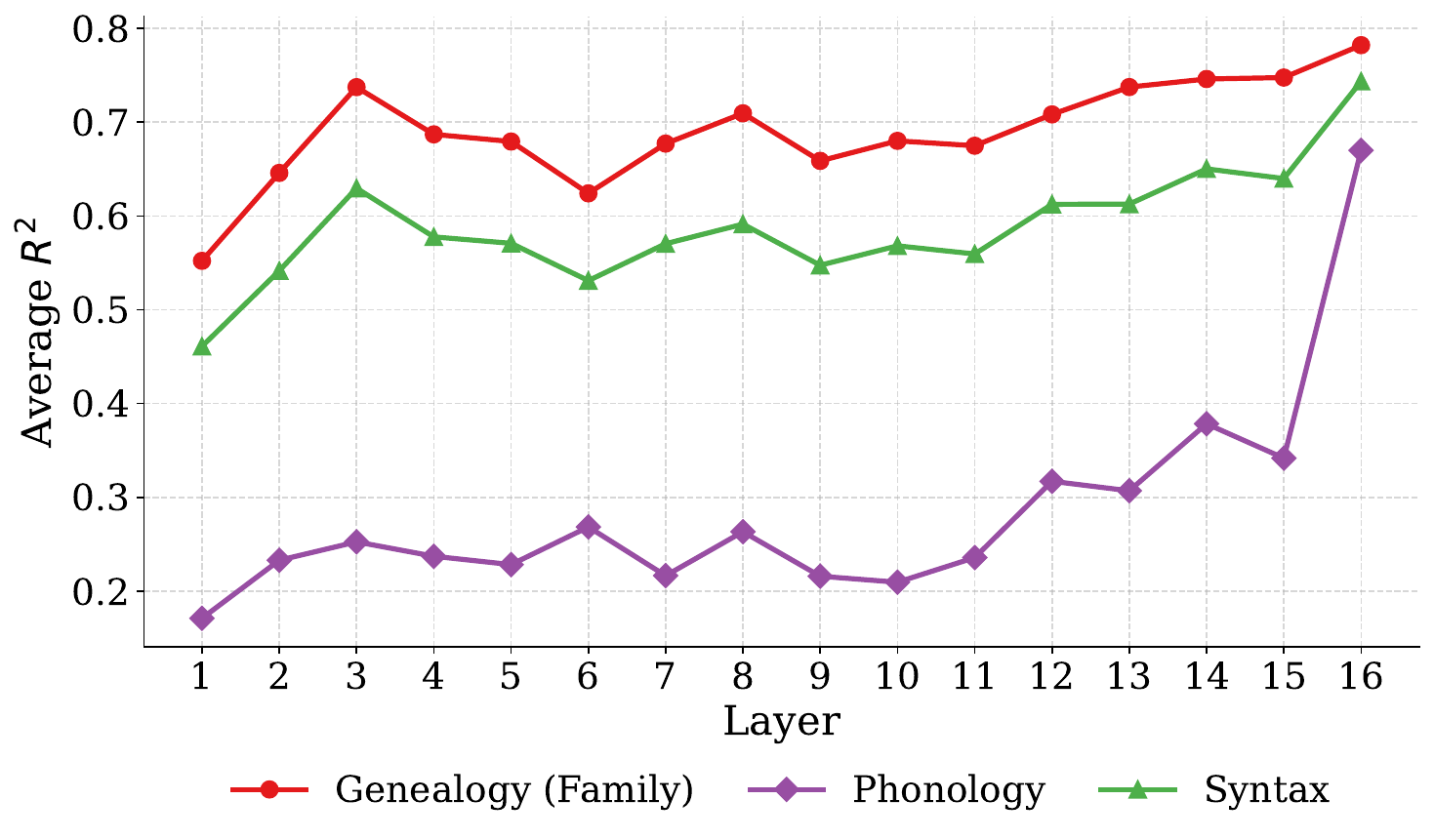}
    \caption{Average probing $R^2$ scores across layers for SAE features in Llama-3.2-1B, grouped by typological family. \textbf{\textcolor{mplred}{Genealogical}} properties are accessible from early layers, while more abstract features such as \textbf{\textcolor{mplpurple}{phonology}} emerge mainly in deeper layers. Corresponding results for raw neurons and Gemma-2-2B show the same hierarchy (Figures~\ref{fig:llama_raw_sae_layerwise},~\ref{fig:gemma_raw_sae_layerwise}).
}
    \label{fig:probing_family_avg}
\end{figure}

\paragraph{Depth-Dependent Emergence of Linguistic Abstraction.}
While invariance determines \emph{where} typological information resides, model depth determines \emph{when} it becomes accessible.
We illustrate this hierarchy using SAE features, where typological trends are most interpretable; raw activations show the same qualitative pattern (Appendix~\ref{app:probing}).
Figure~\ref{fig:probing_family_avg} shows that \textbf{\textcolor{mplred}{genealogical}} properties are linearly decodable from early layers, whereas more abstract \textbf{\textcolor{mplpurple}{phonological}} features emerge only in the deepest layers.
This hierarchy suggests that linguistic abstraction is constructed gradually with depth rather than encoded uniformly across the model.

\paragraph{From Representational Accessibility to Functional Testing.}
Probing shows that typological information becomes increasingly linearly accessible in deeper layers, particularly in script-invariant representations.
However, probing alone does not establish functional necessity.
In Section~\ref{sec:causal}, we therefore test whether units identified by their invariance properties play a causal role in generation. Furthermore, we synthesize these structural findings with downstream semantic competence in Section \ref{sec:discussion}.

\begin{takeaway}
\textbf{Typological structure emerges with depth and is strongest in script-invariant representations}. Abstraction remains distributed across units.
\end{takeaway}

\section{Causal Roles of Script- and Structure-Invariant Units}
\label{sec:causal}

Sections~\ref{sec:romanization}-\ref{sec:probing} show how language-associated units vary with script, word order, and typological structure. We now test whether these distinctions reflect \emph{functional necessity} during generation by performing targeted causal interventions on neuron sets defined solely by their invariance properties. Full experimental details, statistical tests, and qualitative analyses are provided in Appendix~\ref{app:causal}.

\paragraph{Setup.}
All interventions are performed on raw MLP activations. While we focus our main text exposition on Llama-3.2-1B, we concurrently validate all interventions on both Llama-3-8B and Gemma-2-2B to ensure causal effects hold across different architectures and scales. Neuron sets are defined by invariance to script or word-order perturbations (Sections~\ref{sec:romanization},~\ref{sec:shuffling}). For romanization-derived sets, we perform cross-language mean replacement; for shuffling-derived sets, we apply simultaneous zero ablation across all layers. Effects are compared against matched random controls using perplexity on FLORES+ dev examples. Statistical significance is assessed via paired $t$-tests; exact $p$-values are reported in Appendix Tables~\ref{tab:causal_shuffling} and~\ref{tab:causal_romanization}.

\paragraph{Script-Invariant Neurons Support Stable Generation Under Perturbation.}
Using neuron sets derived from the romanization analysis (Section~\ref{sec:romanization}), we perform cross-language mean ablations between Hindi and English (Table~\ref{tab:causal_romanization_main}). \textbf{Overlap neurons}, which remain active across native and romanized scripts, exhibit only mild and asymmetric perplexity changes under cross-language replacement; while statistically significant ($p<0.05$; Table~\ref{tab:causal_romanization}), these effects are small, indicating that these neurons occupy a largely script-invariant subspace. In contrast, \textbf{only-native neurons} show extreme sensitivity: replacing English-only-native activations with Hindi means causes severe degradation, while the reverse yields large apparent perplexity improvements. Qualitative inspection reveals that the latter corresponds to language switching rather than improved modeling, with generations collapsing into fluent English (Appendix~\ref{app:causal}). Crucially, these effects generalize: in both Llama-3-8B and Gemma-2-2B (Appendix Tables~\ref{tab:causal_romanization} and~\ref{tab:romanization_ablation_gemma}), ablating only-native Hindi neurons causes dramatic perplexity changes (e.g., $PPL_{\text{ratio}}^{\text{target}} = 7.74$ in Llama-3-8B), confirming extreme sensitivity. Together, these results causally validate Section~\ref{sec:romanization}, showing that script-specific neurons anchor surface realization and language identity, while script-invariant neurons support stable generation under orthographic perturbation.

\begin{table}[t]
\centering
\small
\setlength{\tabcolsep}{6pt}
\begin{tabular}{l l c c}
\toprule
Language & Neuron set &
$PPL_{\text{ratio}}^{\text{target}}$ &
$PPL_{\text{ratio}}^{\text{random}}$ \\
\midrule
English & overlap     & 0.95 & 0.99 \\
English & only-native & 1.50 & 0.96 \\\hline
Hindi & overlap     & 1.05 & 0.98 \\
Hindi & only-native & 0.31 & 0.97 \\
\bottomrule
\end{tabular}
\caption{
Cross-language mean ablations for romanization-derived neuron sets in Llama-3.2-1B (see Table~\ref{tab:romanization_ablation_gemma} for Gemma-2-2B).
$PPL_{\text{ratio}}^{\text{target}}$ denotes perplexity relative to clean runs, and
$PPL_{\text{ratio}}^{\text{random}}$ reports the same for matched random controls.
All target effects are statistically significant ($p<0.05$; Table~\ref{tab:causal_romanization}).
Ratios below $1$ reflect language switching rather than improved modeling.
}
\label{tab:causal_romanization_main}
\end{table}

\paragraph{Word-Order-Invariant Neurons Support Core Language Modeling.}
We next examine neuron sets derived from the shuffling analysis in Section~\ref{sec:shuffling} using simultaneous zero ablation (Table~\ref{tab:causal_shuffling_main}). Across all languages, \textbf{overlap neurons} -- those that remain active under word-order shuffling -- cause substantially larger perplexity increases than matched random controls, with all effects statistically significant ($p<0.05$; Table~\ref{tab:causal_shuffling}). In contrast, \textbf{only-unshuffled neurons} produce much weaker effects and often reduce perplexity, indicating that order-sensitive signals are largely redundant for generation. This causal dissociation mirrors the identification results in Section~\ref{sec:shuffling}: neurons invariant to structural perturbation are functionally necessary for stable language modeling, while order-sensitive neurons encode auxiliary or brittle patterns. Qualitatively, only overlap-neuron ablations induce systematic failures such as within-word script mixing and abrupt language switching (Appendix Figure~\ref{fig:shuffling_qualitative}), further supporting their causal role. These causal dynamics are highly consistent across architectures, with both Llama-3-8B and Gemma-2-2B exhibiting similar severe degradation and identical qualitative failure modes specifically when shuffling-invariant overlap neurons are ablated (see Appendix Tables~\ref{tab:causal_shuffling} and~\ref{tab:shuffling_ablation_gemma}).

\begin{table}[t]
\centering
\small
\setlength{\tabcolsep}{5pt}
\begin{tabular}{l l c c}
\toprule
Language & Neuron set & $PPL_{\text{ratio}}^{\text{target}}$ & $PPL_{\text{ratio}}^{\text{random}}$ \\
\midrule
English & overlap         & 1.12 & 0.95 \\
English & only-unshuffled & 0.96 & 1.04 \\\hline
Hindi & overlap         & 2.79 & 1.06 \\
Hindi & only-unshuffled & 1.08 & 0.95 \\
\bottomrule
\end{tabular}
\caption{
Zero-ablation results for shuffling-derived neuron sets in Llama-3.2-1B (see Table~\ref{tab:shuffling_ablation_gemma} for Gemma-2-2B).
$PPL_{\text{ratio}}^{\text{target}}$ reports perplexity relative to clean runs after ablating the specified neuron set, and
$PPL_{\text{ratio}}^{\text{random}}$ reports the same for matched random controls.
All overlap-neuron effects differ significantly from random controls ($p<0.05$; Table~\ref{tab:causal_shuffling}).
}
\label{tab:causal_shuffling_main}
\end{table}

\paragraph{Implications for Language Control and Abstraction.}
Across both romanization- and shuffling-based interventions, causal importance consistently tracks invariance to surface perturbations.
Neurons that remain stable under script or word-order variation are more functionally necessary for generation, whereas surface-sensitive neurons primarily anchor realization.
While probing in Section~\ref{sec:probing} shows that typological structure becomes increasingly decodable with depth, our causal interventions do not isolate a small set of neurons whose manipulation selectively disrupts such structure.
Instead, causal effects are associated with invariance properties, suggesting that language control in these models is mediated by robustness to surface variation rather than by a single, localized abstraction module.

\begin{takeaway}
\textbf{Causal importance aligns with invariance to surface perturbations.}
Neurons stable under script or word-order variation are necessary for generation, while probing reflects representational structure rather than direct control.
\end{takeaway}

\section{Discussion and Scaling Analysis}
\label{sec:discussion}

Our results show that multilingual models do not converge to a fully abstract interlingua. Instead, representations are organized around surface-form cues, especially script, while deeper layers support abstraction without unifying script-conditioned subspaces.

\paragraph{Robustness at Scale and Semantic Competence.}
To ensure these findings are not limited by model capacity, we validated our core experiments on larger architectures (Llama-3-8B and Gemma-2-9B). As shown in Appendix~\ref{app:scaling} (Figures~\ref{fig:shared_isolation_8b} --~\ref{fig:layerwise_trend_9b}), representational fragmentation persists at scale: romanized inputs maintain low overlap with native-script counterparts and fail to converge toward English. Conversely, robustness to word-order shuffling remains consistently high across these larger models (c.f. Figure~\ref{fig:shuffle_jaccard_scale}, Appendix~\ref{app:scaling}). Furthermore, to rule out data sparsity (i.e., the models simply failing to comprehend romanized text), we evaluated translation performance. As detailed in Appendix~\ref{app:scaling} (Table~\ref{tab:translation_romanized}), larger models achieve substantial semantic competence on romanized inputs. This confirms that models possess the requisite knowledge, but internally process different scripts through disjoint subspaces as a persistent architectural trait rather than a training deficiency.

\paragraph{Implications for Cross-Lingual Transfer.}
The strong dependence on orthography suggests that cross-lingual transfer is more fragile than often assumed. Romanized inputs neither recover native-script representations nor align with English, even when considering shared language-associated units. Instead, they occupy distinct latent subspaces, helping explain why transliteration or script normalization alone yields limited gains without explicit adaptation or supervision.

\paragraph{Orthography, Control, and Robustness.}
Our findings offer an alternative explanation for prior observations that changing the language or script of a prompt can alter model behavior, including safety-related responses~\citep{deng2024multilingual, yong2023lowresource}. If language-associated units are tightly coupled to orthography, script changes may route inputs through different internal subspaces, yielding divergent outputs. This suggests that some language-based control and jailbreak effects may stem from surface-form routing rather than semantic differences.


\section{Conclusion}

In this study, we show that language-associated units in multilingual LMs are primarily organized around surface-form cues, with script acting as a primary barrier. While typological structure becomes more accessible at deeper layers, our causal analyses reveal that stable generation depends mostly on units invariant to surface perturbations. By validating these findings across model scales, we confirm that LMs process different scripts through disjoint latent spaces despite high semantic competence, showing that multilingual abstraction remains limited by orthography rather than forming a fully unified interlingua.
\section*{Limitations}


While we validate our core representational and causal findings on models up to 9B parameters, evaluating these phenomena on massive-scale frontier models remains an important direction for future work, as models at much larger parameter scales may eventually exhibit different trade-offs between surface-form routing and abstraction. In addition, our analysis centers on feed-forward (MLP) activations and their sparse decompositions, and does not examine other architectural components such as attention heads or embedding layers. Finally, while our interventions assess the causal role of identified units at inference time, we do not study the training dynamics through which these representations emerge.

\section*{Ethical Considerations}
This work analyzes internal representations of multilingual LMs using publicly available pretrained models and established linguistic resources. While we generate and release systematically perturbed (romanized and shuffled) versions of existing evaluation sets, we do not deploy systems in user-facing contexts or evaluate downstream social applications. Our findings highlight how script and surface-form variation can influence internal processing, with potential implications for robustness and safety generalization across languages. We emphasize that our goal is interpretability and analysis rather than exploitation, and we do not propose methods for bypassing safeguards or inducing harmful behavior. Overall, this work aims to support safer and more transparent multilingual model development by clarifying how language-associated representations are organized internally.

\section*{Acknowledgements}
Anwoy Chatterjee gratefully acknowledges the support of the Google PhD Fellowship. Tanmoy Chakraborty acknowledges the support of
the Anusandhan National Research Foundation
(Grant no: DST/INT/USA/NSF-DST/Tanmoy/P-2/2024) and
Rajiv Khemani Young Faculty Chair Professorship in Artificial Intelligence. The authors acknowledge the support of the Google GCP Grant.

\bibliography{custom}


\appendix

\section*{Appendix Contents}
\label{app:toc}

Below we provide an overview of the appendix. These sections are intended to support the core claims by providing methodological details and extended scaling results.

\begin{itemize}[leftmargin=*, itemsep=10pt]

\item \hyperref[app:faq]{\textbf{Appendix A: Frequently Asked Questions (FAQs).}}  
\textit{Addresses common questions regarding representational fragmentation, scale, and causal control.}

\item \hyperref[app:related-work]{\textbf{Appendix B: Extended Related Work.}}  
\textit{Provides a detailed discussion of prior work on multilingual representations, language-associated units, sparse autoencoders, typology, and script effects.}

\item \hyperref[app:lape]{\textbf{Appendix C: Identifying Language-Associated Units with LAPE and SAE-LAPE.}}  
\textit{Explains the identification frameworks and provides the hyperparameter thresholds used for both neurons and sparse features.}

\item \hyperref[app:romanization-setup]{\textbf{Appendix D: Script Perturbation Experiments (Romanization).}}  
\textit{Describes dataset construction, transliteration procedures, and distributional analyses supporting the script romanization findings.}

\item \hyperref[app:shuffling]{\textbf{Appendix E: Structural Perturbation Experiments (Word Shuffling).}}  
\textit{Reports supplementary results on shuffled inputs, aggregate overlap analyses, and stability of activation statistics.}

\item \hyperref[app:probing]{\textbf{Appendix F: Probing Typological Structure Across Layers.}}  
\textit{Details the ridge regression probing setup and provides comparative layer-wise informativeness across models.}

\item \hyperref[app:causal]{\textbf{Appendix G: Causal Interventions on Invariant Neuron Sets.}}  
\textit{Provides simultaneous ablation protocols and qualitative failure modes showing script mixing during generation.}

\item \hyperref[app:scaling]{\textbf{Appendix H: Extended Scaling and Semantic Competence Analysis.}}  
\textit{Documents the persistence of representational fragmentation at the 8B and 9B scales and provides translation benchmarks ruling out data sparsity as an explanation.}

\end{itemize}

\section{Frequently Asked Questions (FAQs)}
\label{app:faq}

\begin{enumerate}[leftmargin=*, itemsep=6pt]

\item \textbf{Do language-associated units imply the existence of a universal interlingua?}  
No. While language-associated units are clearly identifiable and can influence model behavior, our results show that they are predominantly sensitive to surface-form cues such as script and token distribution. 

\item \textbf{Is the observed script sensitivity simply an artifact of tokenization?}  
Tokenization necessarily introduces distinct input embeddings across scripts, but our analysis goes beyond early-layer effects. We observe that alignment remains low even in intermediate layers, indicating that script sensitivity is not merely a tokenizer artifact but reflects persistent representational fragmentation within the model.

\item \textbf{Why use 1B and 2B models for the main exposition?}  
We center our primary exposition on Llama-3.2-1B and Gemma-2-2B to enable extensive, computationally intensive representational sweeps and causal interventions across many layers and languages. However, to ensure our findings are not artifacts of limited capacity, we explicitly validate our core experiments on larger models (Llama-3-8B and Gemma-2-9B), confirming these representational properties hold at scale.

\item \textbf{Do these findings generalize to larger models?}  
Yes. As detailed in our scaling analysis, we validate our findings on Llama-3-8B and Gemma-2-9B. We observe that representational fragmentation under script variation, as well as robustness under structural perturbation, persist at these larger scales. Crucially, this fragmentation remains even though these models exhibit strong semantic translation competence on romanized inputs, demonstrating that script-conditioned subspaces are a persistent architectural trait rather than a symptom of undertraining or data sparsity.

\item \textbf{Does strong probing performance imply functional importance?}  
No. Probing reveals that typological properties become increasingly linearly accessible in deeper layers, but causal interventions show that functional importance aligns with invariance to surface perturbations. This reinforces the view that linear decodability does not imply causal control.

\item \textbf{Why analyze both raw neurons and SAE features?}  
Raw neurons directly govern model behavior, while SAE features provide an interpretable decomposition of these activations. Analyzing both allows us to separate functional relevance from interpretability and avoid over-attributing abstract meaning to sparse features alone.

\item \textbf{What is the main takeaway for interpreting \textit{language-associated neurons}?}  
Language-associated units exist and matter, but they primarily reflect surface-form processing rather than abstract language identity. 

\end{enumerate}

\section{Extended Related Work}
\label{app:related-work}

\subsection{Language-Associated Units and Multilingual Representations}

Understanding how multilingual LMs encode language identity has become a central question in interpretability and cross-lingual modeling. Early multilingual neural machine translation (NMT) systems already suggested that jointly trained models do not form a fully language-agnostic interlingua, but instead organize representations in a partially shared space structured by language identity and similarity \citep{johnson-etal-2017-googles, kudugunta-etal-2019-investigating}. Subsequent analyses showed that encoder representations cluster by genealogical and typological proximity, with high-resource languages occupying more stable regions of the latent space \citep{pires-etal-2019-multilingual, libovicky-etal-2020-language}.

More recently, investigation at the neuron level has provided evidence that language identity can be localized to specific internal units. \citet{lape} introduced LAPE to identify neurons that preferentially activate for individual languages in multilingual LMs, showing that a small subset of neurons, often concentrated in early and late layers, exerts disproportionate control over language selection. Contemporary works also showed that targeted interventions on such neurons can reliably steer output language, even without modifying input prompts \citep{kojima-etal-2024-multilingual,gurgurov2025languagearithmeticssystematiclanguage,rahmanisa2025unveilinginfluenceamplifyinglanguagespecific}. These observations establish that language control is not purely emergent at the output layer but is mediated by identifiable internal mechanisms.

Earlier representational studies, however, caution against interpreting such units as encoding abstract language identity \citep{wu-dredze-2020-languages, libovicky2019languageneutralmultilingualbert}. Analyses of multilingual NMT and representation spaces show substantial mixing across languages, particularly in middle layers, with language separation re-emerging closer to the output where lexical constraints dominate \citep{kudugunta-etal-2019-investigating}. This layered organization parallels findings in bilingual cognition, where shared semantic representations coexist with partially segregated lexical and orthographic processing streams \citep{MARIAN200370,Costa2014-jq}.

Our work builds on this literature but departs in emphasis. Rather than asking whether language-associated units exist, we ask what linguistic properties they encode. Specifically, we test whether such units reflect abstract language identity or are instead driven by surface-form cues such as script and token distributions, a distinction that remains underexplored in prior neuron-level studies. While our primary exposition focuses on highly capable 1B and 2B parameter auto-regressive models to enable computationally intensive feature sweeps, we explicitly validate our core findings on larger architectures (up to 9B parameters) to ensure our conclusions regarding orthography and abstraction hold at scale.

\subsection{Sparse Autoencoders and Feature-Level Interpretability}

SAEs have recently emerged as a promising tool for disentangling dense transformer activations into more interpretable, monosemantic latent features. The central idea -- that sparsity can separate overlapping signals into distinct dimensions -- has strong precedents in vision, where network dissection methods link individual units to human-interpretable concepts \citep{DBLP:conf/cvpr/BauZKO017}. In language models, sparse methods have been shown to isolate features corresponding to factual recall, formatting, or syntactic regularities that are difficult to identify in dense representations \citep{huben2024sparse, marks2025sparse}.

Several recent works extend SAEs to large language models at scale. For instance, recently \citet{shi2025routesparseautoencoderinterpret} proposed RouteSAE which introduces routing mechanisms that propagate sparse features across layers, improving interpretability while maintaining model performance. Open-source SAE frameworks further demonstrate that sparse latents can support causal interventions and analyses in modern transformer models \citep{lieberum2024gemmascopeopensparse}. In multilingual settings, \citet{sae_lape} and \citet{deng-etal-2025-unveiling} show that SAE features can align with semantic concepts across languages, motivating the use of sparse representations for cross-lingual interpretability.

Our work leverages this progress but reframes the goal. We first identify language-associated sparse features as well as raw model neurons, by using \textsc{SAE-LAPE} \citep{sae_lape} and LAPE \citep{lape} respectively. We then systematically analyze their sensitivity to script, word order, and typological structure. Unlike prior studies that focus primarily on semantic or task-level concepts, we center our analysis on linguistic abstraction, explicitly separating representational alignment (as revealed by probing) from functional necessity (as tested via causal intervention), echoing critiques of probing as a standalone interpretability tool \citep{hewitt-liang-2019-designing, belinkov-2022-probing}.

\subsection{Typology, Script, and Romanization Effects}

Linguistic typology has long been used to study cross-lingual similarity and transfer in multilingual models. The URIEL and \texttt{lang2vec} framework provides structured vectors encoding genealogical, geographical, phonological, and syntactic properties for various languages \citep{lang2vec}. Subsequent work shows that typological information becomes increasingly linearly accessible in deeper layers of multilingual transformers, suggesting a gradual emergence of abstraction \citep{rama-etal-2020-probing}.

Orthography and script introduce an additional, often confounding, dimension. Prior work in multilingual language identification shows that script cues dominate early decisions, and that romanized or transliterated text can significantly degrade performance when script information is not explicitly modeled \citep{jauhiainen-etal-2019-language}. In representation learning, transliteration and script normalization have been shown to alter clustering structure in multilingual embedding spaces, sometimes improving transfer but often creating mismatches between surface form and linguistic identity \citep{artetxe-etal-2020-cross, moosa-etal-2023-transliteration}.

Recent interpretability studies suggest that these effects extend to internal model mechanisms. Analyses of bilingual and multilingual models show that changing script can reroute activations through different internal pathways, even when lexical content is preserved \citep{saji-etal-2025-romanlens, trinley2025languagesdoesaya23think, muller-etal-2021-unseen, lu-etal-2025-paths}. Our work builds on these observations by systematically comparing native-script and romanized inputs under a unified neuron- and feature-identification framework, revealing that script changes induce near-complete reorganization of language-associated units. Importantly, we show that this fragmentation persists even in deeper layers where typological information is linearly decodable, indicating that abstraction and control are distributed across parallel, script-bound subspaces rather than unified into a single interlingua.

\section{Identifying Language-Associated Units with LAPE and SAE-LAPE}
\label{app:lape}

This appendix summarizes the methods used to identify language-associated units in our analysis.

\subsection{LAPE for Raw Neurons}

Language Activation Probability Entropy (LAPE) quantifies how selectively an individual neuron responds to different languages.
Given a multilingual corpus, for each neuron $j$ at layer $\ell$ and language $k$, we compute the activation probability
\[
P^{(\ell)}_{j,k} = \mathbb{E}\left[\mathbb{I}\big(a^{(\ell)}_j > 0\big)\;\middle|\;\text{language } k\right],
\]
where $a^{(\ell)}_j$ denotes the neuron activation and $\mathbb{I}(\cdot)$ is the indicator function.
The vector of activation probabilities across languages is $\ell_1$-normalized to form a distribution, and its entropy is computed as
\[
\text{LAPE}^{(\ell)}_j = - \sum_k P'^{(\ell)}_{j,k} \log P'^{(\ell)}_{j,k}.
\]
Low entropy indicates that a neuron activates predominantly for a small subset of languages.
Neurons with sufficiently low entropy and a dominant language are identified as \emph{language-associated}.

\subsection{SAE-LAPE for Sparse Features}

SAE-LAPE extends the LAPE criterion to sparse latent features obtained from Sparse Autoencoders (SAEs).
SAEs are trained on feed-forward (MLP) activations to decompose dense representations into a sparse set of latent features.
Each SAE feature is treated analogously to a neuron: we compute its activation probability per language based on whether the feature is active for a given token.
The same entropy-based criterion is then applied to identify language-associated sparse features.

To ensure robustness, we restrict attention to features that are active for a non-trivial fraction of tokens and examples within at least one language.
This enables language association analysis at the level of sparse, interpretable features rather than individual neurons.

\subsection{Hyperparameters and Implementation Details}
\label{app:lape_hyperparams}

All LAPE and SAE-LAPE analyses share a common entropy-based framework for measuring language selectivity, differing primarily in their filtering criteria and membership assignment rules.

\paragraph{Activation Statistics.}
For both methods, activation probabilities are computed over a multilingual corpus by aggregating token-level activations within each language.
A unit (raw neuron or SAE latent) is considered \emph{active} for a token if its activation exceeds zero.
Activation probabilities are normalized across languages prior to entropy computation.

\paragraph{SAE-LAPE Hyperparameters.}
SAE-LAPE operates on sparse latent features extracted from Sparse Autoencoders trained on MLP activations.
To exclude noisy or overly idiosyncratic features, we apply two pre-selection thresholds:
(i) an \emph{example rate} of 0.98, requiring a latent to be active in at least 98\% of examples within at least one language, and
(ii) a \emph{high-frequency latent (HFL) rate} of 0.1, requiring activation on at least 10\% of tokens in that language.
Latents failing either criterion have their entropy set to infinity and are excluded from selection.

Language membership for SAE latents is determined using a relative top-$k$ criterion.
A latent $f$ is considered present in language $l$ if its activation probability satisfies
\[
P(f \mid l) \ge 0.8 \times \max_{l' \in L} P(f \mid l'),
\]
where the threshold ratio of 0.8 is fixed across all experiments.
This relative criterion allows features to be shared across a small number of languages when desired.
Depending on the configuration, we further restrict selection to latents that are either unique to a single language (\texttt{lang\_specific}) or shared by an exact number of languages (\texttt{lang\_shared}).

A methodological adaptation was required for Gemma models. Because the original SAE-LAPE implementation was designed for cardinally constrained \textit{Top-K} SAEs (as used for Llama), we introduced an additional filtering step for Gemma's \textit{JumpReLU} SAEs by restricting the analysis to the top-200 active latents by activation magnitude per token. While this introduces minor variance, the macro-level representational trends remain highly consistent across both SAE architectures.

\paragraph{LAPE Hyperparameters for Raw Neurons.}
For raw model neurons, which are typically denser and more polysemantic, we adopt a more conservative, percentile-based filtering strategy.
We compute the 95th percentile of activation probabilities across all neurons and languages, and discard neurons whose activation probability never exceeds this threshold in any language.
Among the remaining candidates, we select the lowest-entropy neurons corresponding to the top 1\% most language-selective units.

Language assignment for these neurons uses an absolute activation criterion: a neuron is attributed to language $l$ if its activation probability exceeds the same 95th-percentile threshold.
This approach emphasizes globally salient, language-skewed neurons rather than fine-grained feature sharing. All models share the same setup.

\paragraph{Outputs.}
Both methods export identified units with identified languages(s), activation probabilities, and entropy values.

Overall, SAE-LAPE prioritizes consistent, interpretable sparsified features with controlled cross-lingual sharing, while LAPE for raw neurons focuses on identifying the most strongly language-skewed units in dense representations. Table~\ref{tab:sae_lape_params} and Table~\ref{tab:lape_params} summarize the thresholds and hyperparameters used for SAE-LAPE and LAPE respectively. 
\begin{table}[t]
\centering
\small
\setlength{\tabcolsep}{2pt}
\begin{tabular}{l c}
\toprule
\textbf{Parameter} & \textbf{Value} \\
\midrule
Activation indicator & Latent $z > 0$ \\

Aggregation level & Token + example \\

Minimum example rate & $0.98$ \\

Minimum HFL rate & $0.10$ \\

Top-$k$ threshold ratio & $0.80$ \\



Entropy for invalid features & $\infty$ \\

Llama Top-K SAE: $k$-value & 32 \\

Gemma JumpReLU SAE: enforced Top-K & 200 \\

\bottomrule
\end{tabular}
\caption{
Hyperparameters used for SAE-LAPE identification of language-associated sparse latent features.
Rates are computed per layer over the multilingual corpus.
}
\label{tab:sae_lape_params}
\end{table}

\begin{table}[t]
\centering
\small
\setlength{\tabcolsep}{2pt}
\begin{tabular}{l c}
\toprule
\textbf{Parameter} & \textbf{Value} \\
\midrule
Activation indicator & $a > 0$ \\

Aggregation level & Token-level \\

Activation percentile (filter rate) & $95^{\text{th}}$ percentile \\

Entropy selection fraction & Lowest 1\% neurons \\

Language assignment threshold & $95^{\text{th}}$ percentile (global) \\


Inactive neuron handling & Discarded \\

\bottomrule
\end{tabular}
\caption{
Hyperparameters used for LAPE-based identification of language-associated raw neurons.
Activation percentiles are computed globally across all neurons and languages.
}
\label{tab:lape_params}
\end{table}

\subsection{Usage in This Work}

In this paper, LAPE and SAE-LAPE are used strictly as \emph{identification tools} for selecting language-associated neurons and sparse features.
All subsequent analyses -- including romanization, shuffling, probing, and causal interventions -- are conducted on these identified units.
We do not assume that low entropy alone implies abstract linguistic control or causal importance.

\section{Script Perturbation Experiments (Romanization)}
\label{app:romanization-setup}

\subsection{Experimental Setup}
\paragraph{Datasets.}
We use the \texttt{dev} split of FLORES+, which provides sentence-aligned multilingual data across typologically diverse languages. For South Asian languages, we additionally consider the Dakshina dataset to assess the effects of context-aware romanization, noting that these corpora are not sentence-aligned and are therefore used only for supplementary analysis.

\paragraph{Language Selection.}
Our experiments cover Hindi (hi), Marathi (mr), Bengali (bn), Urdu (ur), Russian (ru), Bulgarian (bg), Japanese (ja), Chinese (zh), Korean (ko), English (en), and Spanish (es), spanning multiple writing systems and including both closely related and typologically distant language pairs.

\paragraph{Romanization Procedure and Diacritics.}
Romanized text is generated using the ICU Transliterator. For applicable languages, we construct both diacritic-preserving and ASCII-only variants by removing diacritics via Unicode normalization, enabling controlled analysis of sub-phonemic orthographic cues. The whole pipeline is run thrice: (a) with the native datasets, (b) with the diacritics-romanization datasets, (c) with the diacritics-free-romanization datasets. Then the resulting neuron sets are compared.

\paragraph{Metrics.}
Overlap between language-associated feature sets is quantified using Jaccard similarity. We additionally compute cross-language overlaps to assess whether romanization induces increased sharing with English or other Latin-script languages.

\subsection{Supplementary Romanization Analysis Across Models and Representations}
\label{app:romanization_supplementary}

This appendix extends Section~\ref{sec:romanization} by documenting the full set of romanization diagnostics across all evaluated model and representation configurations. While the main text focuses on feature identity and overlap, here we examine (i) aggregate neuron-sharing structure across all languages, and (ii) distributional effects of romanization on activation behavior of language-specific neurons.

\paragraph{Aggregate Neuron Sharing Under Orthographic Variation.}
We begin by reporting aggregate Venn diagrams computed jointly over all languages, restricted to language-specific neurons identified independently per input condition. For each configuration, we plot Venn diagrams for units shared among at most three languages, comparing native-script inputs, romanized inputs with diacritics, and romanized inputs without diacritics.
This representation captures all low-order sharing behavior, ensuring a fair balance between specificity and coverage.

\begin{figure}
    \centering
    \includegraphics[width=1\linewidth]{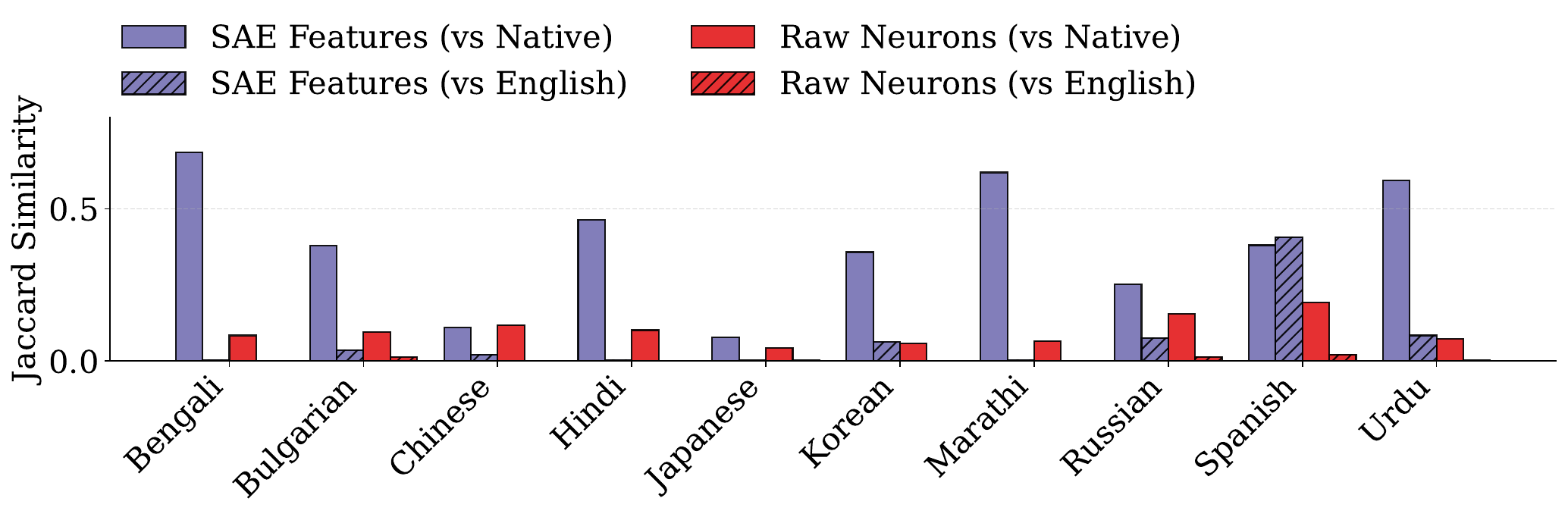}
    \caption{
    Jaccard similarity between language-associated units identified from Romanized inputs and those from Native-script or English inputs in Gemma-2-2B. Results are shown for both \textbf{\textcolor{mplred}{raw neurons}} and \textbf{\textcolor{mplpurplesae}{SAE features}}. Romanized inputs exhibit low overlap with their native-script counterparts and near-zero overlap with English in both representations, indicating limited cross-script alignment without convergence to English.}
    \label{fig:jaccard_sae_raw_native_english_gemma}
\end{figure}

\begin{figure*}[t]
    \centering

    \includegraphics[width=\linewidth]{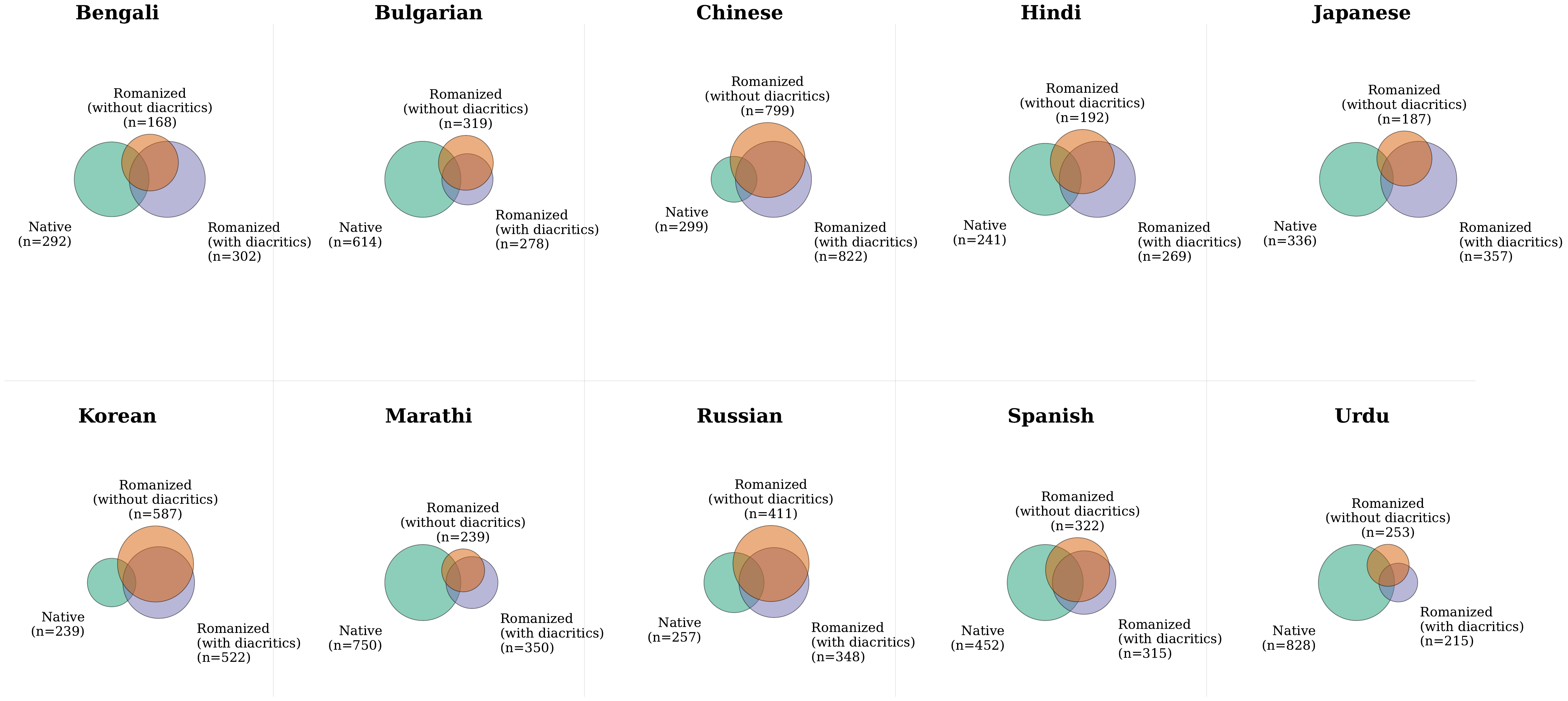}

    \vspace{3em}

    \includegraphics[width=\linewidth]{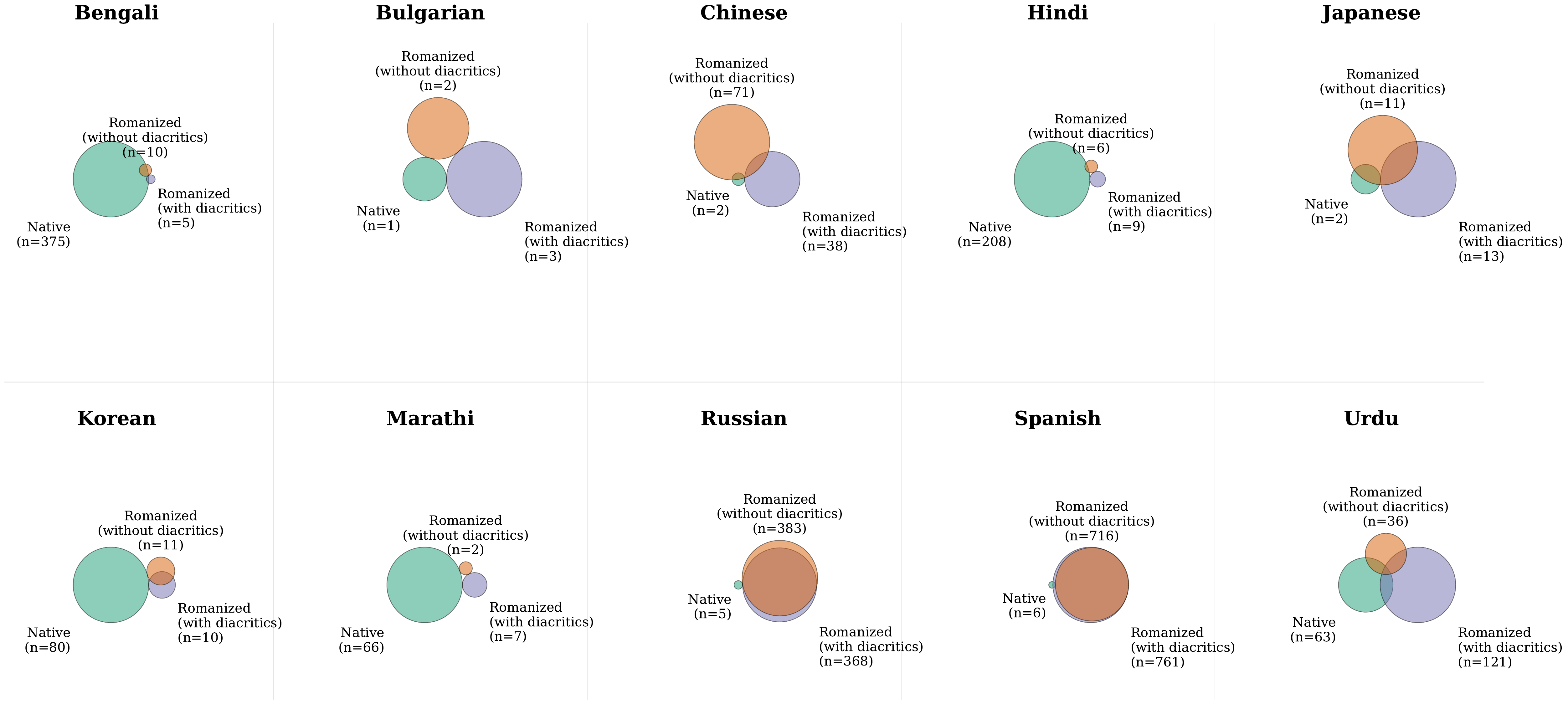}

    \caption{
    Aggregate degree-3 Venn diagrams of language-specific neurons under orthographic variation for \textbf{Gemma-2-2B}.
    Degree-3 denotes the union of neurons shared by up to three languages.
    Panels correspond to raw MLP and SAE representations under diacritics-preserving and diacritics-removed romanization.}
    \label{fig:romanization_venn_gemma}
\end{figure*}

\begin{figure*}[t]
    \centering

    \includegraphics[width=\linewidth]{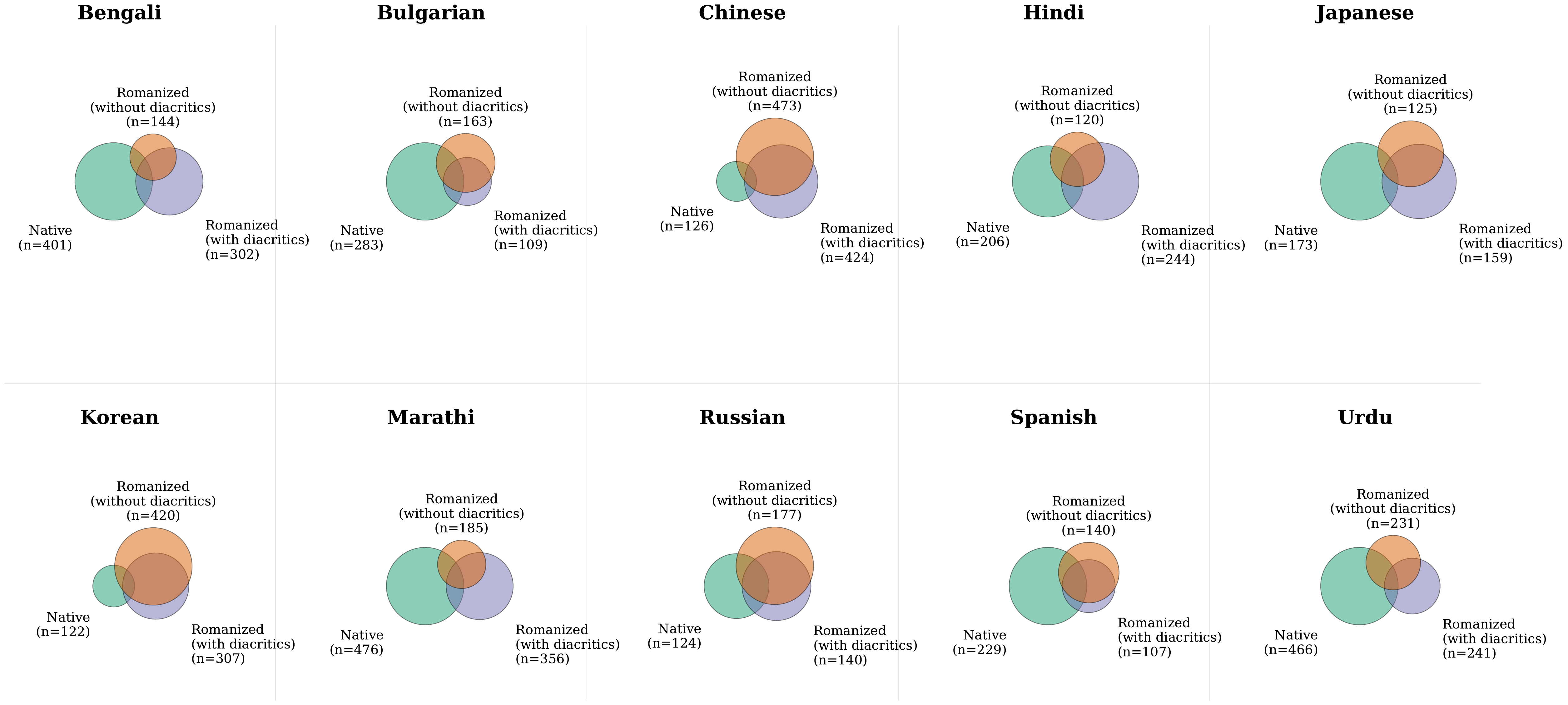}

    \vspace{3em}

    \includegraphics[width=\linewidth]{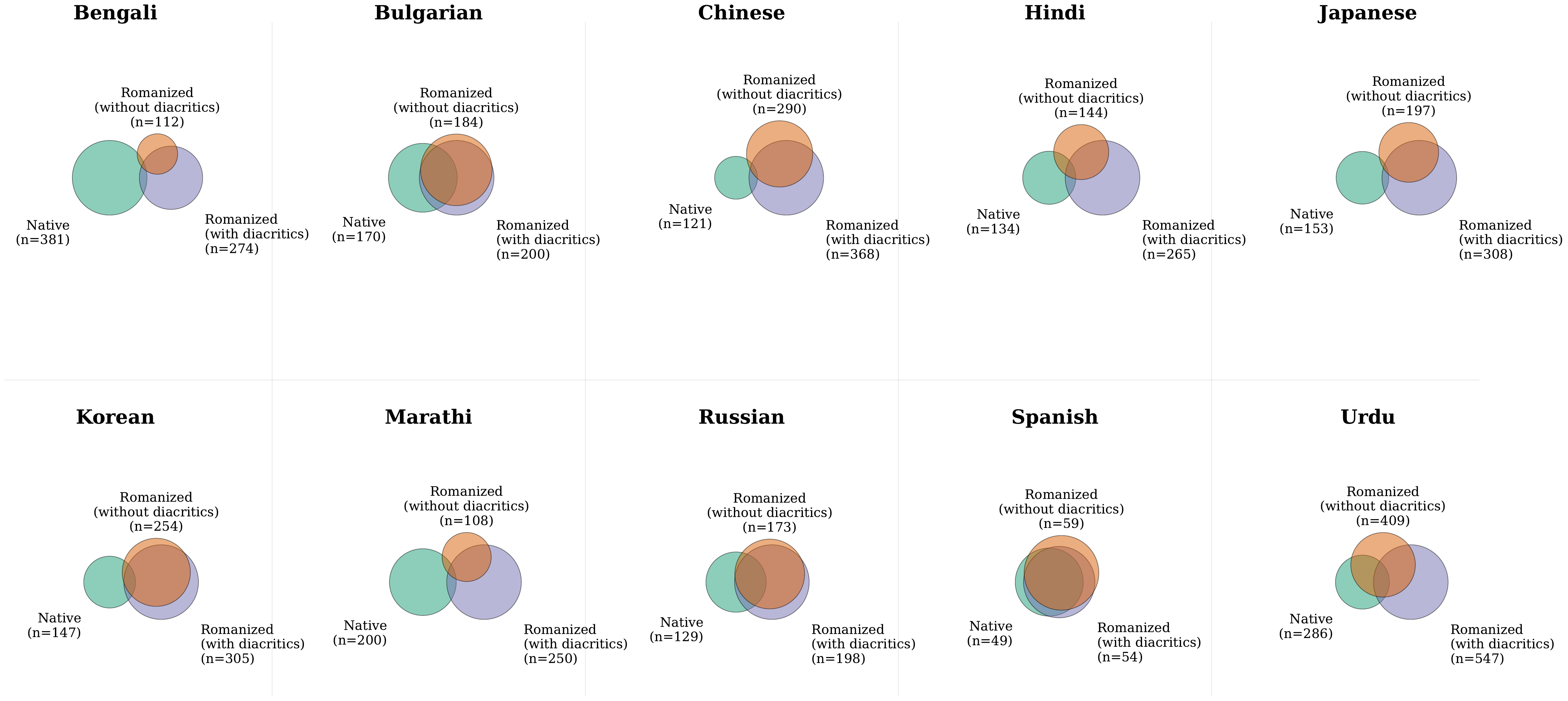}

    \caption{
    Aggregate degree-3 Venn diagrams of language-specific neurons under orthographic variation for \textbf{Llama-3.2-1B}.
    Degree-3 denotes the union of neurons shared by up to three languages.
    Panels correspond to raw MLP and SAE representations under diacritics-preserving and diacritics-removed romanization.}
    \label{fig:romanization_venn_llama}
\end{figure*}

Figures~\ref{fig:romanization_venn_gemma} and~\ref{fig:romanization_venn_llama} summarize these results across Gemma and Llama, under both raw MLP and SAE representations. Across all available configurations, aggregate overlap between native and romanized variants remains low. Overlap between the two romanized variants is slightly higher but remains limited, indicating that even minor orthographic perturbations such as diacritic removal induce substantial reassignment of language-specific neurons. 
Figure \ref{fig:jaccard_sae_raw_native_english_gemma} shows these trends for Gemma, consistent with the trends for Llama from the main text. These aggregate results confirm that the effects reported per-language in the main text persist at the multilingual level.


\paragraph{Distributional Effects of Romanization.}
Overlap-based analyses describe neuron \emph{reuse}, but do not capture how retained neurons behave. We therefore analyze activation statistics under native versus romanized inputs for both (i) the complete sets of neurons active in each condition, and (ii) the subset of neurons overlapping between native and romanized representations.

\begin{figure}[!h]
    \centering
    \includegraphics[width=0.9\linewidth]{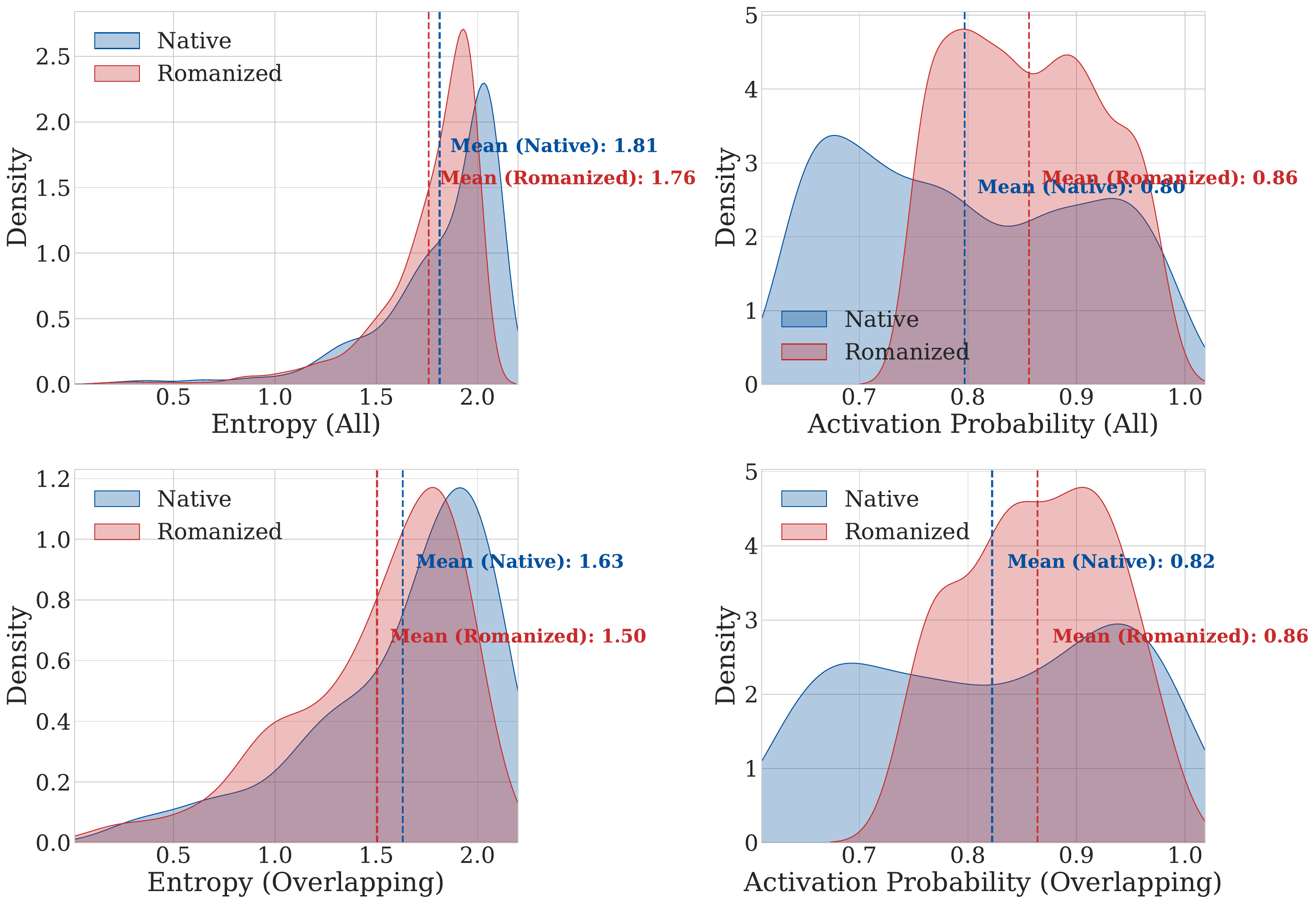}
    \vspace{0.6em}
    \includegraphics[width=0.9\linewidth]{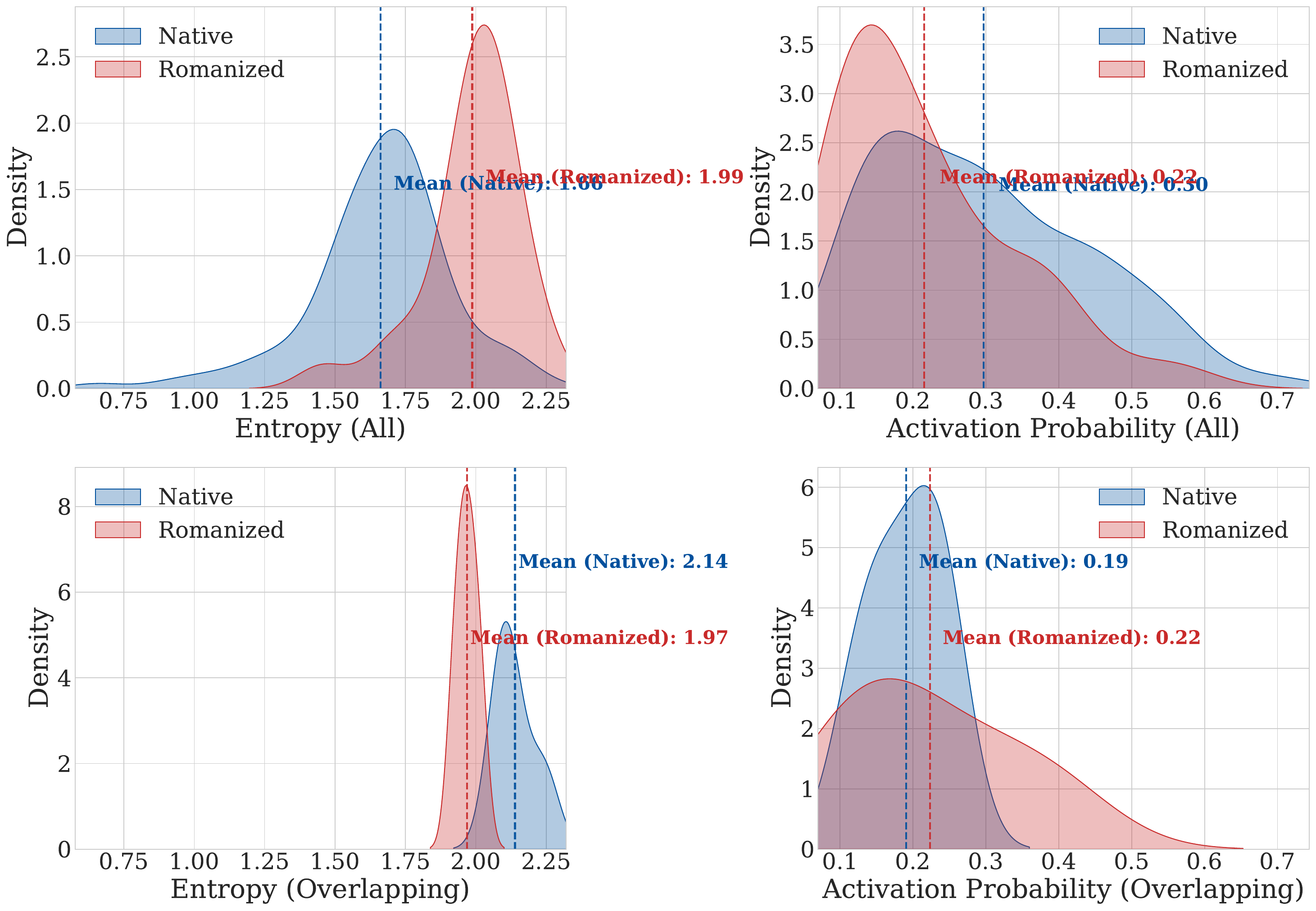}
    \caption{
    Activation probability and entropy distributions for language-specific neurons under native vs.\ romanized inputs (Gemma-2-2B).
    Top: raw MLP; Bottom: SAE.}
    \label{fig:romanization_dist_gemma}
\end{figure}

\begin{figure}[!h]
    \centering
    \includegraphics[width=0.9\linewidth]{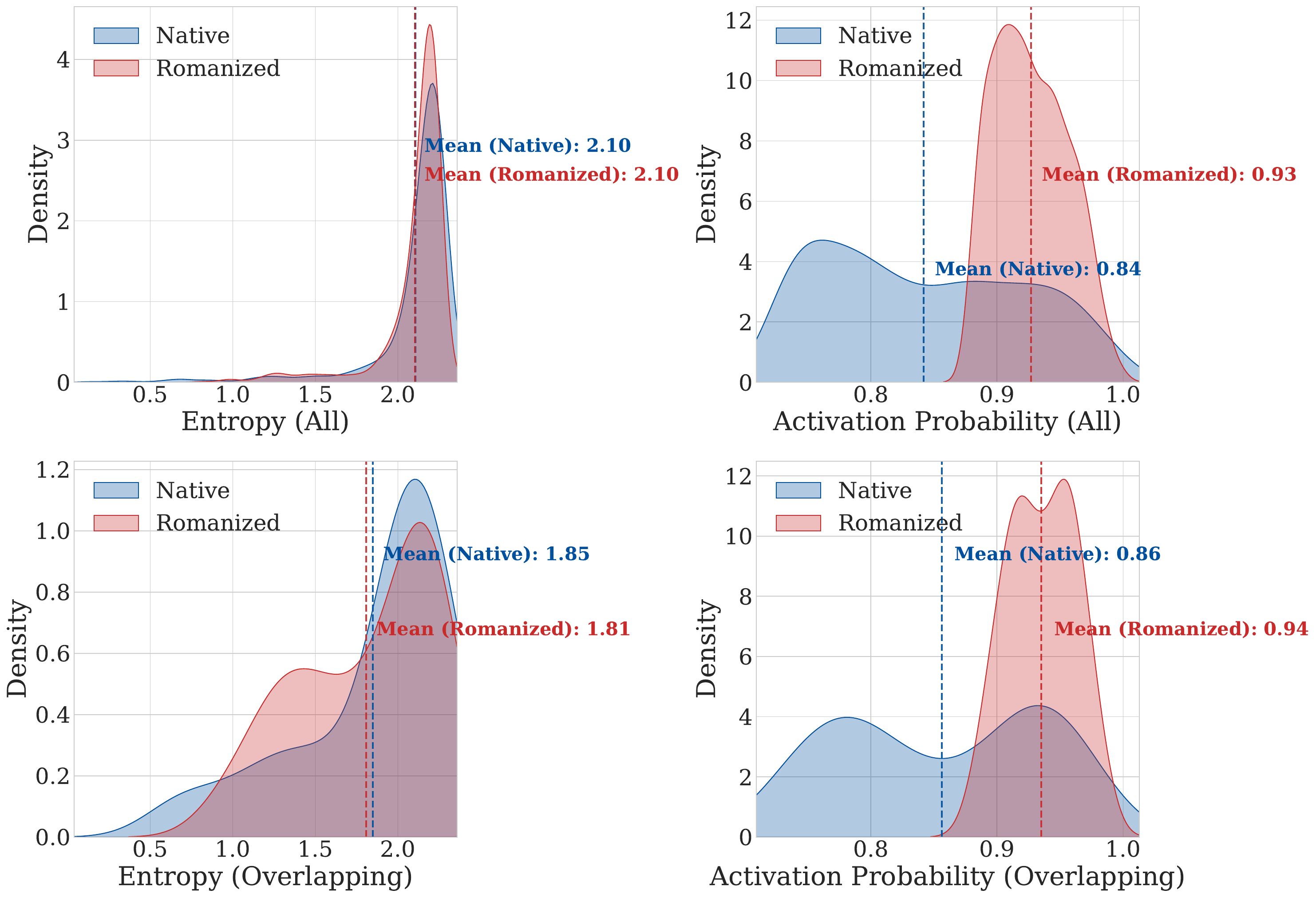}
    \vspace{0.6em}
    \includegraphics[width=0.9\linewidth]{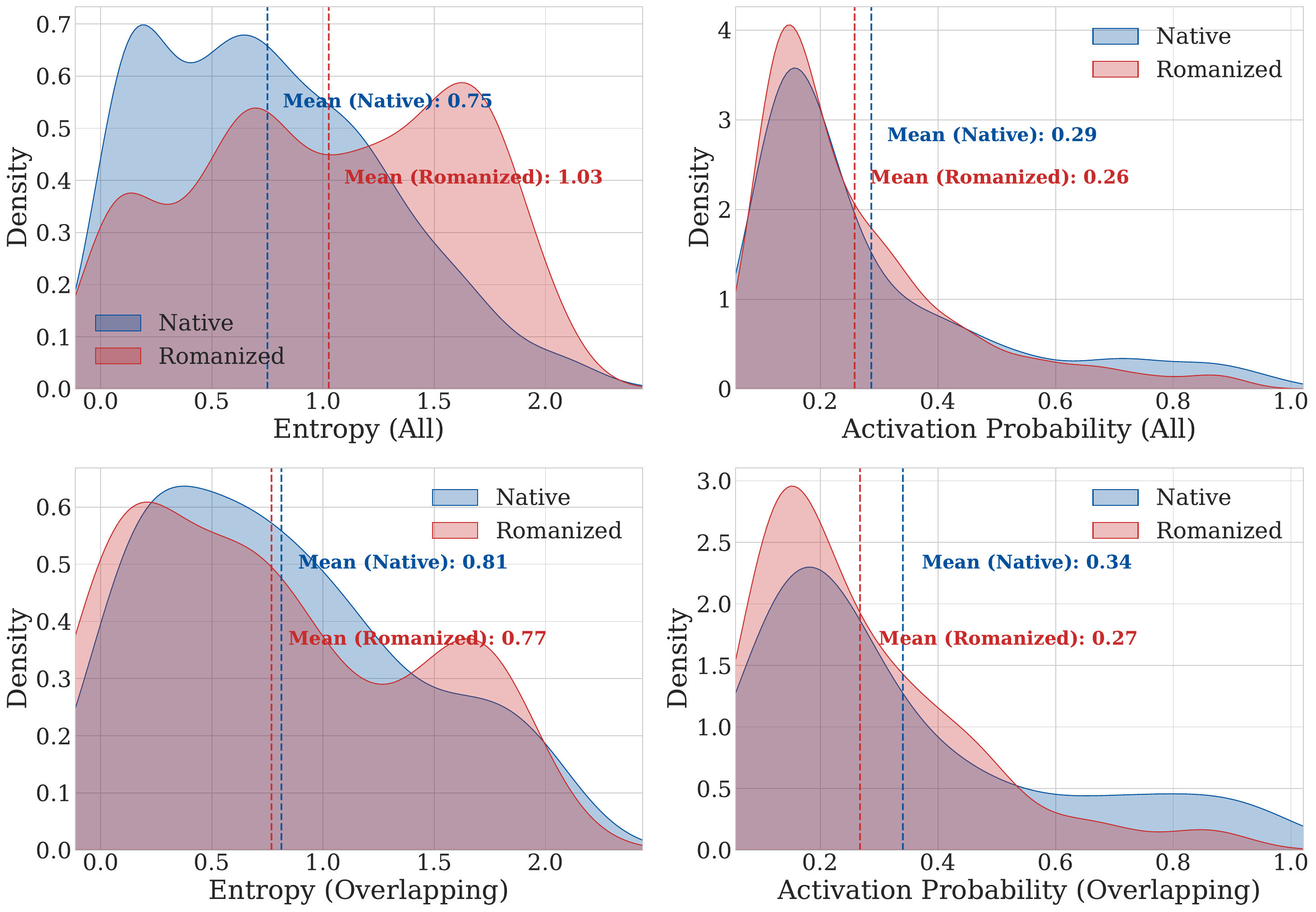}
    \caption{
    Activation probability and entropy distributions for language-specific neurons under native vs.\ romanized inputs (Llama-3.2-1B).
    Top: raw MLP; Bottom: SAE.}
    \label{fig:romanization_dist_llama}
\end{figure}

Figures~\ref{fig:romanization_dist_gemma} and~\ref{fig:romanization_dist_llama} report these distributions for Gemma and Llama across raw and SAE representations. Across all available configurations, romanization induces clear distributional shifts in both activation probability and entropy. These shifts are observed both when considering complete neuron sets and when restricting to overlapping neurons, indicating that the effects are not solely driven by changes in neuron identity. Moreover, the shifts are substantially larger than those observed under shuffling baselines, suggesting structured changes in activation dynamics rather than random variance.

\paragraph{Representation-Specific Distributional Trends.}
For raw MLP representations, romanization consistently shifts activation probability mass toward higher values while reducing entropy, indicating more concentrated and decisive neuron firing. This effect is pronounced for Gemma, whereas for Llama the entropy reduction is comparatively mild, despite similar probability shifts.

For SAE representations, distributional shifts are again substantial, but the directionality is less consistent across configurations. In particular, both entropy and activation probability may increase or decrease depending on the setup. However, the overall magnitude of these shifts is larger for Gemma than for Llama, suggesting that sparse representations in Gemma are more sensitive to orthographic perturbations.

\begin{figure*}[!h]
    \centering
    \includegraphics[width=1\linewidth]{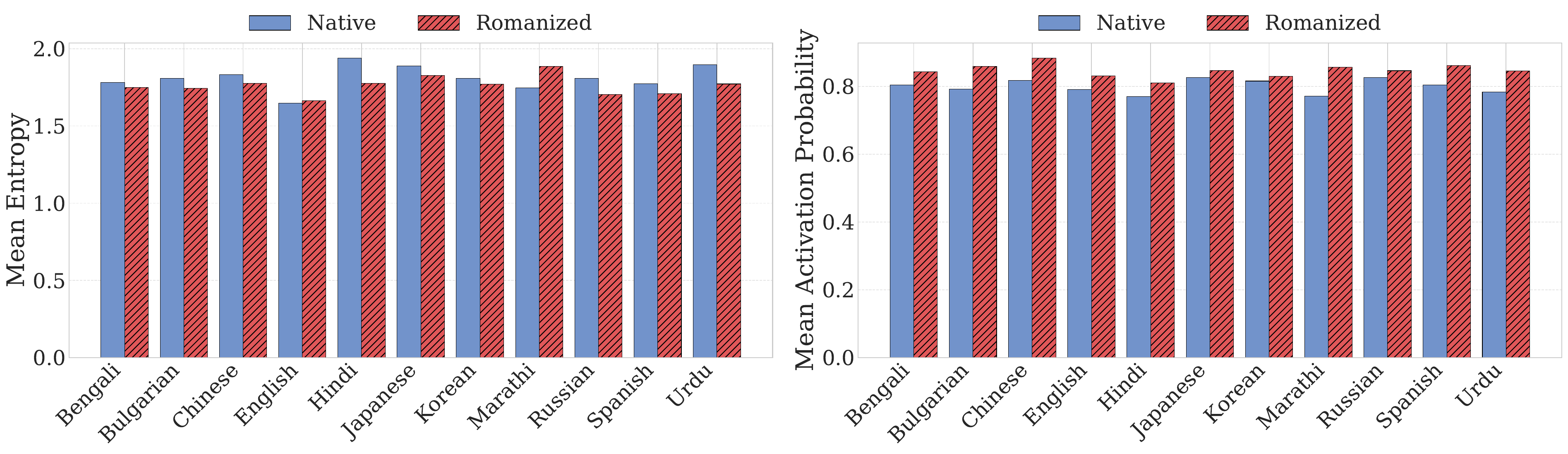}
    \vspace{0.7em}
    \includegraphics[width=1\linewidth]{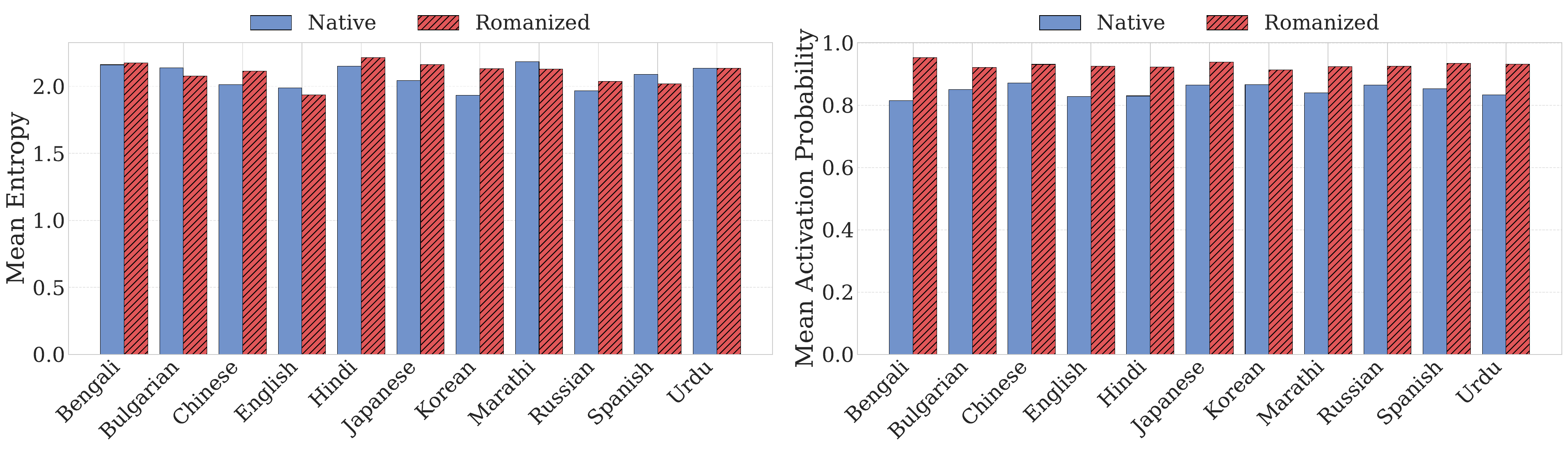}
    \caption{
    Mean activation statistics across languages for native and romanized inputs, for the raw MLP LAPE-identified features.
    Top: Gemma-2-2B; Bottom: Llama-3.2-1B.}
    \label{fig:romanization_means}
\end{figure*}

\paragraph{Stability of Mean Activation Statistics.}
Finally, we report mean activation statistics averaged across languages and neurons. Despite strong neuron-level redistribution and distributional shifts, mean activation values remain largely stable across native and romanized inputs, indicating that romanization reallocates activation mass without substantially altering global magnitude. Figure~\ref{fig:romanization_means} summarizes these values for the raw activations.

\paragraph{Summary.}
Together with Section~\ref{sec:romanization}, these results show that orthographic variation affects both the allocation and dynamics of language-specific neurons. Degree-3 analyses confirm that low-order sharing remains limited even when allowing pairwise reuse, while distributional statistics reveal structured activation shifts under romanization that are not captured by identity-based overlap alone.


\begin{figure}[!h]
    \centering
    \includegraphics[width=\columnwidth]{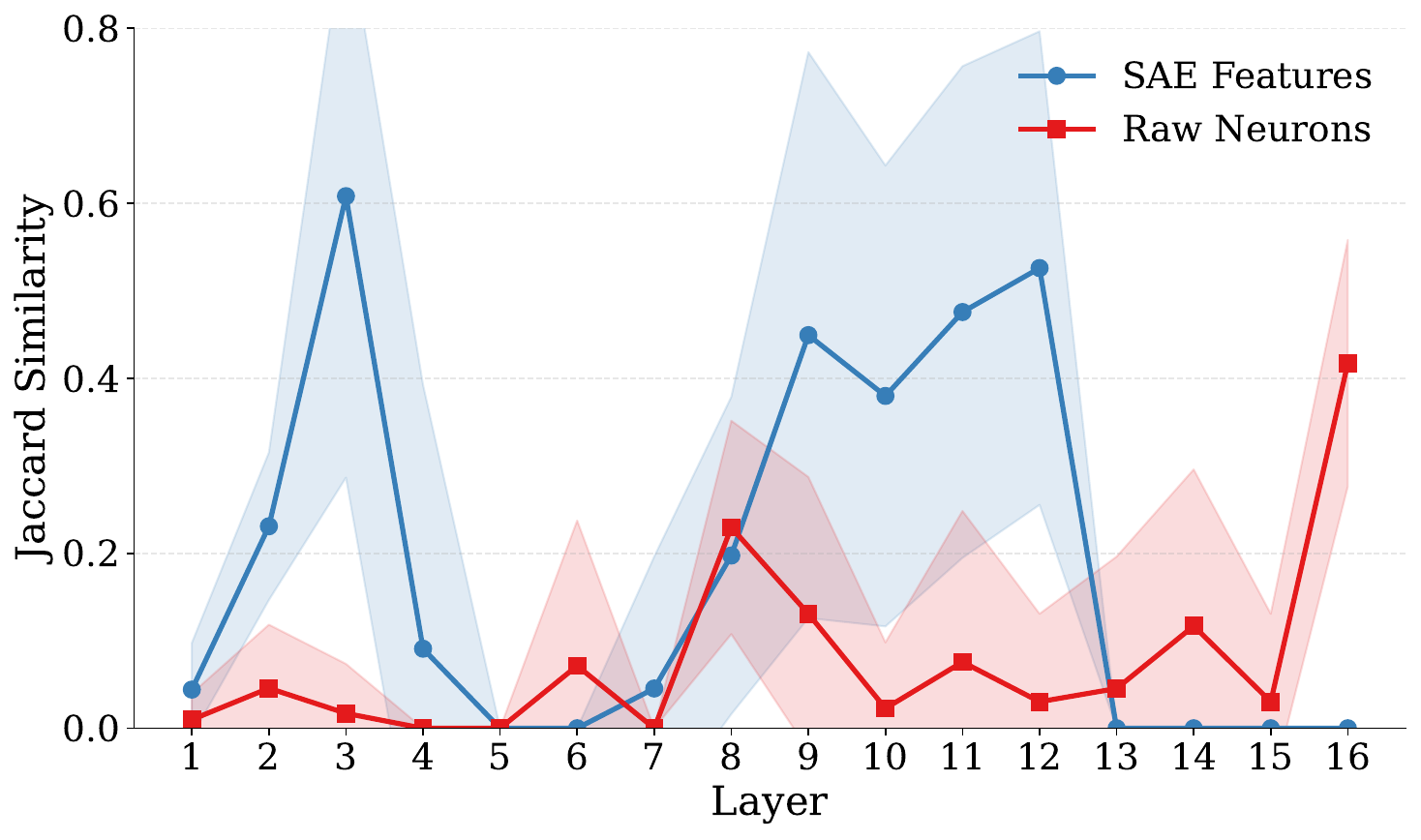}
    \caption{Layer-wise alignment between language-associated units for Native and Romanized inputs in Gemma-2-2B. The \textbf{\textcolor{mplred}{red line}} denotes average Jaccard similarity for \textbf{\textcolor{mplred}{raw neurons}}, and the \textbf{\textcolor{mplblue}{blue line}} for \textbf{\textcolor{mplblue}{SAE features}}; shaded regions indicate standard deviation across languages. Both raw neurons and SAE features show a mid-layer increase in overlap. However, in all cases, alignment remains far from convergence, indicating that representational separation persists beyond input tokenization.
}
    \label{fig:layerwise_trend_gemma}
\end{figure}

    
    
    
    


\subsection{Probing--Romanization Interaction: Typological Alignment of Neuron Subsets}
\label{app:romanization_probing_correlation}
This subsection analyzes how typological structure, as measured by \texttt{lang2vec} probing, distributes across neuron subsets induced by romanization. While earlier sections establish that romanization reorganizes language-specific features, here we ask whether this reorganization correlates with the degree to which neurons encode linguistic typology.

\paragraph{Setup.}
For each layer, model, and representation (raw MLP or SAE), neurons are partitioned into four disjoint subsets based on their activity under native and romanized inputs:
(i) \textit{native-only} neurons,
(ii) \textit{romanized-only} neurons,
(iii) \textit{overlap} neurons active under both conditions,
and (iv) a \textit{baseline} consisting of all neurons in the layer.
For each subset, we compute the average family-wise maximum probing $R^2$ score across neurons for the three typological feature families used in the final analysis: \texttt{fam}, \texttt{syntax}, and \texttt{phonology}. All plots in this section report these averages using the \texttt{specific\_mean} metric.

\paragraph{Consistency Across Models and Representations.}
Across all model and representation configurations, the qualitative behavior of these curves is remarkably consistent. Baseline probing values are generally lower than those obtained from more selective neuron subsets. An exception arises for Gemma, where neurons active only for native inputs sometimes fall below the baseline. In the Gemma raw setting, probing values are comparatively similar across subsets, indicating weaker separation between neuron groups.

\paragraph{Overlap Neurons Encode Stronger Typological Structure.}
The most robust result is that the \textit{overlap} subset consistently exhibits substantially higher probing $R^2$ scores than all other subsets. This pattern holds across all models, representations, feature families, and romanization conditions. Neurons that remain active across both native and romanized inputs are therefore not only orthography-invariant, but also more strongly aligned with linguistic typology than neurons that respond selectively to a single script variant.

\paragraph{Model- and Representation-Level Effects.}
Consistent with prior probing analyses, Gemma achieves higher absolute probing scores than Llama across all neuron subsets. Within Llama, SAE representations exhibit markedly lower $R^2$ values than raw MLP activations, often by a large margin. Crucially, however, the dominance of the overlap subset persists even in these lower-signal regimes, indicating that the relationship between orthographic stability and typological alignment is robust to overall representational strength.

\paragraph{Preservation of Typological Hierarchy.}
Across all neuron subsets and configurations, the relative ordering of feature families remains unchanged:
\[
\texttt{fam} \;>\; \texttt{syntax} \;>\; \texttt{phonology}.
\]
Romanization-induced partitioning thus modulates the \emph{magnitude} of typological alignment, but not its hierarchical structure.

\begin{figure}[!h]
    \centering
    \includegraphics[width=\linewidth]{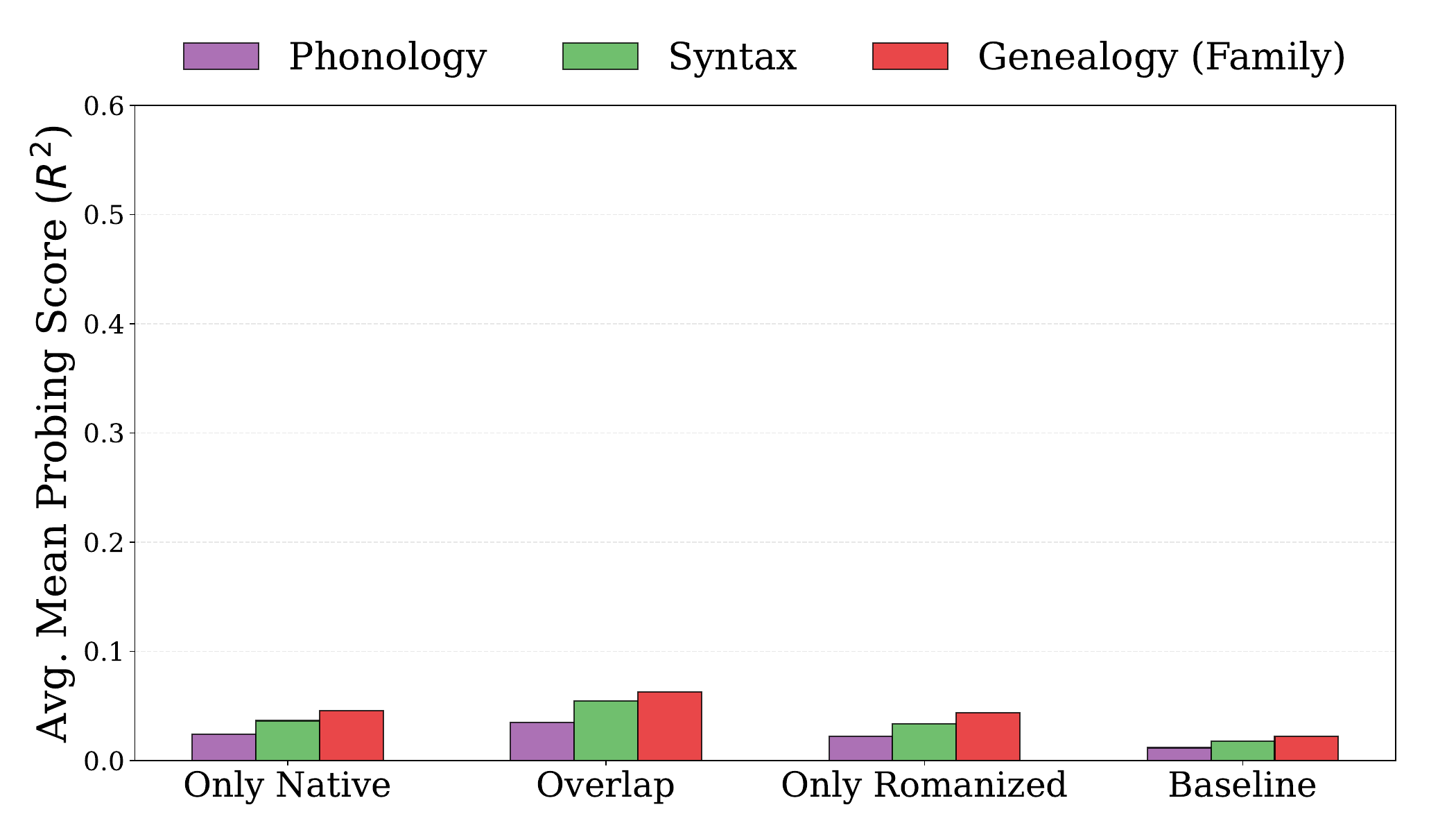}
    \caption{
    Average family-wise maximum probing $R^2$ scores across neuron subsets induced by romanization
    (Llama-3.2-1B, SAE).
    Overall probing scores are lower, but overlap neurons remain dominant.}
    \label{fig:romanization_probing_llama_sae}
\end{figure}

\begin{figure}[!h]
    \centering
    \includegraphics[width=\linewidth]{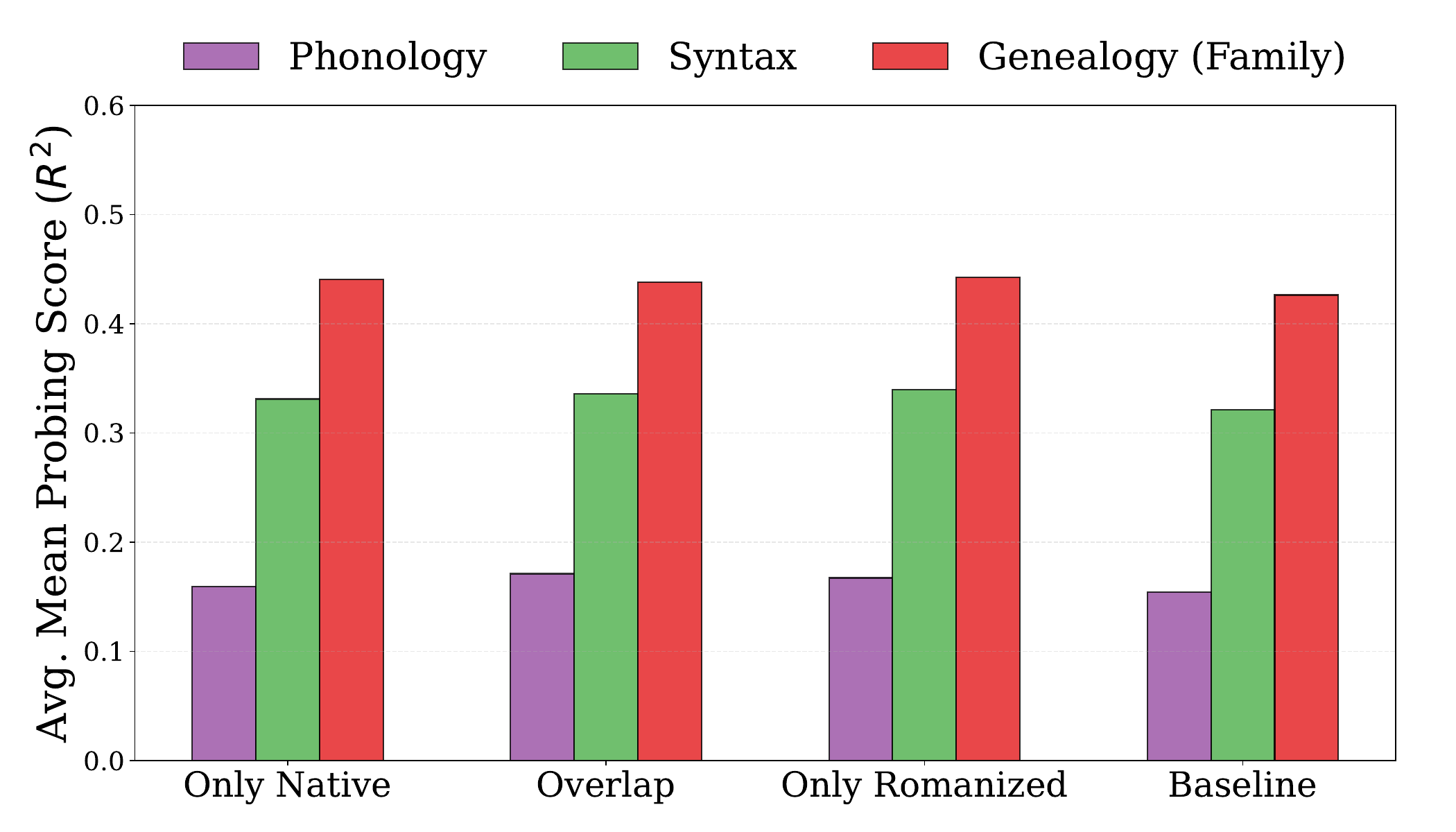}
    \caption{
    Average family-wise maximum probing $R^2$ scores across neuron subsets induced by romanization
    (Gemma-2-2B, raw MLP).
    Scores are closer across subsets, with native-only neurons occasionally falling below baseline.}
    \label{fig:romanization_probing_gemma_raw}
\end{figure}

\begin{figure}[!h]
    \centering
    \includegraphics[width=\linewidth]{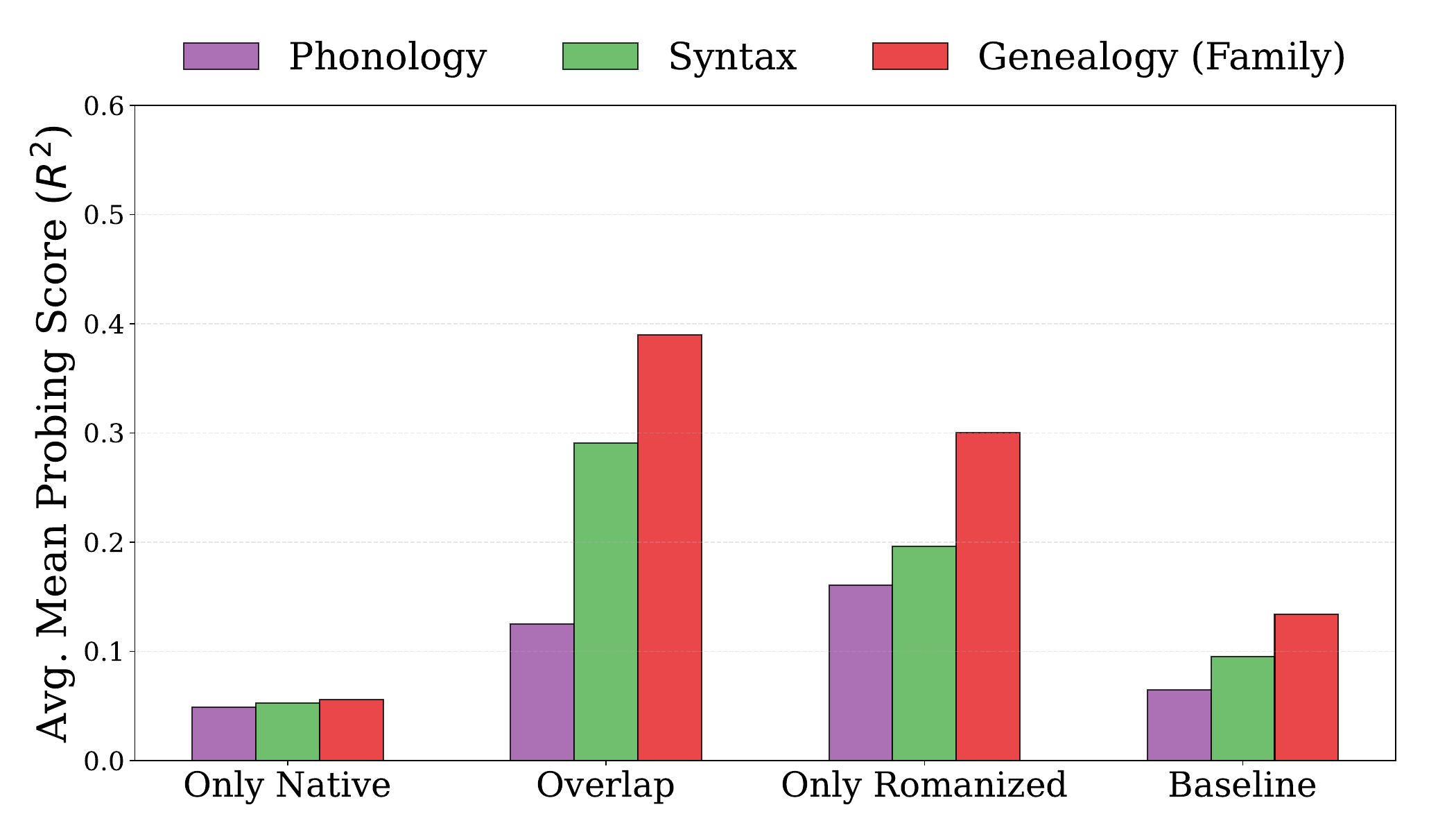}
    \caption{
    Average family-wise maximum probing $R^2$ scores across neuron subsets induced by romanization
    (Gemma-2-2B, SAE).
    Overlap neurons continue to show stronger typological alignment despite increased sparsity.}
    \label{fig:romanization_probing_gemma_sae}
\end{figure}

\paragraph{Representative Results.}
Figures~\ref{fig:romanization_probing_llama_raw}--\ref{fig:romanization_probing_gemma_sae} show representative results for Llama and Gemma under both raw and SAE representations with diacritics-preserving romanization. Analogous trends are observed for the diacritics-removed setting.

\paragraph{Summary.}
Together, these results establish a systematic association between orthographic robustness and linguistic abstraction. Neurons that are preserved across romanization transformations consistently encode stronger typological structure than neurons that are sensitive to script variation. Romanization thus serves as a diagnostic tool that reveals not only representational fragmentation, but also the locus of stable linguistic abstraction within multilingual models.

\section{Structural Perturbation Experiments (Word Shuffling)}
\label{app:shuffling}

\paragraph{Datasets.} 
Following the original setup, we use a combination of three datasets.
\begin{itemize}[nosep, wide, labelwidth=!, labelindent=0pt]
    \item[(i)] \textbf{XNLI:} 1{,}000 examples from the \texttt{train} split (en, de, fr, hi, es, th, bg, ru, tr, vi).
    \item[(ii)] \textbf{PAWS-X:} 1{,}000 examples from the \texttt{train} split (en, de, fr, es, ja, ko, zh).
    \item[(iii)] \textbf{FLORES+:} 997 examples from the \texttt{dev} split (15+ languages).
\end{itemize}

\paragraph{Procedure.}
For each dataset, we apply the LAPE and SAE-LAPE pipelines twice: (a) on sentences in their natural word order, (b) on sentences where words within each prompt are randomly permuted. All other parameters are held fixed. The resulting neuron sets are compared.

\subsection{Supplementary Shuffling Analyses Across Models and Representations}
\label{app:shuffling_supplementary}

This appendix provides additional analyses for the shuffling experiments reported in Section~\ref{sec:shuffling}. While the main text focuses on language-level stability and aggregate trends, here we document neuron-level overlap structure and distributional behavior across all model and representation configurations.

\begin{figure}[t]
    \centering
    \includegraphics[width=\linewidth]{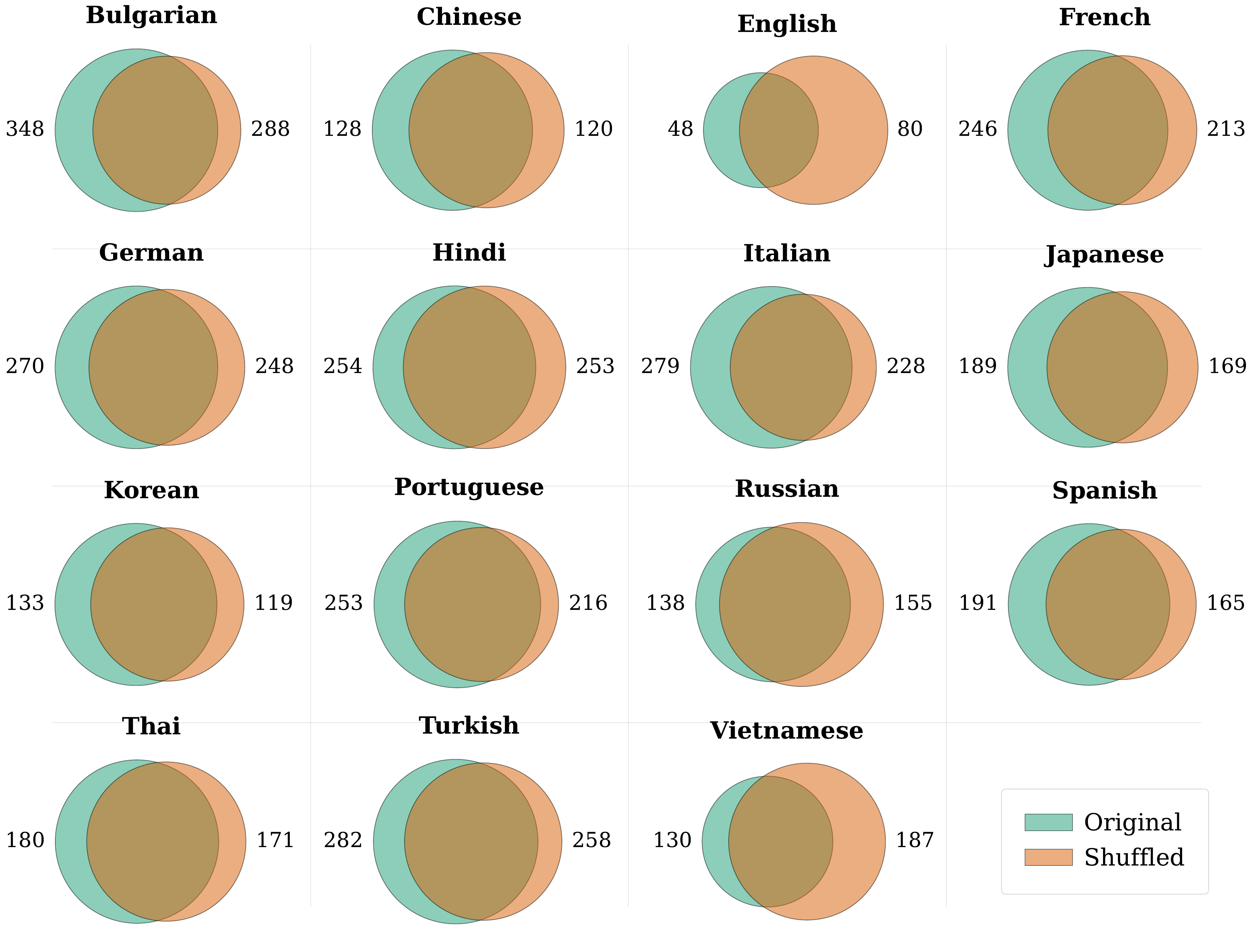}
    \vspace{0.6em}
    \includegraphics[width=\linewidth]{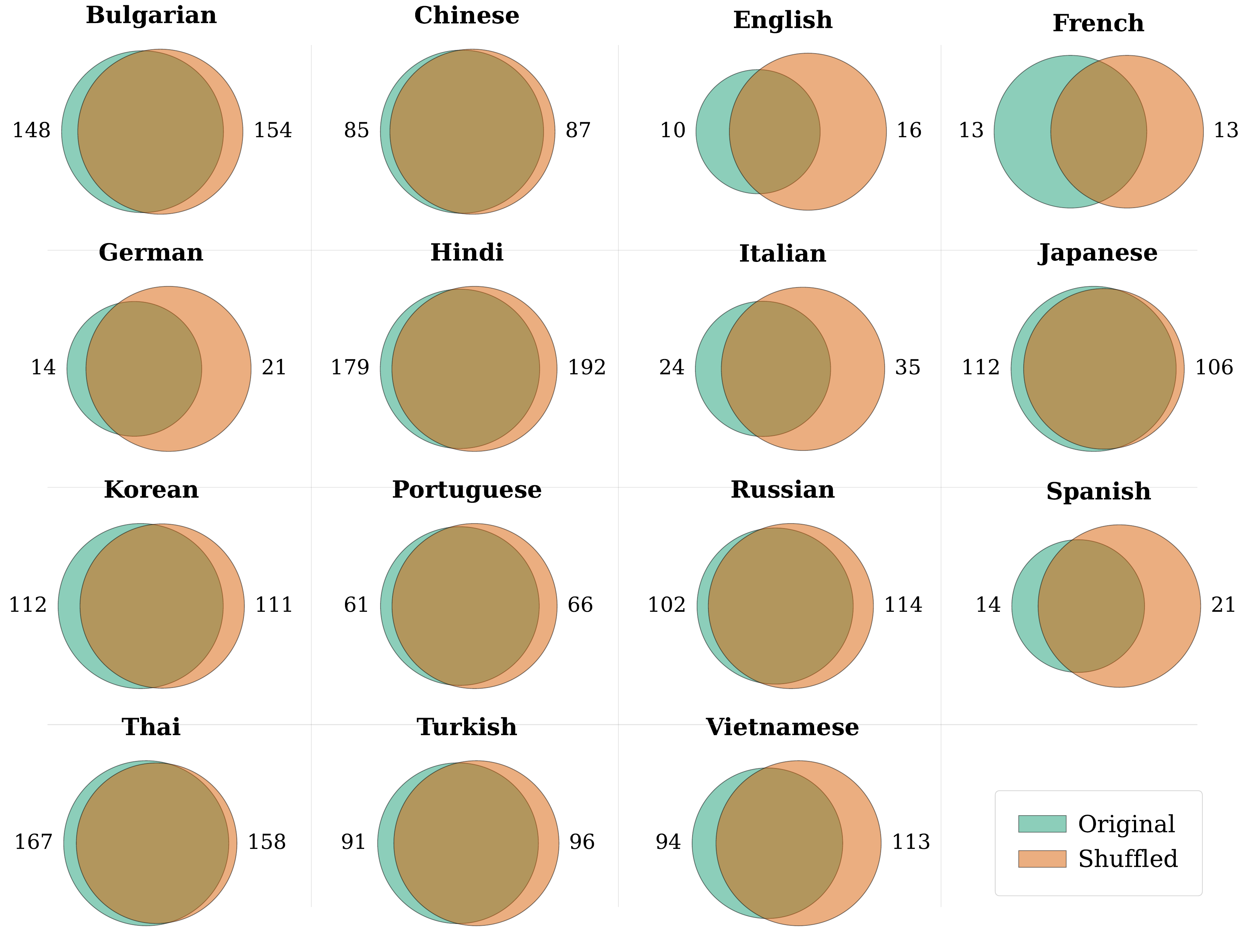}
    \caption{
    Aggregate degree-based Venn diagrams comparing features from original and shuffled inputs in Llama-3.2-1B.
    Top: raw MLP; Bottom: SAE.
    High overlap indicates stability of neuron identity under word-order perturbation.}
    \label{fig:shuffling_venn_llama}
\end{figure}

\begin{figure}[t]
    \centering
    \includegraphics[width=\linewidth]{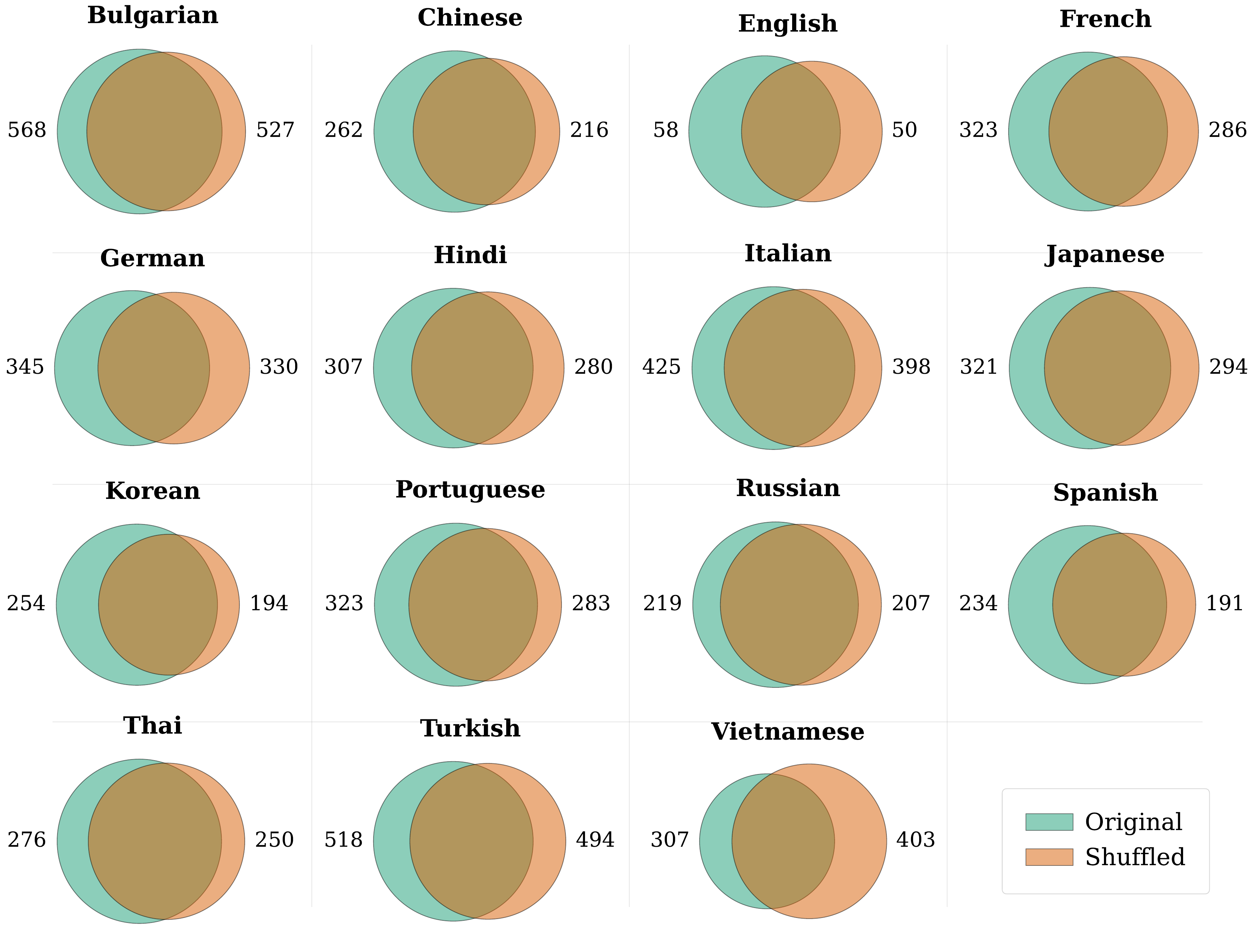}
    \vspace{0.6em}
    \includegraphics[width=\linewidth]{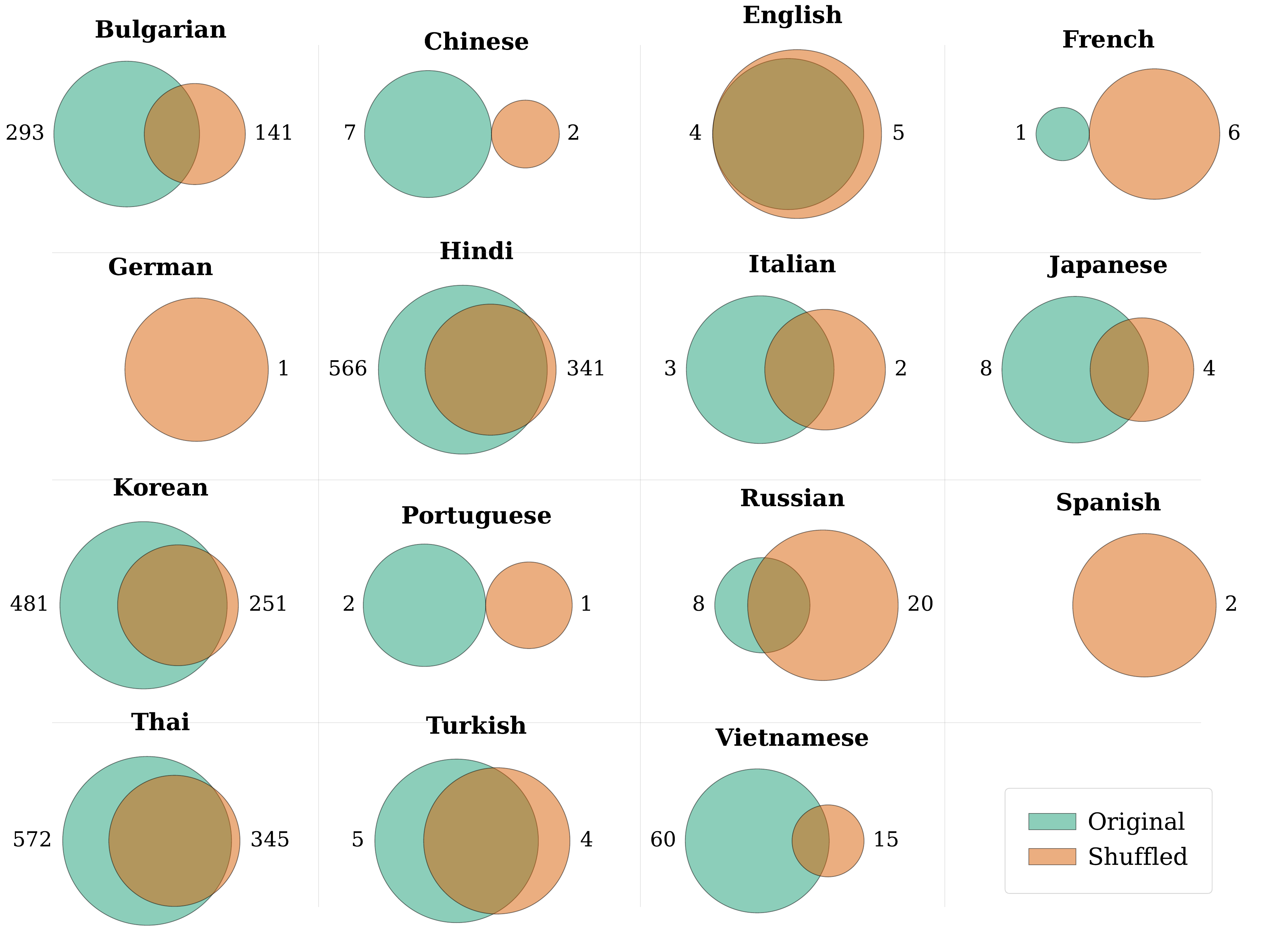}
    \caption{
    Aggregate degree-based Venn diagrams comparing features from original and shuffled inputs in Gemma-2-2B.
    Top: raw MLP (degree~3); Bottom: SAE (degree~5).
    When sufficient neurons are identified, high overlap is preserved under shuffling.}
    \label{fig:shuffling_venn_gemma}
\end{figure}

\paragraph{Aggregate Neuron Overlap Under Shuffling.}
We first examine neuron overlap between features identified from original and word-shuffled inputs, aggregated across all languages. Figures~\ref{fig:shuffling_venn_llama} and~\ref{fig:shuffling_venn_gemma} show degree-based Venn diagrams for Llama and Gemma, respectively, under raw and SAE representations.

Across all configurations, overlap between original and shuffled feature sets remains high, indicating that shuffling preserves feature identity at the neuron level. This confirms that the stability observed at the language level in the main text also holds when aggregating across neurons. 

For Gemma SAE, the absolute number of identified neurons is small for certain languages, making low-degree overlap estimates unstable. In this case, we report overlap up to degree~5 rather than degree~3. When restricting attention to settings with sufficient numbers of identified neurons, high overlap is consistently recovered, in line with other configurations.

\begin{figure}[t]
    \centering
    \includegraphics[width=\linewidth]{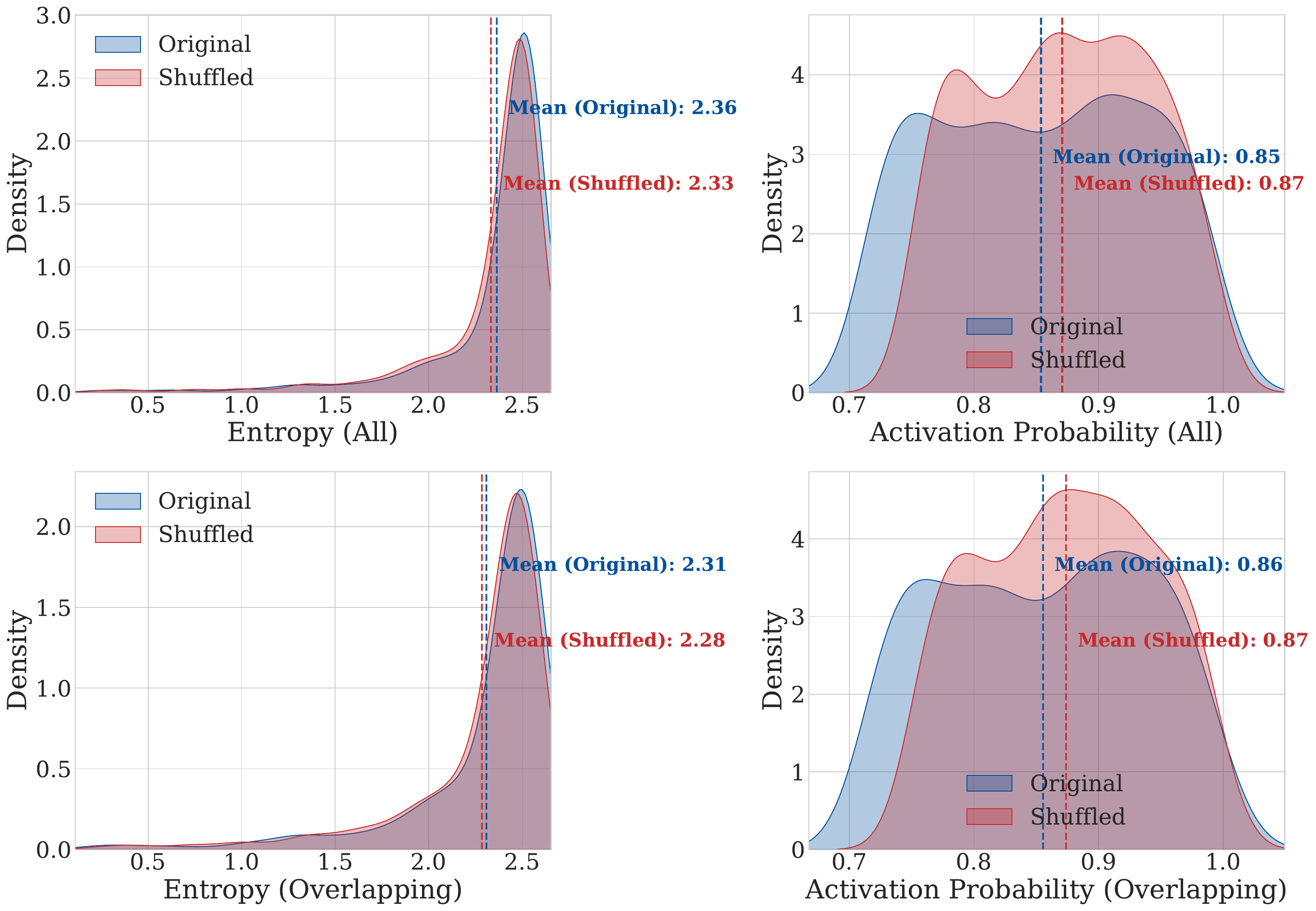}
    \vspace{0.6em}
    \includegraphics[width=\linewidth]{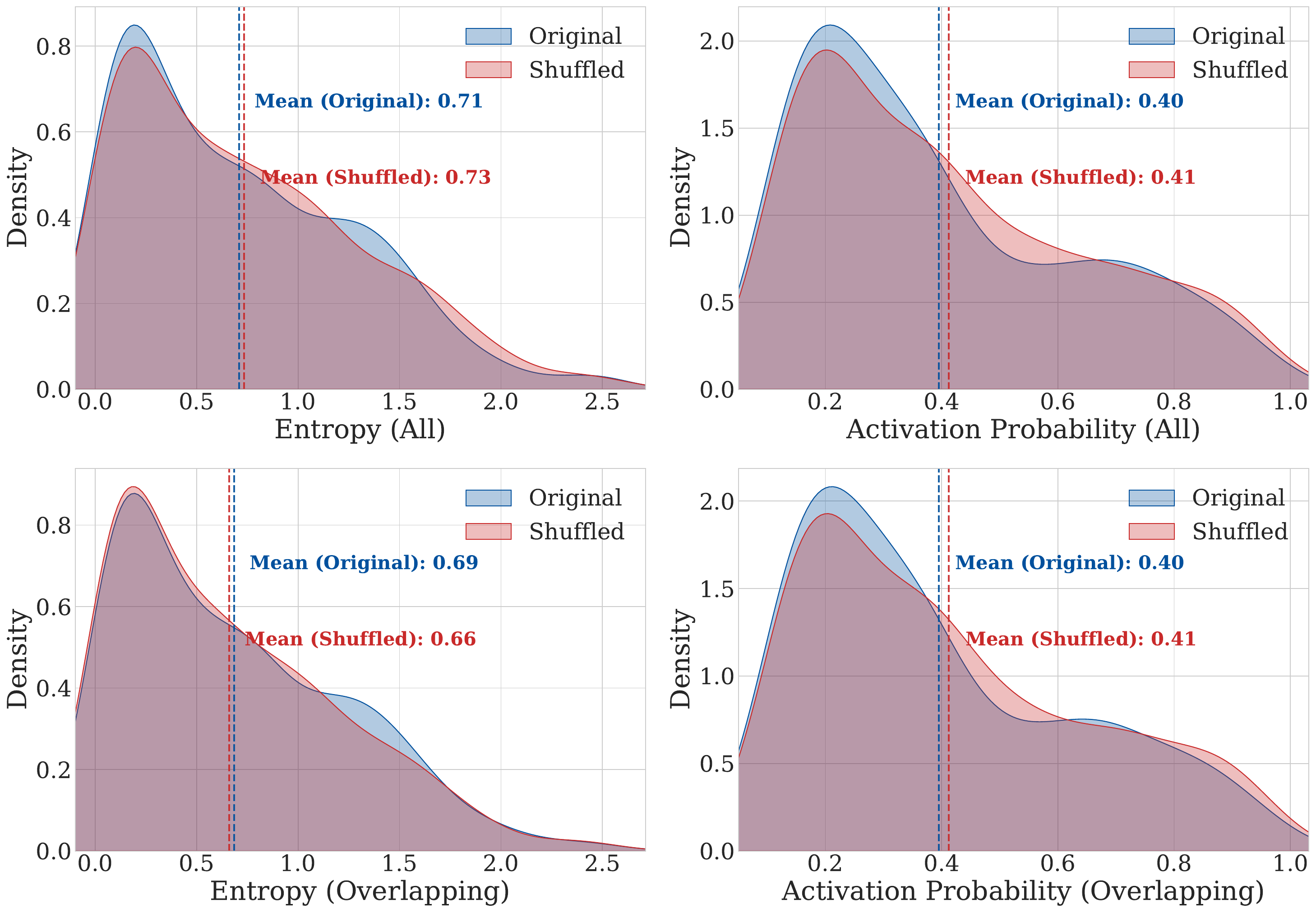}
    \caption{
    Activation entropy and selection probability distributions for original and shuffled inputs in Llama-3.2-1B.
    Top: raw MLP; Bottom: SAE.
    The near-identical distributions indicate minimal distributional shift under shuffling.}
    \label{fig:shuffling_dist_llama}
\end{figure}

\begin{figure}[t]
    \centering
    \includegraphics[width=\linewidth]{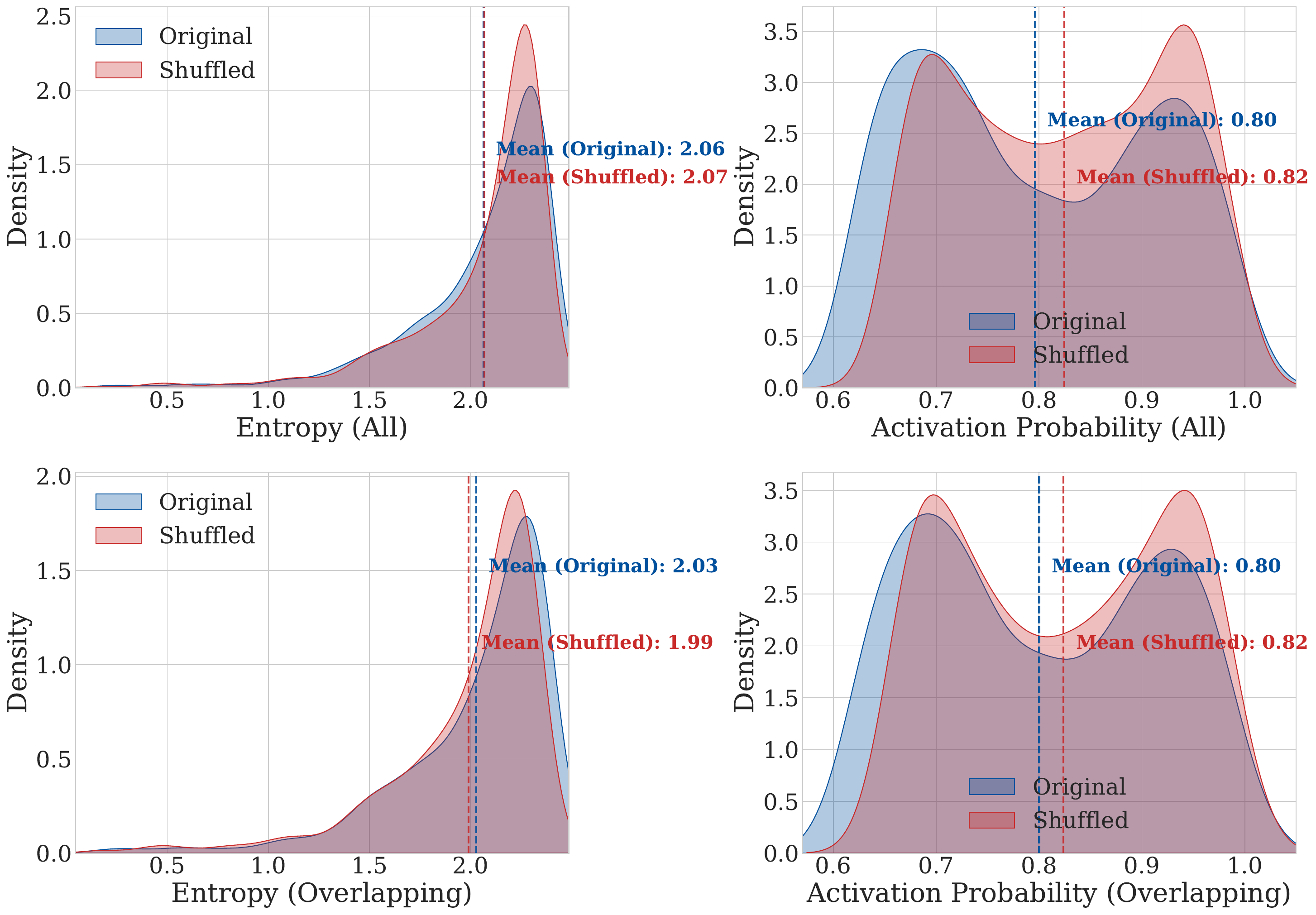}
    \vspace{0.6em}
    \includegraphics[width=\linewidth]{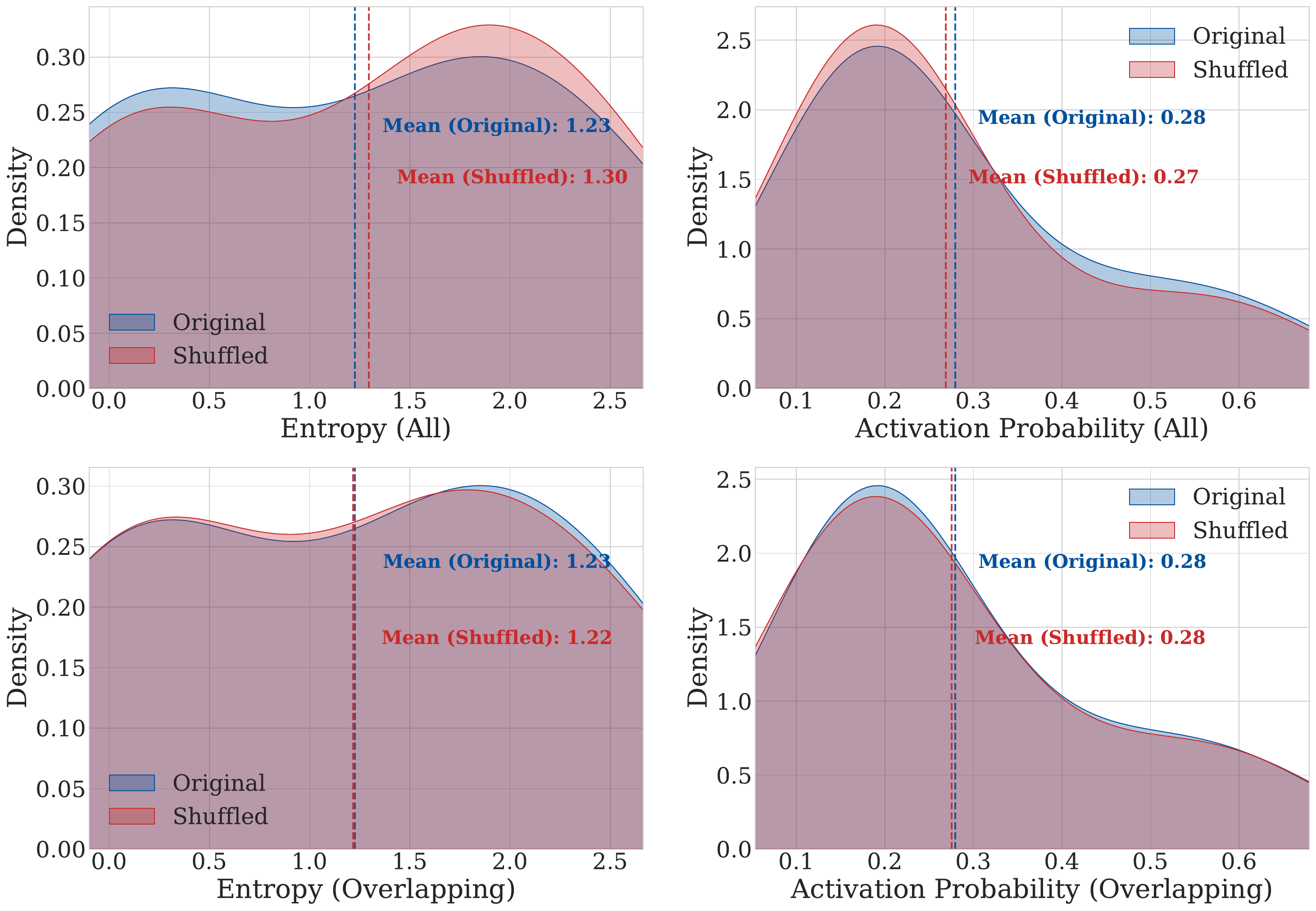}
    \caption{
    Activation entropy and selection probability distributions for original and shuffled inputs in Gemma-2-2B.
    Top: raw MLP; Bottom: SAE.
    Distributional shifts remain small across representations.}
    \label{fig:shuffling_dist_gemma}
\end{figure}

\paragraph{Distributional Stability of Activation Statistics.}
Beyond feature identity, we analyze whether shuffling induces shifts in activation behavior. Figures~\ref{fig:shuffling_dist_llama} and~\ref{fig:shuffling_dist_gemma} compare distributions of activation entropy and selection probability for original versus shuffled inputs, aggregated across languages.
Across all models and representations, the distributions are nearly overlapping, with only minor shifts in their means. This remains true when restricting the analysis to overlapping features (results omitted for brevity), indicating that neurons preserved under shuffling also maintain stable activation profiles.

\begin{figure*}[t]
    \centering
    \includegraphics[width=\linewidth]{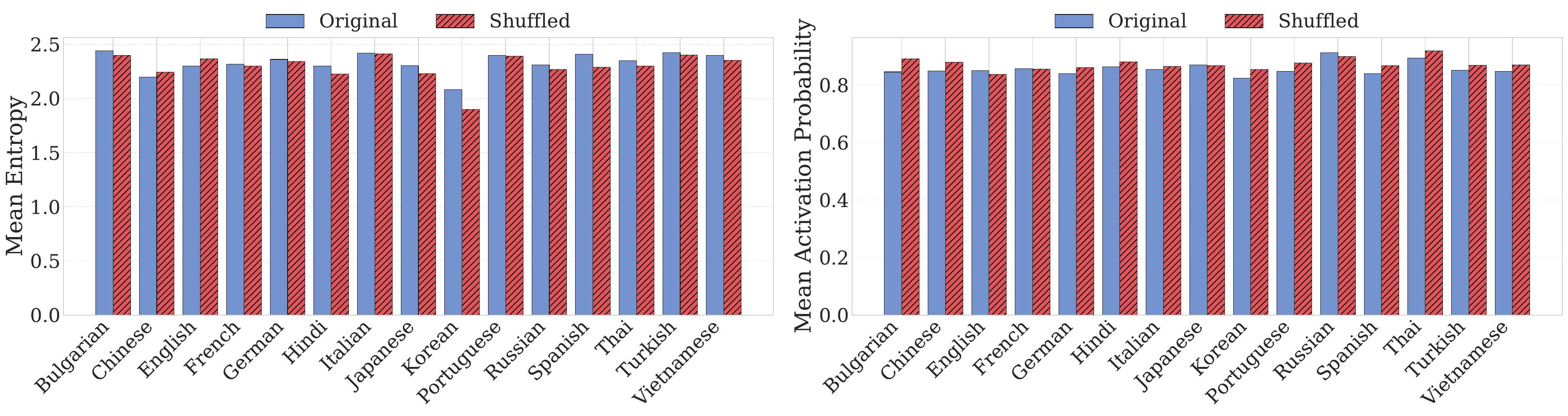}
    \vspace{0.6em}
    \includegraphics[width=\linewidth]{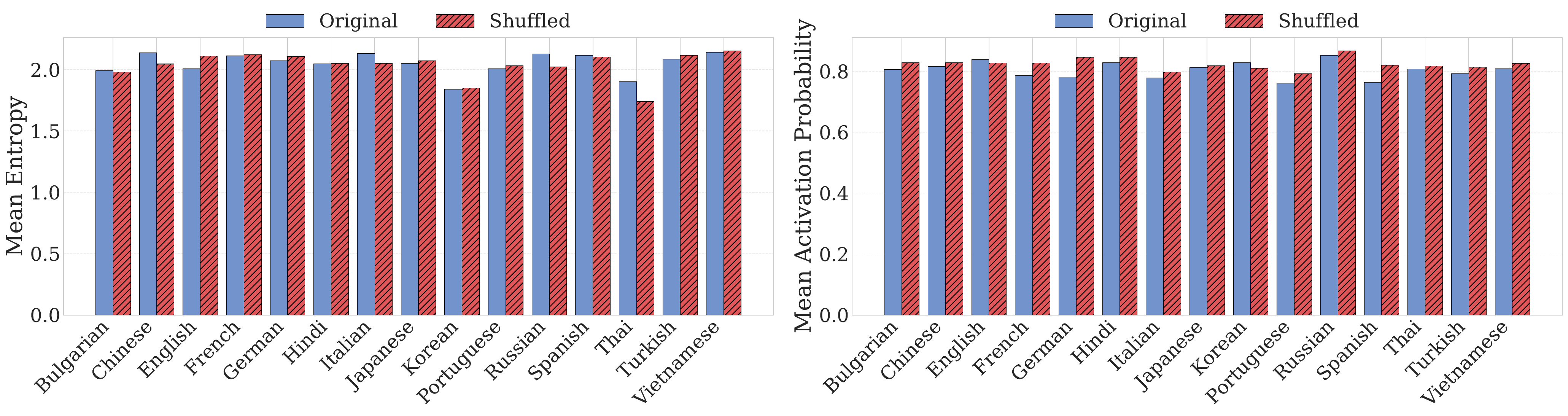}
    \caption{
    Mean activation entropy and selection probability across languages before and after shuffling.
    Top: Llama-3.2-1B; Bottom: Gemma-2-2B (raw MLP).
    Mean-level changes are small, consistent with distribution-level stability.}
    \label{fig:shuffling_means}
\end{figure*}

\paragraph{Mean Activation Statistics.}
Finally, we report mean activation statistics aggregated across languages. As shown in Figure~\ref{fig:shuffling_means}, mean entropy and selection probability change only marginally under shuffling, reiterating that syntactic perturbation does not significantly reweight feature activity.

\paragraph{Summary.}
Together, these supplementary analyses reinforce the robustness conclusions in Section~\ref{sec:shuffling}. Word-order shuffling preserves both neuron identity and activation statistics across models and representations. Differences observed in low-neuron regimes (e.g., Gemma SAE) are attributable to feature sparsity rather than systematic sensitivity to syntactic structure, further supporting the view that language-associated features primarily reflect token-level and distributional regularities.

\begin{figure}[!t]
    \centering
    \includegraphics[width=\columnwidth]{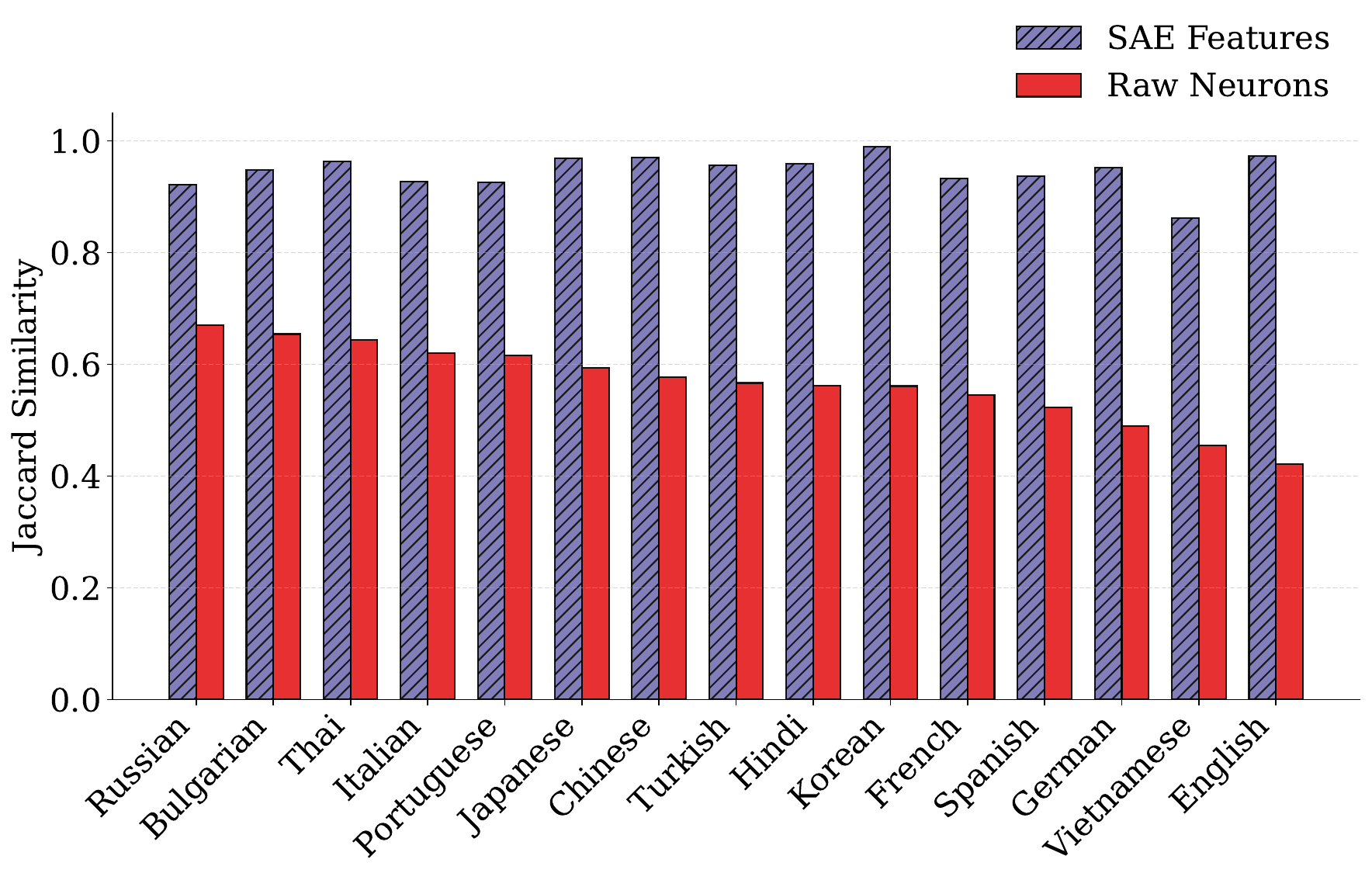}
    \caption{
Jaccard similarity between language-associated units identified from original and word-shuffled text in Gemma-2-2B.
\textbf{\textcolor{mplred}{Raw neurons}} exhibit consistently moderate-to-high overlap across languages, indicating robustness to word-order perturbation.
\textbf{\textcolor{mplpurplesae}{SAE features}} also show high overlap, revealing robustness of sparse features to local distributional patterns disrupted by shuffling.
}
    \label{fig:shuffle_jaccard_gemma}
\end{figure}

    
    
    
    

\subsection{Probing--Shuffling Interaction: Typological Alignment Under Syntactic Perturbation}
\label{app:shuffling_probing_correlation}

This subsection analyzes how sensitivity to word-order shuffling correlates with typological structure, as measured by \texttt{lang2vec} probing. In contrast to romanization, shuffling preserves surface form and token identity while disrupting local syntactic order. We therefore examine how typological alignment distributes across neuron subsets that differ in their stability under shuffling.

\paragraph{Setup.}
For each layer, model, and representation (raw MLP or SAE), neurons are partitioned into four disjoint subsets based on their activity under original and shuffled inputs:
(i) \textit{normal-only} neurons (active only for original text),
(ii) \textit{shuffled-only} neurons,
(iii) \textit{overlap} neurons active under both conditions,
and (iv) a \textit{baseline} consisting of all neurons in the layer.
For each subset, we compute the average family-wise maximum probing $R^2$ score across neurons for the three typological feature families used throughout the paper: \texttt{fam}, \texttt{syntax}, and \texttt{phonology}.
All plots report mean values aggregated across layers; we use \texttt{degree3\_mean} for all configurations, except for Gemma SAE where \texttt{degree5\_mean} is used due to low neuron counts in some languages.

\begin{figure}[t]
    \centering
    \includegraphics[width=\linewidth]{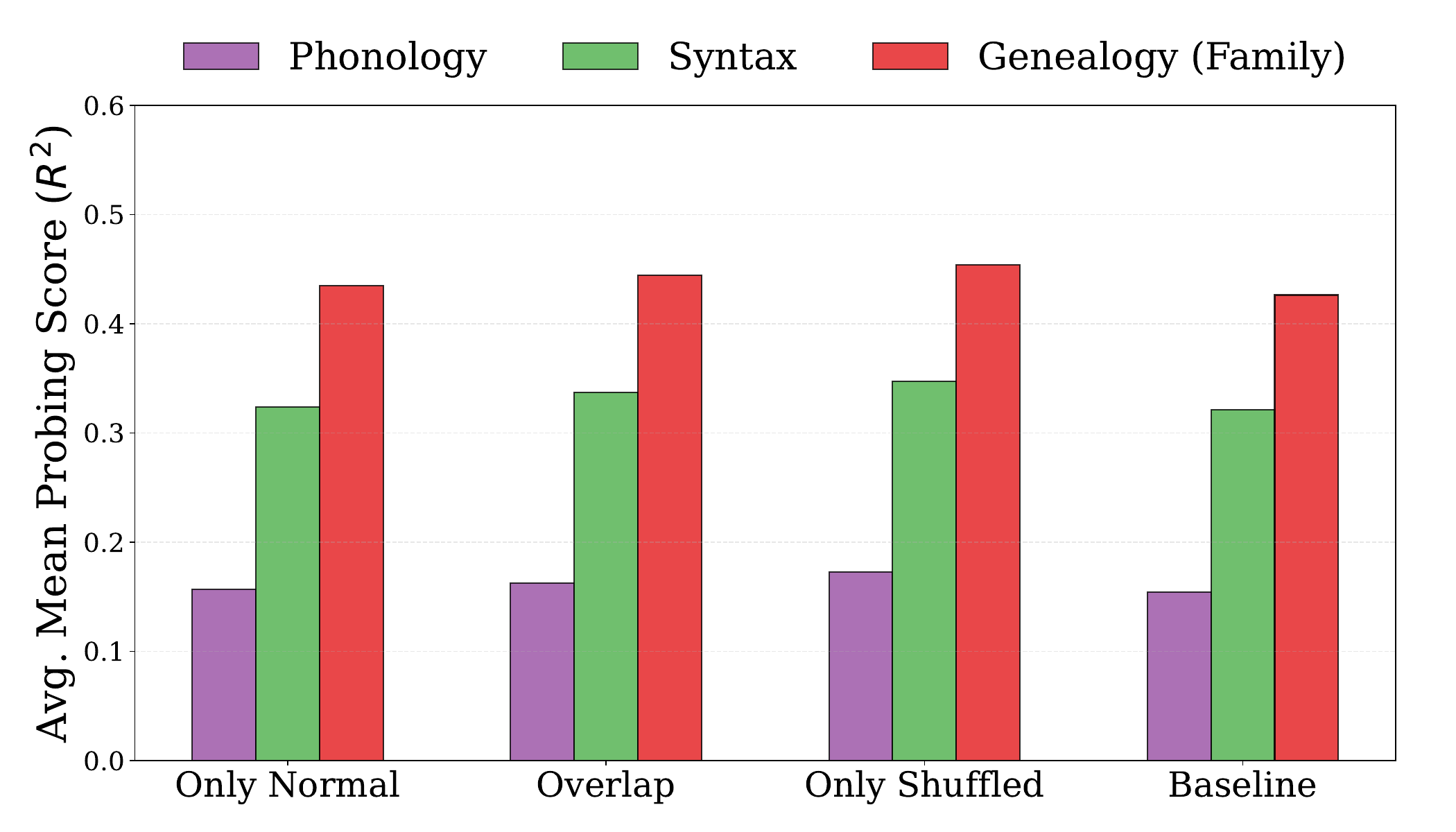}
    \caption{
    Average family-wise maximum probing $R^2$ scores across neuron subsets under shuffling
    (Gemma-2-2B, raw MLP).
    Typological alignment is similar across normal-only, shuffled-only, and overlap subsets.}
    \label{fig:shuffling_probing_gemma_raw}
\end{figure}

\paragraph{Raw Representations Show Uniform Typological Alignment.}
Figures~\ref{fig:shuffling_probing_llama_raw} and~\ref{fig:shuffling_probing_gemma_raw} show results for raw MLP representations in Llama and Gemma.
In both models, probing scores are remarkably similar across the \textit{normal-only}, \textit{shuffled-only}, and \textit{overlap} subsets.
This indicates that, at the level of distributed raw activations, sensitivity to word-order perturbation is largely decoupled from typological alignment.
Neurons that respond selectively to shuffled inputs are no less typologically informative than those that respond to original inputs.

\begin{figure}[t]
    \centering
    \includegraphics[width=\linewidth]{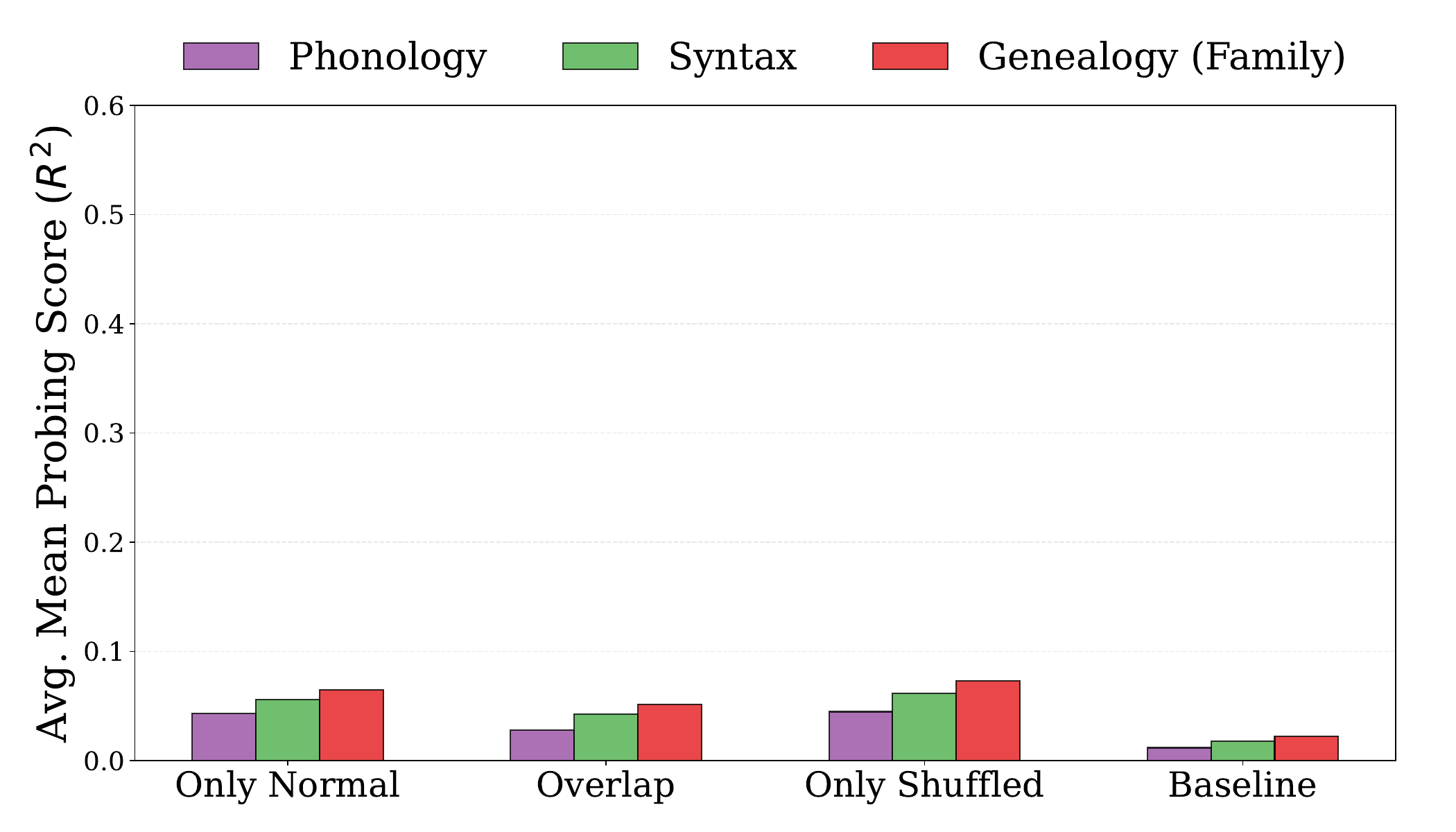}
    \caption{
    Average family-wise maximum probing $R^2$ scores across neuron subsets under shuffling
    (Llama-3.2-1B, SAE).
    Condition-specific subsets dominate overlap neurons; baseline scores remain lowest.}
    \label{fig:shuffling_probing_llama_sae}
\end{figure}

\begin{figure}[t]
    \centering
    \includegraphics[width=\linewidth]{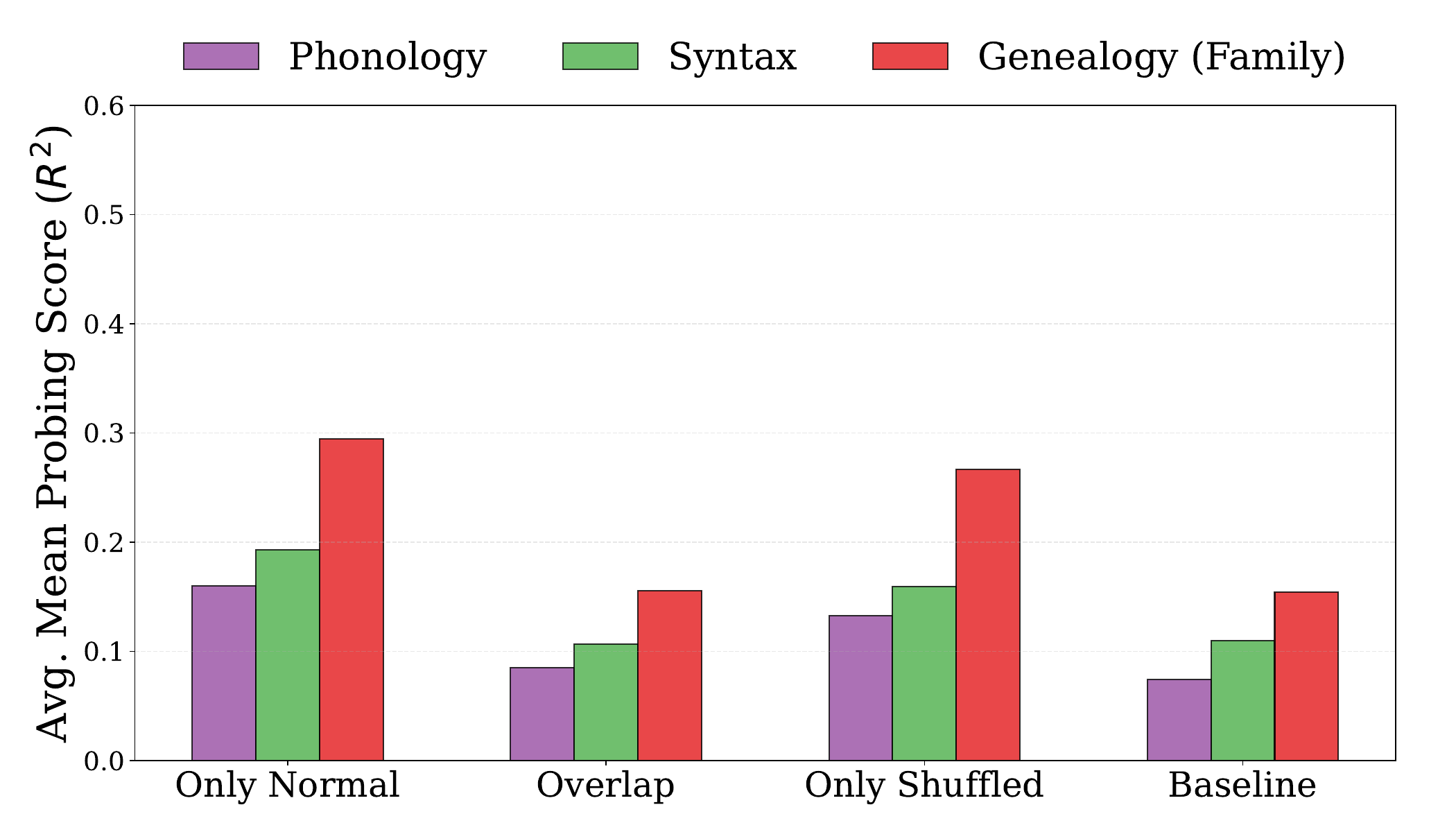}
    \caption{
    Average family-wise maximum probing $R^2$ scores across neuron subsets under shuffling
    (Gemma-2-2B, SAE).
    Results use degree-5 aggregation due to low neuron counts.
    As in Llama, condition-specific subsets show stronger typological alignment than overlap neurons.}
    \label{fig:shuffling_probing_gemma_sae}
\end{figure}

\paragraph{SAE Representations Expose a Structured Hierarchy.}
A different pattern emerges for SAE representations (Figures~\ref{fig:shuffling_probing_llama_sae} and~\ref{fig:shuffling_probing_gemma_sae}).
For both Llama and Gemma, we observe a consistent ordering:
\[
\textit{normal-only} \;\approx\; \textit{shuffled-only} \;>\; \textit{overlap}.
\]
That is, neurons selective to a single condition -- whether original or shuffled -- exhibit stronger typological alignment than neurons that remain active across both.
This contrasts sharply with the romanization setting, where overlap neurons were most informative, and suggests that invariance to word-order perturbation does not preferentially select for typologically informative features in sparse representations.

\paragraph{Baseline Effects in Llama.}
In Llama, baseline probing scores are substantially lower than those of any condition-specific subset, for both raw and SAE representations.
This gap is less pronounced in Gemma.
The result suggests that in Llama, typological information is concentrated in a relatively small subset of neurons, and is diluted when averaging across the full layer.

\paragraph{Preservation of Typological Hierarchy.}
Across all models, representations, and neuron subsets, the relative ordering of feature families remains unchanged:
\[
\texttt{fam} \;>\; \texttt{syntax} \;>\; \texttt{phonology}.
\]
Thus, while shuffling-sensitive partitioning modulates the strength of typological alignment, it does not alter the underlying hierarchy of linguistic information.

\paragraph{Representative Results.}
Figure~\ref{fig:shuffling_probing_llama_raw}, along with Figures~\ref{fig:shuffling_probing_llama_sae}--\ref{fig:shuffling_probing_gemma_sae}, show the full set of results for all configurations.


\paragraph{Summary.}
Together, these results indicate that robustness to syntactic perturbation is not a reliable indicator of typological abstraction.
In raw representations, typological information is broadly distributed and largely insensitive to shuffling-based partitioning.
In contrast, sparse representations reveal that neurons invariant to shuffling are not necessarily those most aligned with linguistic typology, highlighting a clear qualitative difference between orthographic and syntactic perturbations.

\section{Probing Typological Structure Across Layers}
\label{app:probing}

\subsection{Experimental Setup}

This section describes the probing framework used to relate neuron- and SAE-feature activations to typological properties of languages. 

\paragraph{Activation Extraction.}
For each language and layer, we extract mean activations corresponding to either raw model hidden states or SAE latents, depending on the probing condition.

Given a model layer $\ell$ and a selected set of neurons or SAE features $\mathcal{N}_\ell$, we collect activations over a multilingual dataset as follows. For each minibatch, we extract the hidden states at layer $\ell$ (or the corresponding SAE latent activations) and average over both batch and token dimensions. These per-batch means are then aggregated across batches to obtain a single activation vector per language and layer: $
\mathbf{x}_{\ell}^{(k)} \in \mathbb{R}^{|\mathcal{N}_\ell|}$, where $k$ indexes languages. Activations are collected from the FLORES+ dataset using the \texttt{train} split, with batch size 16. 

\paragraph{Typological Features.}
Typological targets are loaded from \texttt{lang2vec} features. Each feature set corresponds to a matrix $\mathbf{Y} \in \mathbb{R}^{L \times F}$, where $L$ is the number of languages and $F$ the number of typological dimensions.

Feature sets include syntactic, phonological, and inventory-based features, as well as genealogical family and geographic coordinates. Prior to probing, feature dimensions with zero variance across the selected languages are removed to ensure well-defined regression targets.

\paragraph{Regression Setup.}
Probing is formulated as a set of \emph{univariate} regression problems. For each neuron or feature $n \in \mathcal{N}_\ell$ and each typological dimension $f$, we fit a linear model across languages:
\[
y^{(k)}_f = \beta_{n,f} \, x^{(k)}_n + \epsilon^{(k)},
\]
where $x^{(k)}_n$ denotes the mean activation of neuron $n$ for language $k$.

To stabilize estimation under small sample sizes, we use ridge regression with regularization coefficient $\lambda = 1.0$. Importantly, each neuron is probed independently, i.e., regressions are single-predictor models rather than multivariate probes.

\paragraph{Cross-Validation and Evaluation.}
Probe quality is assessed using $5$-fold cross-validation over languages. In each fold, regression coefficients are estimated on the training languages and evaluated on held-out languages. The coefficient of determination ($R^2$) is computed for each neuron–feature pair on the test split.
For numerical stability and efficiency, regression is implemented in closed form and evaluated in blocks over both neuron and feature dimensions. For each neuron $n$ and feature $f$, the final probe score is obtained by averaging $R^2$ across folds:
\[
R^2_{n,f} = \frac{1}{K} \sum_{k=1}^{K} {R^2_{n,f}}^{(k)}.
\]

Neuron–feature pairs with undefined $R^2$ values (e.g., due to zero variance in the target) are excluded.

\subsection{Detailed Layerwise Probing Comparisons}
\label{sec:appendix_probing_detailed}

Here we provide a detailed layerwise analysis of probing results for the three typological feature families used in the final experiments: \texttt{fam}, \texttt{syntax}, and \texttt{phonology}. We focus on (i) differences between raw MLP activations and SAE representations, and (ii) cross-model differences between Llama-3.2-1B and Gemma-2-2B. All plots report layerwise averages of maximum $R^2$ scores per feature family.

\begin{figure}[!t]
    \centering
    \includegraphics[width=0.9\linewidth]{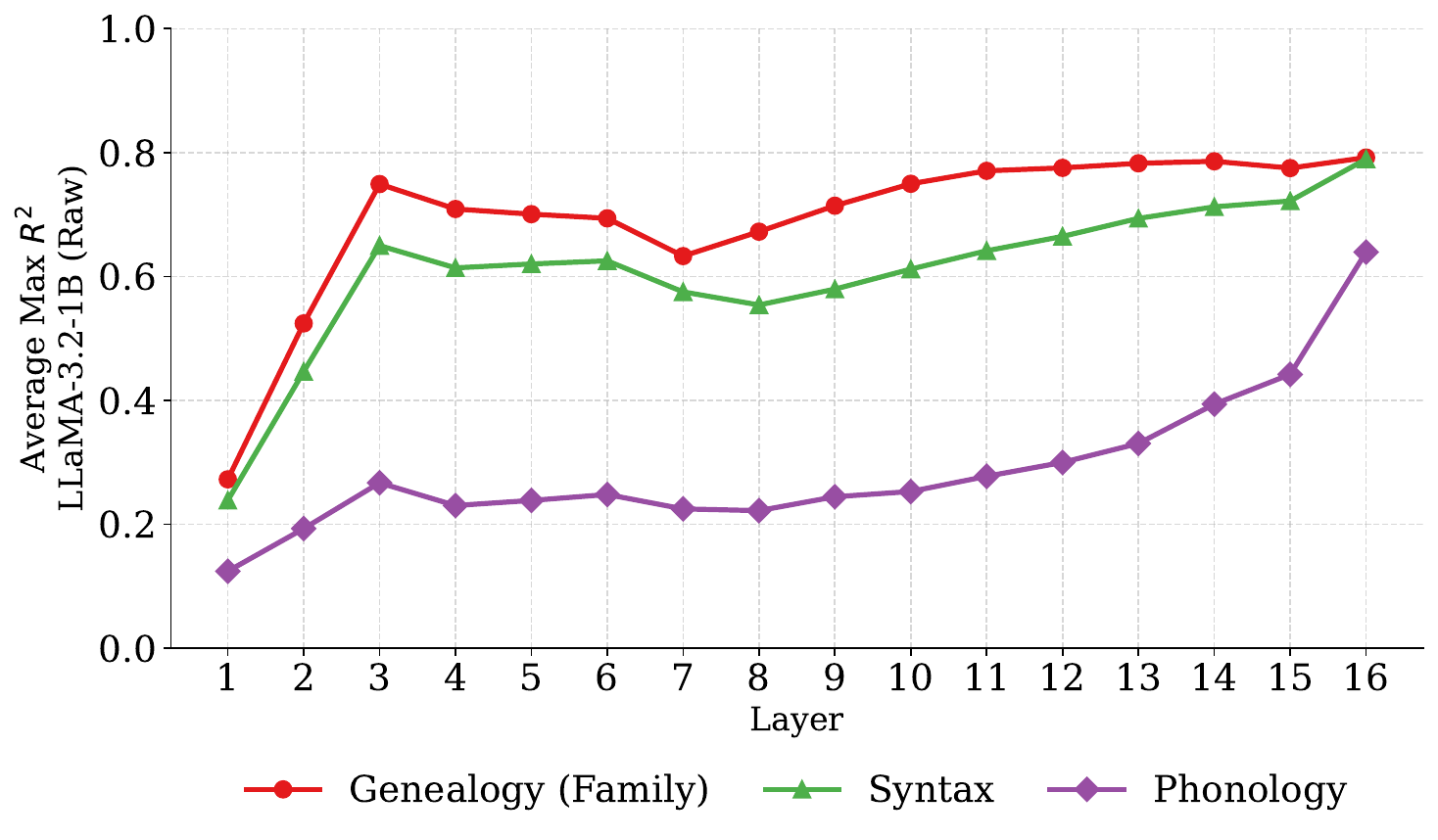}
    \vspace{0.6em}
    \includegraphics[width=0.9\linewidth]{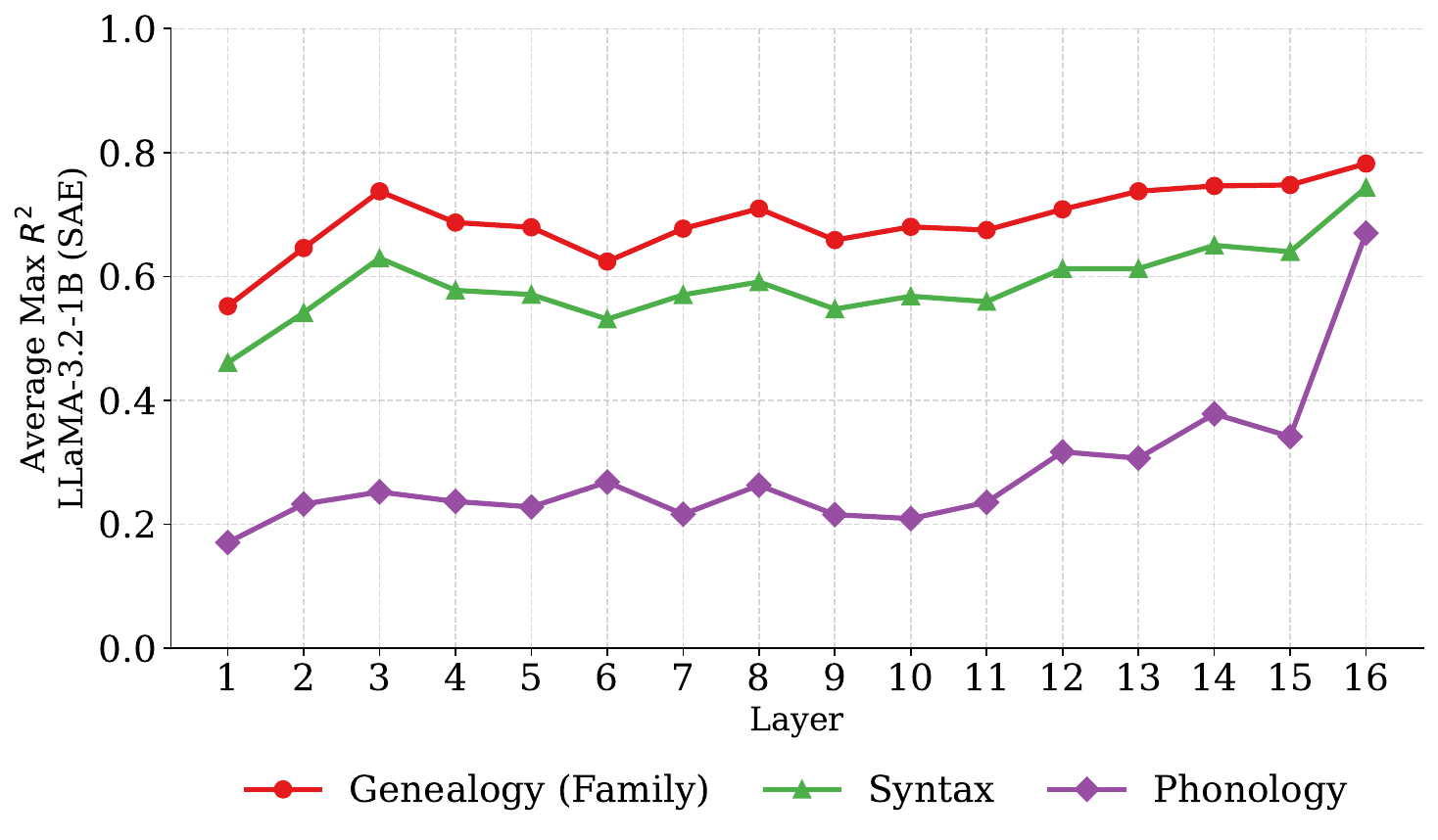}
    \caption{
    Layerwise probing performance in Llama-3.2-1B.
    \textbf{Top:} Raw MLP activations.
    \textbf{Bottom:} SAE features.
    SAE representations are comparatively stronger in early layers, while raw activations dominate in later layers.}
    \label{fig:llama_raw_sae_layerwise}
\end{figure}

\begin{figure}[!t]
    \centering
    \includegraphics[width=0.9\linewidth]{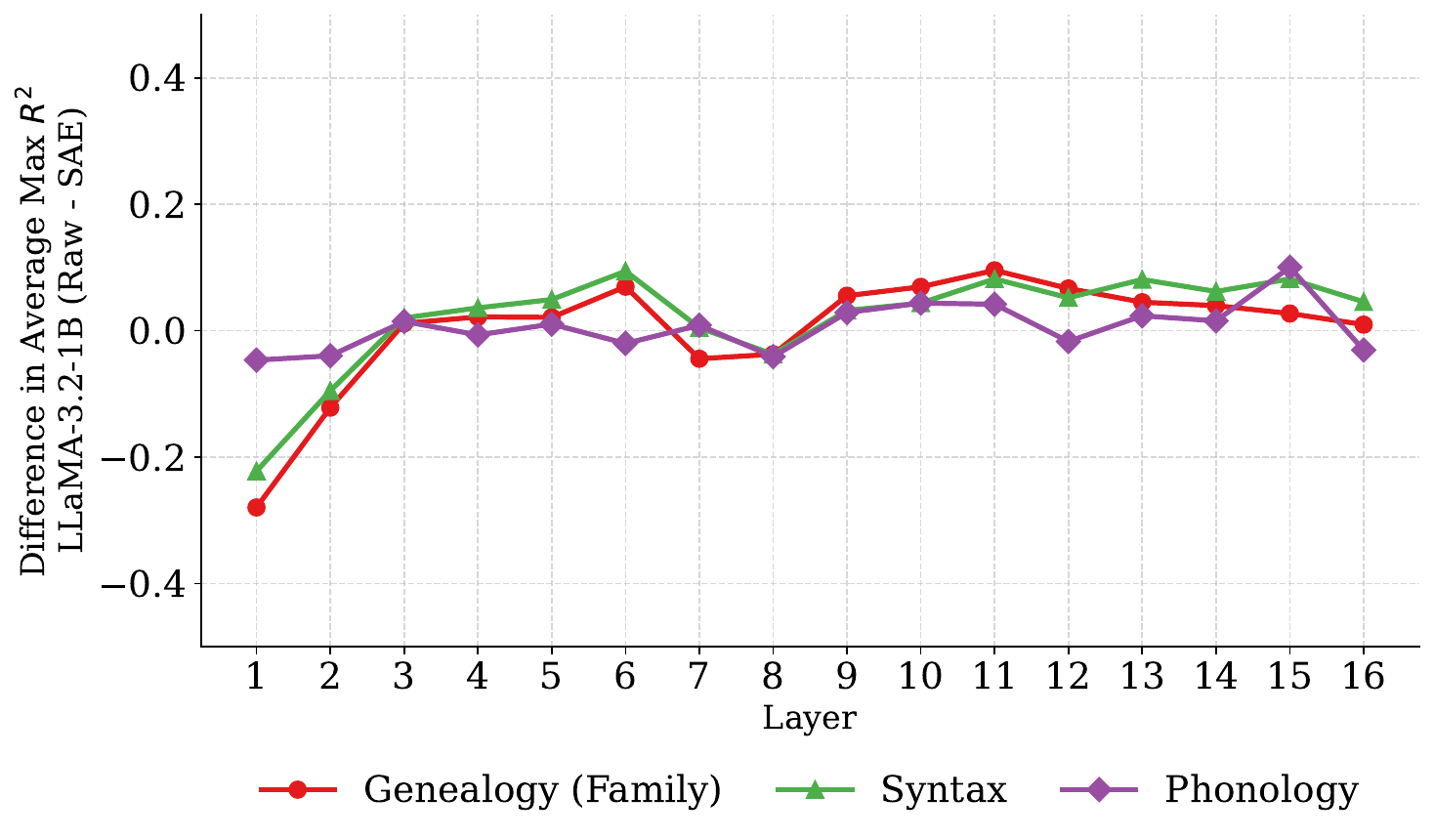}
    \caption{
    Raw minus SAE probing score differences for Llama-3.2-1B.
    Negative values in shallow layers indicate higher SAE informativeness, while the gradual shift toward positive values reflects increasing raw dominance with depth.}
    \label{fig:llama_raw_vs_sae_diff}
\end{figure}

\paragraph{Raw vs.\ SAE representations in Llama.}
Figure~\ref{fig:llama_raw_sae_layerwise} shows the layerwise probing trends for Llama raw and SAE representations, while Figure~\ref{fig:llama_raw_vs_sae_diff} visualizes their differences directly.
In early layers, SAE features are more informative than raw MLP activations for all three feature families, resulting in negative raw--SAE differences. This indicates that SAE training amplifies weak but structured typological signals that are only diffusely present in shallow raw activations. As depth increases, this advantage steadily diminishes, and the difference approaches positive values, indicating that raw representations become more linearly informative in deeper layers. This transition reflects a shift from early sparse amplification to richer distributed encoding in later layers.

\begin{figure}[!t]
    \centering
    \includegraphics[width=0.9\linewidth]{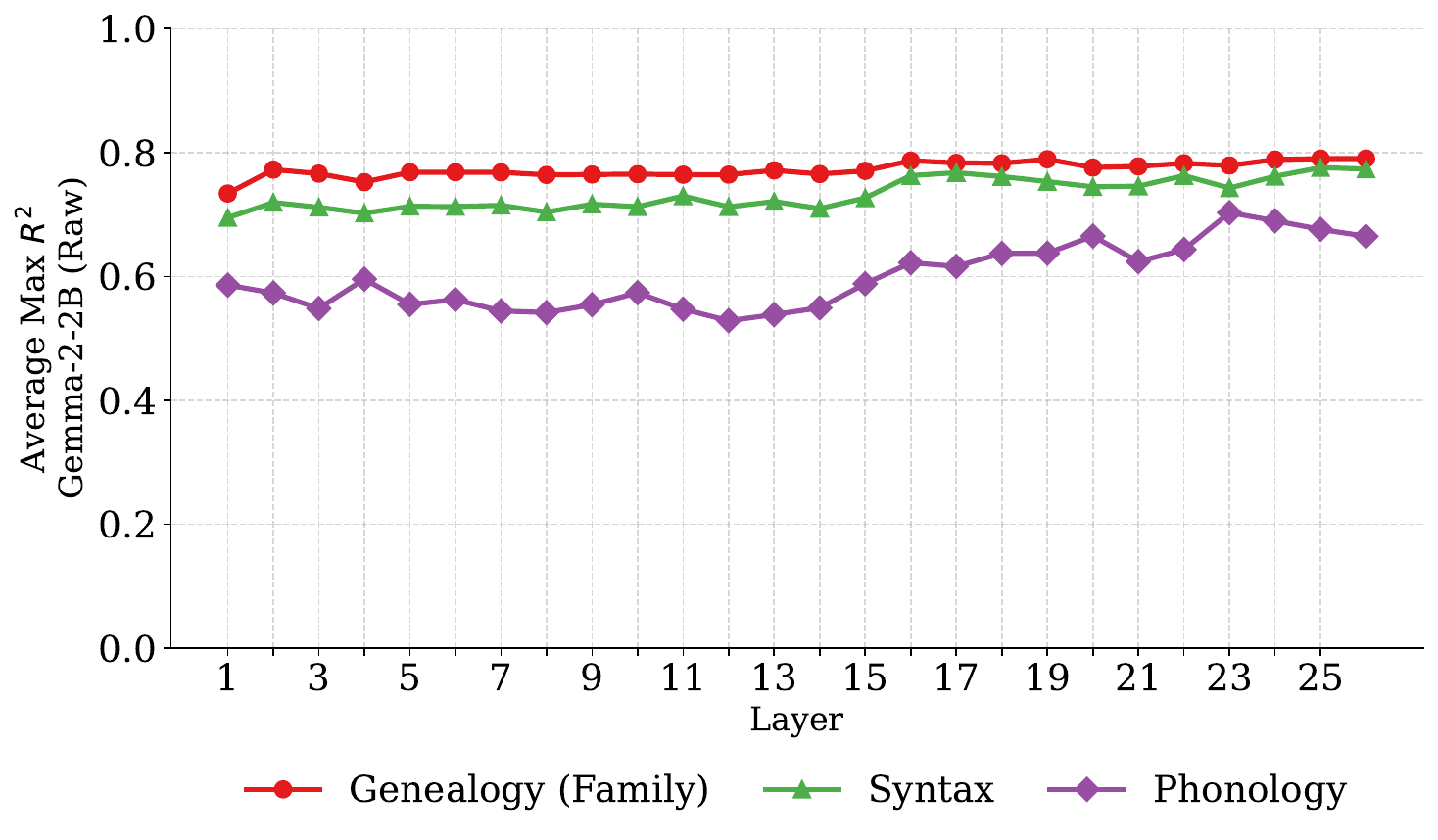}
    \vspace{0.6em}
    \includegraphics[width=0.9\linewidth]{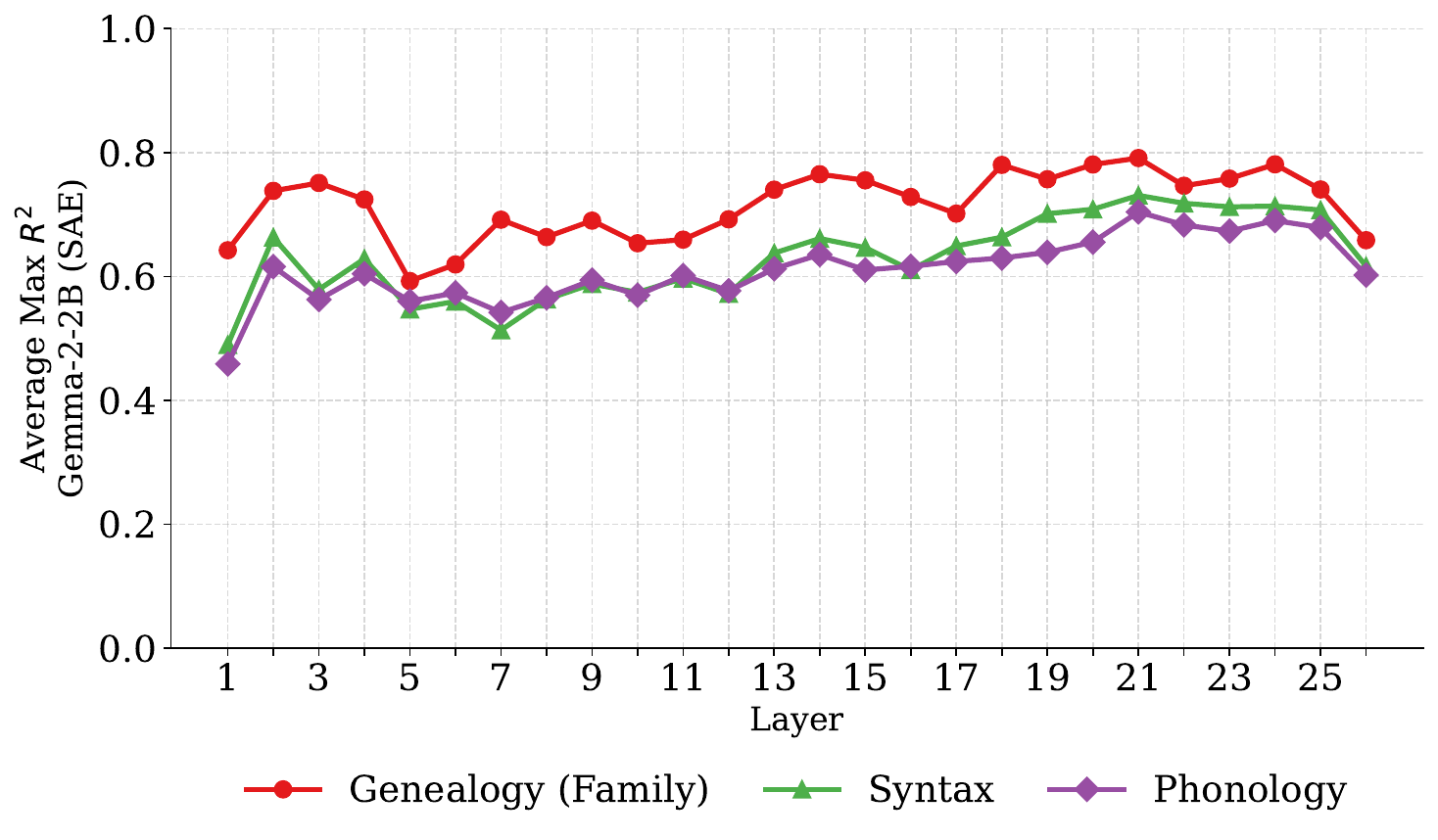}
    \caption{
    Layerwise probing performance in Gemma-2-2B.
    \textbf{Top:} Raw MLP activations.
    \textbf{Bottom:} SAE features.
    Raw representations dominate for \texttt{fam} and \texttt{syntax}, while SAE features retain stronger phonological signals across layers.}
    \label{fig:gemma_raw_sae_layerwise}
\end{figure}

\begin{figure}[!t]
    \centering
    \includegraphics[width=0.9\linewidth]{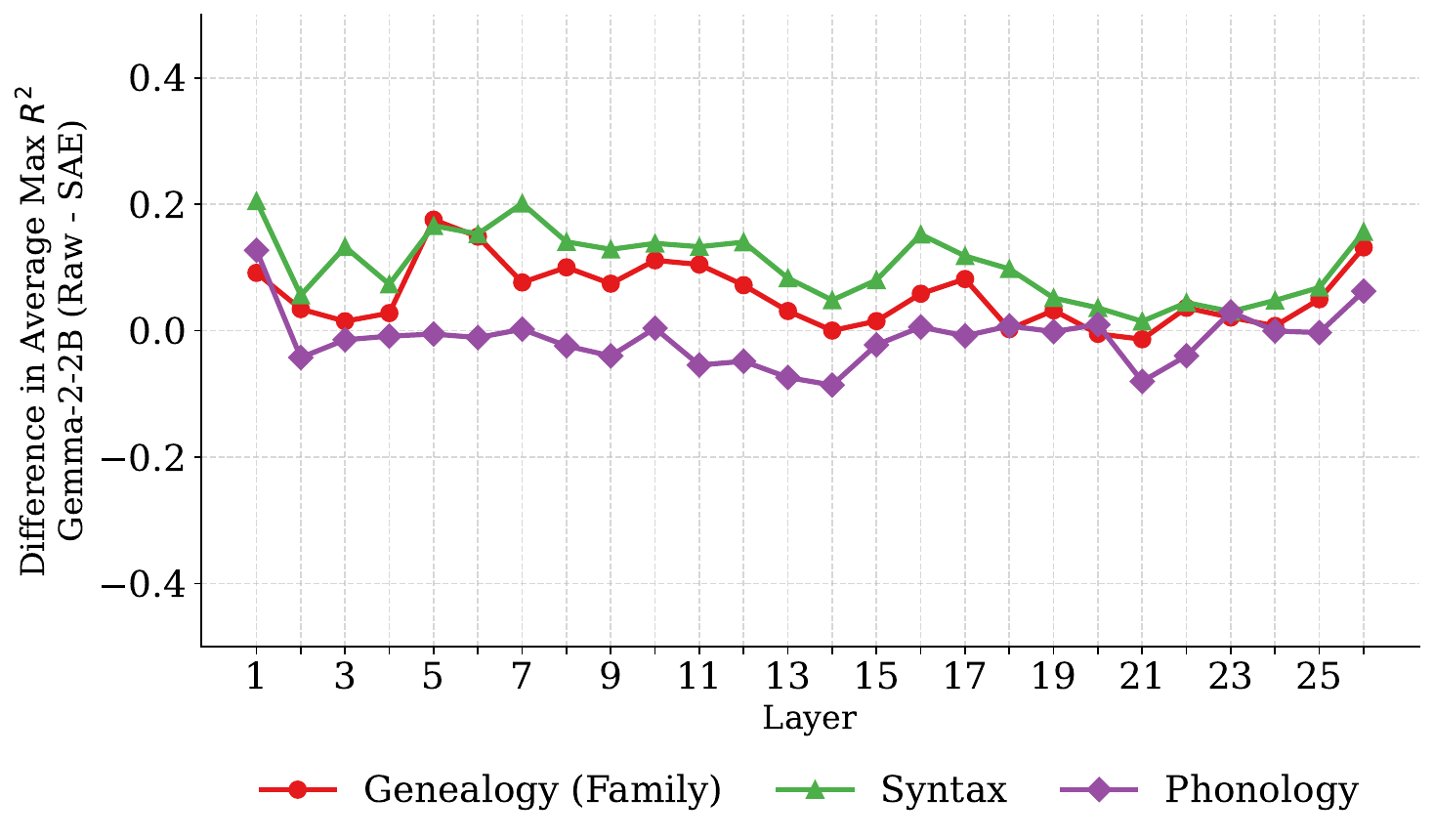}
    \caption{
    Raw minus SAE probing score differences for Gemma-2-2B.
    Differences are stable across depth: positive for \texttt{fam} and \texttt{syntax}, and negative for \texttt{phonology}.}
    \label{fig:gemma_raw_vs_sae_diff}
\end{figure}

\paragraph{Raw vs.\ SAE representations in Gemma.}
The corresponding Gemma plots are shown in Figures~\ref{fig:gemma_raw_sae_layerwise} and~\ref{fig:gemma_raw_vs_sae_diff}. Unlike Llama, Gemma exhibits a more stable relationship between raw and SAE representations across layers.
For \texttt{fam} and \texttt{syntax}, raw activations are consistently more informative than SAE features, yielding positive differences across depth. In contrast, \texttt{phonology} shows consistently negative differences, indicating that Gemma SAEs preferentially preserve phonological structure relative to raw MLP activations. This feature-specific asymmetry suggests that sparse factorization interacts differently with lower-level sound-related abstractions than with genealogical or syntactic structure.

\begin{figure}[!t]
    \centering
    \includegraphics[width=0.9\linewidth]{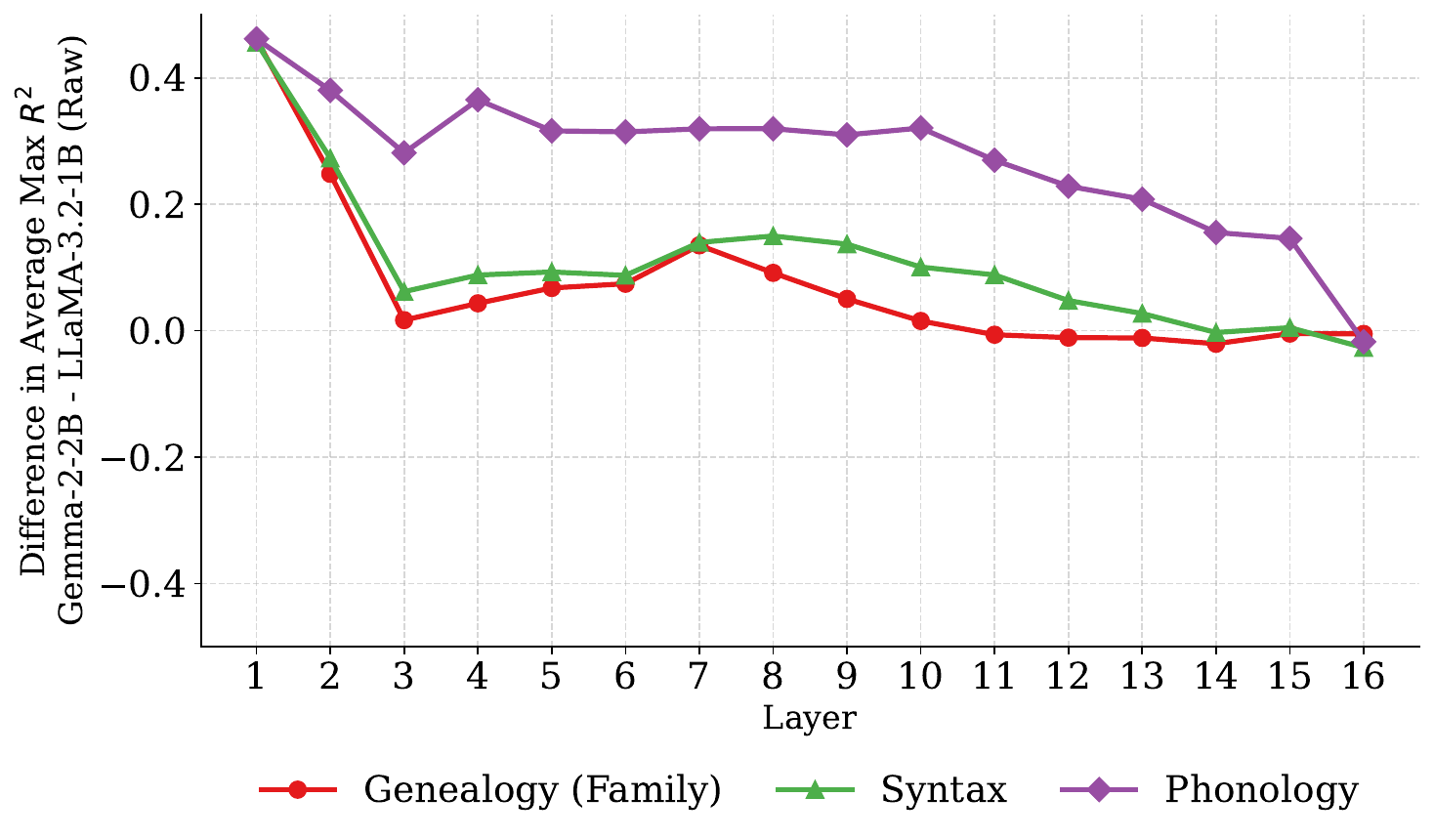}
    \vspace{0.6em}
    \includegraphics[width=0.9\linewidth]{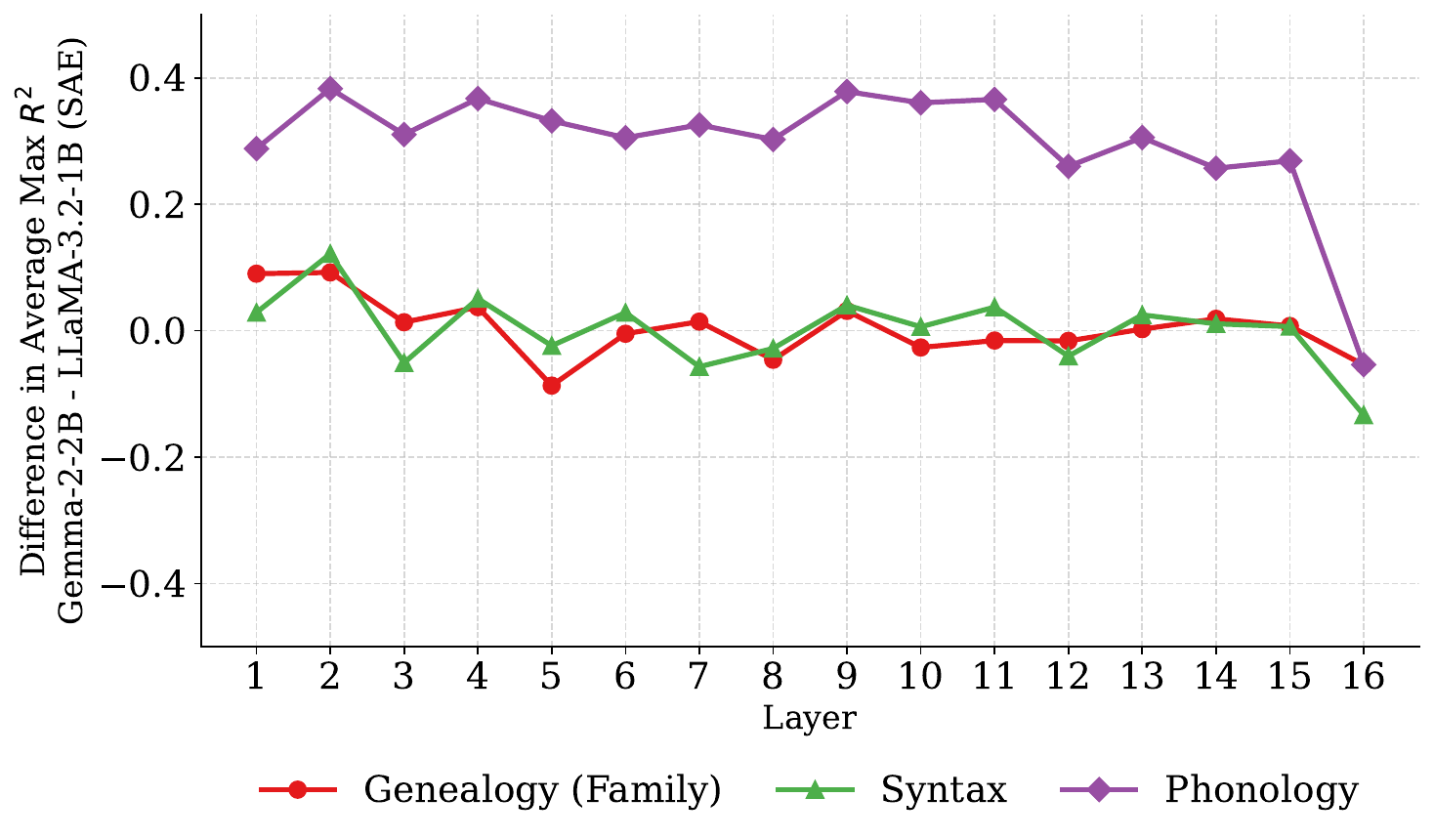}
    \caption{
    Cross-model comparison of probing performance.
    \textbf{Top:} Raw MLP activations.
    \textbf{Bottom:} SAE features.
    Raw representations show large early-layer differences, especially for phonology, followed by sharp convergence with depth, while SAE representations compress these disparities.}
    \label{fig:cross_model_comparison}
\end{figure}

\paragraph{Cross-Model Comparison: Llama vs.\ Gemma.}
Figure~\ref{fig:cross_model_comparison} presents direct comparisons between Llama and Gemma under matched representational settings.
Raw MLP activations exhibit stark cross-model differences in shallow layers for all three feature families, with phonology showing substantially larger gaps than \texttt{fam} or \texttt{syntax}. Moreover, raw cross-model differences decrease sharply with depth, producing a pronounced downward trend across all feature families. This suggests that early typological representations are strongly shaped by architectural and tokenizer-specific factors, while deeper layers converge toward more similar abstractions.
In contrast, SAE representations substantially attenuate these differences. Although phonology remains the most discriminative feature family, the overall magnitude and depth-dependence of cross-model differences are reduced, indicating that sparse representations emphasize later-stage, shared abstractions over model-specific surface variation.

\paragraph{Summary.}
These detailed comparisons show that sparse autoencoding reshapes typological structure in a depth-, model-, and feature-dependent manner. Llama SAEs transiently enhance early-layer typological accessibility, Gemma SAEs selectively favor Phonology features, and phonology consistently emerges as the most sensitive axis for cross-model differences -- particularly in shallow raw representations.

\section{Causal Interventions on Invariant Neuron Sets}
\label{app:causal}

This appendix reports causal intervention experiments designed to assess whether neuron subsets identified via invariance-based analyses are functionally necessary for multilingual language modeling. We intervene on neuron sets defined by their stability under controlled input perturbations: word-order shuffling and script romanization, done in sections \ref{sec:shuffling} and \ref{sec:romanization} respectively. All experiments are conducted on the first 100 examples per language from the FLORES+ dataset.

\subsection{Neuron Selection via Shuffling and Romanization}

Neuron subsets are derived from earlier analyses that characterize neuron behavior under targeted surface perturbations.

\paragraph{Shuffling-Based Neuron Sets.}
Using word-level shuffling experiments, neurons are categorized into:
\begin{itemize}[nosep, wide, labelwidth=!, labelindent=0pt]
    \item[(i)]\textbf{Overlap neurons:} Neurons consistently identified under both normal and shuffled inputs. These neurons are invariant to word-order perturbations and are hypothesized to encode structurally necessary representations.
    \item[(ii)] \textbf{Only-unshuffled neurons:} Neurons identified only under normal inputs and absent under shuffled conditions. These neurons are sensitive to surface word order and local syntactic structure.
\end{itemize}

\paragraph{Romanization-Based Neuron Sets.}
Using native-script versus romanized inputs, neurons are grouped into:
\begin{itemize}[nosep, wide, labelwidth=!, labelindent=0pt]
    \item[(i)] \textbf{Overlap neurons:} Neurons shared across native and romanized scripts, hypothesized to encode script-invariant representations.
    \item[(ii)] \textbf{Only-native neurons:} Neurons active only for native-script inputs and whose functional signature disappears under romanization, indicating sensitivity to surface orthography.
\end{itemize}

Across both regimes, \emph{overlap neurons} are defined by invariance to the corresponding perturbation, while non-overlap neurons capture sensitivity to surface form.  
For all experiments, matched \textbf{random control sets} are constructed by sampling an equal number of neurons uniformly from the overall neuron pool of the model.

\subsection{Intervention Protocol}

All experiments are conducted on raw model activations.

\paragraph{Ablation Scope.}
To avoid layer-local confounds, we apply \emph{simultaneous ablation across all layers}. For each layer $\ell$, activations of the selected neuron set are modified during the forward pass.

\paragraph{Ablation Types.}
We consider:
\begin{itemize}[nosep, wide, labelwidth=!, labelindent=0pt]
    \item[(i)] \textbf{Zero ablation for shuffling-based neuron sets:} Activations are set to zero.
    \item[(ii)] \textbf{Cross-language mean ablation for romanization-based neuron sets:} Activations are replaced by mean activation vectors computed from another language.
\end{itemize}

Mean vectors are computed over the corresponding FLORES+ split of the source language.

\subsection{Evaluation Metrics and Statistical Testing}

For each example, we compute clean and patched perplexities ($PPL_{\text{clean}}$, $PPL_{\text{patch}}$), perplexity ratios, and perplexity deltas ($\Delta PPL$).  
Paired-sample $t$-tests compare targeted ablations against matched random controls over the 100 examples, with significance assessed at $p < 0.05$.


\subsection{Causal Intervention Results: Llama-3.2-1B and Llama-3-8B}

\subsubsection{Shuffling-Based Zero Ablation}









\begin{table*}[!t]
\centering
\small
\setlength{\tabcolsep}{5.5pt}
\begin{tabular}{l l| c c c| c c c}
\toprule
Lang & Category &
$PPL_{ratio}^{\text{target}}$ &
$PPL_{ratio}^{\text{ctrl}}$ &
$p$ (ratio) &
$\Delta PPL^{\text{target}}$ &
$\Delta PPL^{\text{ctrl}}$ &
$p$ ($\Delta$) \\
\midrule
\multicolumn{8}{c}{\textbf{Llama-3.2-1B}} \\
\midrule
en & overlap
& 1.116 & 0.954 & $1.5{\times}10^{-50}$
& +272.3 & $-$108.6 & $2.2{\times}10^{-35}$ \\
en & only-unshuffled
& 0.963 & 1.044 & $3.0{\times}10^{-48}$
& $-$87.1 & +99.6 & $1.9{\times}10^{-34}$ \\
hi & overlap
& 2.786 & 1.055 & $2.2{\times}10^{-19}$
& +1914.0 & +204.9 & $1.6{\times}10^{-12}$ \\
hi & only-unshuffled
& 1.083 & 0.947 & $4.1{\times}10^{-40}$
& +228.6 & $-$133.3 & $4.6{\times}10^{-10}$ \\
fr & overlap
& 1.118 & 1.030 & $5.4{\times}10^{-6}$
& +114.6 & +74.6 & 0.145 \\
fr & only-unshuffled
& 0.935 & 0.957 & $2.8{\times}10^{-10}$
& $-$130.7 & $-$85.8 & $5.5{\times}10^{-7}$ \\
zh & overlap
& 1.217 & 0.960 & $2.4{\times}10^{-17}$
& +952.6 & $-$523.2 & $2.4{\times}10^{-24}$ \\
zh & only-unshuffled
& 0.936 & 0.982 & $5.7{\times}10^{-17}$
& $-$695.3 & $-$210.4 & $4.1{\times}10^{-16}$ \\
\midrule
\multicolumn{8}{c}{\textbf{Llama-3-8B}} \\
\midrule
en & overlap
& 0.925 & 0.992 & $3.7{\times}10^{-8}$
& $-$23.3 & $-$1.7 & $5.3{\times}10^{-6}$ \\
en & only-unshuffled
& 0.832 & 0.993 & $1.3{\times}10^{-59}$
& $-$40.1 & $-$1.1 & $1.8{\times}10^{-23}$ \\
hi & overlap
& 4.348 & 1.021 & $4.4{\times}10^{-12}$
& +289.7 & +10.5 & $2.2{\times}10^{-10}$ \\
hi & only-unshuffled
& 0.859 & 1.017 & $4.0{\times}10^{-30}$
& $-$26.8 & +3.1 & $4.0{\times}10^{-10}$ \\
fr & overlap
& 0.959 & 0.955 & 0.813
& $-$29.1 & $-$15.1 & 0.134 \\
fr & only-unshuffled
& 0.867 & 1.020 & $4.6{\times}10^{-19}$
& $-$59.1 & +15.2 & $6.8{\times}10^{-12}$ \\
zh & overlap
& 1.236 & 0.990 & $1.1{\times}10^{-17}$
& +53.0 & $-$5.0 & $4.6{\times}10^{-17}$ \\
zh & only-unshuffled
& 0.934 & 0.982 & $2.2{\times}10^{-14}$
& $-$20.9 & $-$6.4 & $8.0{\times}10^{-11}$ \\
\bottomrule
\end{tabular}
\caption{
Causal zero-ablation results for shuffling-derived neuron sets across models (Llama models, raw neurons).
Values report means over the first 100 FLORES+ examples per language.
Control sets consist of matched random neurons with identical cardinality.
Llama-3-8B shows qualitatively similar directional effects to Llama-3.2-1B, with stronger amplification in Hindi and Chinese, while maintaining robustness patterns under only-unshuffled conditions.
}
\label{tab:causal_shuffling}
\end{table*}
Table~\ref{tab:causal_shuffling} reports results for shuffling-derived neuron sets. Across languages, ablation of \emph{overlap neurons} induces the largest and most consistent degradations. For instance, Hindi in Llama-3.2-1B exhibits the strongest effect, with $PPL$ ratios approaching $2.8$ and $\Delta PPL$ exceeding $+1900$. Chinese shows similarly pronounced degradation, while English and French exhibit smaller but still significant effects. In all cases, overlap ablations degrade performance substantially more than matched random controls.

In contrast, \emph{only-unshuffled neurons} yield weaker and sometimes inverted effects. For English and French, ablation leads to reductions in perplexity relative to clean runs. Hindi shows a small increase, but far weaker than overlap ablations, while Chinese exhibits negative $\Delta PPL$ despite statistical significance. These patterns indicate that only-unshuffled neurons encode order-sensitive surface regularities that are largely redundant for language modeling.

Moreover, the larger model (Llama-3-8B) produces stronger effects for Hindi and Chinese as compared to the (Llama-3.2-1B).

\begin{figure*}[!h]
    \centering
    \includegraphics[width=0.9\linewidth]{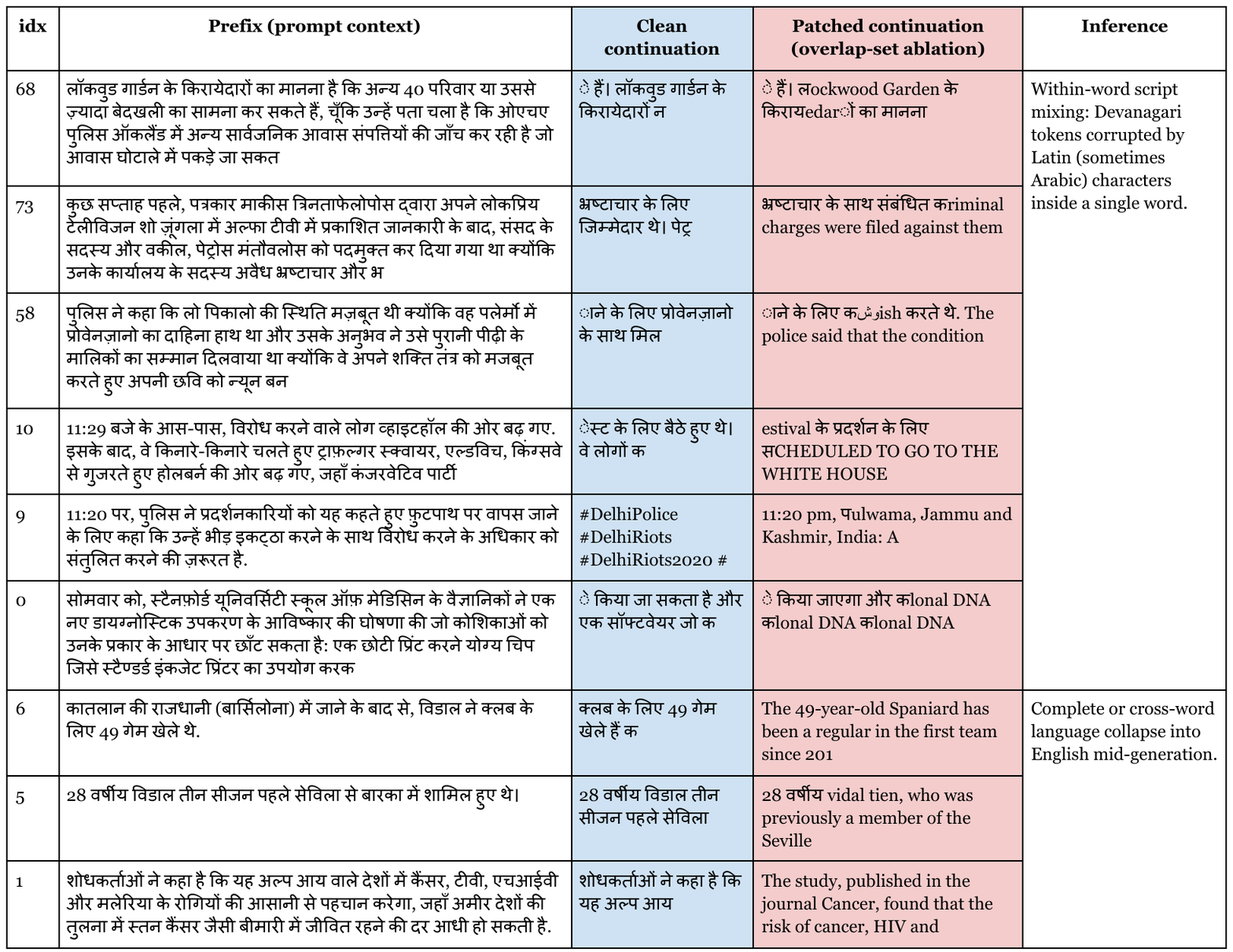}
    \caption{
Representative Hindi generations under shuffling-based overlap-neuron ablation (Llama-3.2-1B, raw).
Each row shows the input prefix, clean continuation, and ablated continuation for the same example.
While clean generations preserve Devanagari script integrity, overlap-neuron ablations induce within-word mixed-script corruption, abrupt language switching, and topic drift.
Such token-internal script mixing is not observed under matched random or Only-unshuffled neuron ablations.
}

    \label{fig:shuffling_qualitative}
\end{figure*}

\paragraph{Qualitative Effects Under Shuffling.}
As shown in Figure~\ref{fig:shuffling_qualitative}, ablation of overlap neurons induces systematic qualitative failures in Hindi. Most notably, we observe \emph{within-word script mixing}, where individual lexical items combine Devanagari and Latin characters (e.g., mixed-script morphemes). This phenomenon is not observed under random or Only-unshuffled ablations, where script switching -- if present -- occurs only at word boundaries. These qualitative failures align with the large perplexity degradation and indicate that overlap neurons play a role in maintaining subword-level orthographic coherence.

\subsubsection{Romanization-Based Cross-Language Mean Ablation}





\begin{table*}[!t]
\centering
\small
\setlength{\tabcolsep}{6pt}
\begin{tabular}{l | l | c c c | c c c}
\toprule
Lang & Category &
$PPL_{ratio}^{\text{target}}$ &
$PPL_{ratio}^{\text{ctrl}}$ &
$p$ (ratio) &
$\Delta PPL^{\text{target}}$ &
$\Delta PPL^{\text{ctrl}}$ &
$p$ ($\Delta$) \\
\midrule
\multicolumn{8}{c}{\textbf{Llama-3.2-1B}} \\
\midrule
en & overlap
& 0.947 & 0.991 & $6.9{\times}10^{-45}$
& $-$127.1 & $-$21.1 & $9.7{\times}10^{-28}$ \\
en & only-native
& 1.498 & 0.955 & $5.0{\times}10^{-89}$
& +1176.9 & $-$104.9 & $3.1{\times}10^{-40}$ \\
hi & overlap
& 1.047 & 0.982 & $9.5{\times}10^{-34}$
& +79.1 & $-$53.8 & $1.7{\times}10^{-7}$ \\
hi & only-native
& 0.312 & 0.970 & $1.2{\times}10^{-38}$
& $-$1800.5 & $-$92.5 & $7.7{\times}10^{-11}$ \\
\midrule
\multicolumn{8}{c}{\textbf{Llama-3-8B}} \\
\midrule
en & overlap
& 0.946 & 0.989 & $2.9{\times}10^{-3}$
& $-$15.4 & $-$2.0 & $1.1{\times}10^{-3}$ \\
en & only-native
& 0.817 & 0.999 & $1.2{\times}10^{-18}$
& $-$46.4 & +0.2 & $2.1{\times}10^{-10}$ \\
hi & overlap
& 1.062 & 0.990 & $2.1{\times}10^{-12}$
& +7.8 & $-$3.4 & $3.6{\times}10^{-3}$ \\
hi & only-native
& 7.738 & 0.948 & $3.5{\times}10^{-29}$
& +1326.6 & $-$15.4 & $2.8{\times}10^{-11}$ \\
\bottomrule
\end{tabular}
\caption{
Cross-language mean ablation results for romanization-derived neuron sets (raw neurons).
Rows indicate forward-pass language; mean activations are taken from the opposite language.
Results are averaged over the first 100 FLORES+ examples.
Control sets consist of matched random neurons with identical cardinality.
Llama-3-8B exhibits qualitatively similar patterns to Llama-3.2-1B: only-native Hindi neurons cause dramatic perplexity changes when ablated during Hindi inference ($PPL_{ratio}^{\text{target}} = 7.74$), while overlap neurons produce modest effects across both models.
}
\label{tab:causal_romanization}
\end{table*}

Table~\ref{tab:causal_romanization} summarizes results for cross-language ablations between Hindi and English for both Llama models. Replacing overlap neuron activations across languages yields relatively mild effects. English shows a slight decrease in perplexity, while Hindi shows a modest increase. The small magnitude of these effects suggests that overlap neurons encode representations that are largely invariant to script and language identity.

In contrast, replacing only-native neuron activations leads to extreme effects. Generations in both languages are severely affected. Qualitative inspection for the Hindi-to-English ablations in Llama-3.2-1B reveals that these reductions arise from language switching rather than improved Hindi modeling: many generations abandon Hindi entirely and continue fluently in English which has higher likelihood under the model. Sometimes the generations switch to other languages like Bengali, which attributes to the drastic increase in perplexity for Llama-3-8B. 


\subsection{Causal Intervention Results: Gemma-2-2B}
\label{app:causal_results}

We repeat the same analyses on Gemma-2-2B to assess cross-model consistency.

\subsubsection{Shuffling-Based Zero Ablation}

\begin{table*}[!t]
\centering
\small
\setlength{\tabcolsep}{5.5pt}
\begin{tabular}{l| l| c c c| c c c}
\toprule
Lang & Category &
$PPL_{ratio}^{\text{target}}$ &
$PPL_{ratio}^{\text{ctrl}}$ &
$p$ (ratio) &
$\Delta PPL^{\text{target}}$ &
$\Delta PPL^{\text{ctrl}}$ &
$p$ ($\Delta$) \\
\midrule
en & overlap
& 3.045 & 0.312 & $3.5{\times}10^{-81}$
& +588.9 & $-$191.2 & $2.5{\times}10^{-39}$ \\

en & only-unshuffled
& 0.799 & 0.312 & $2.8{\times}10^{-130}$
& $-$56.5 & $-$191.2 & $1.1{\times}10^{-47}$ \\

hi & overlap
& 1.109 & 0.953 & 0.173
& $-$34.2 & $-$7.6 & $1.0{\times}10^{-3}$ \\

hi & only-unshuffled
& 0.397 & 2.557 & $8.1{\times}10^{-50}$
& $-$102.7 & +289.9 & $3.6{\times}10^{-22}$ \\

fr & overlap
& 1.547 & 0.952 & $1.2{\times}10^{-89}$
& +116.1 & $-$10.0 & $4.4{\times}10^{-36}$ \\

fr & only-unshuffled
& 1.260 & 1.403 & $3.1{\times}10^{-40}$
& +56.3 & +90.4 & $4.4{\times}10^{-24}$ \\

zh & overlap
& 0.842 & 0.150 & $2.0{\times}10^{-52}$
& $-$58.8 & $-$241.5 & $1.3{\times}10^{-56}$ \\

zh & only-unshuffled
& 0.082 & 3.152 & $6.0{\times}10^{-76}$
& $-$259.8 & +645.7 & $8.5{\times}10^{-38}$ \\
\bottomrule
\end{tabular}
\caption{
Causal zero-ablation results for shuffling-derived neuron sets (Gemma-2-2B, raw).
Values report means over the first 100 FLORES+ examples per language.
Control sets consist of matched random neurons with identical cardinality.
}
\label{tab:shuffling_ablation_gemma}
\end{table*}

\begin{figure*}[!h]
    \centering
    \includegraphics[width=0.8\linewidth]{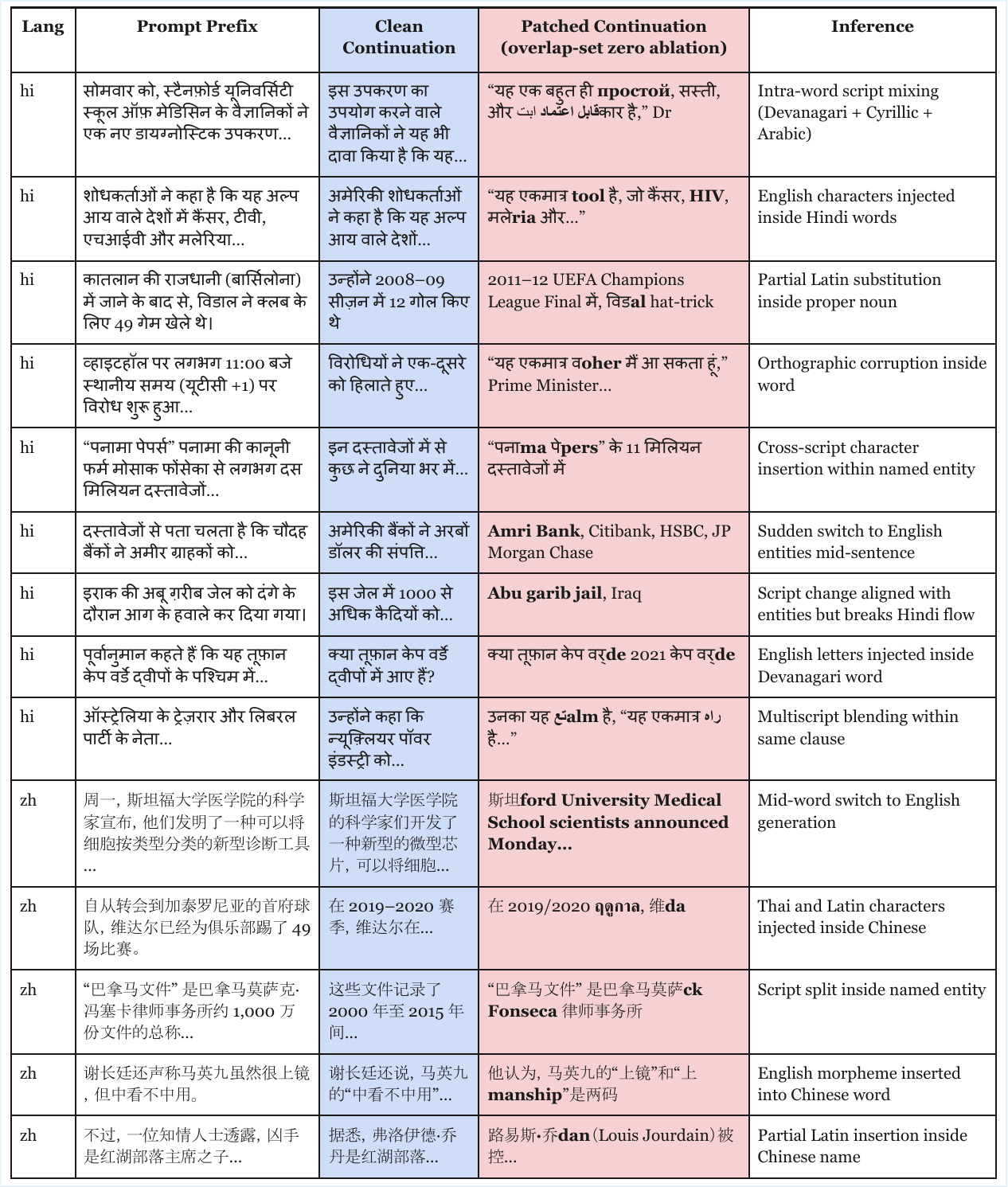}
    \caption{Qualitative examples of model behavior under shuffling-based overlap ablation (Gemma-2-2B, raw). Ablation of overlap neurons induces systematic script mixing, including partial Latin insertions and mixed-script morphemes occurring within words. Such orthographic violations are not observed for random or only-normal neuron ablations.}
    \label{fig:gemma-gen-ex-acl}
\end{figure*}


\begin{table*}[!th]
\centering
\small
\setlength{\tabcolsep}{6pt}
\begin{tabular}{l | l | c c c | c c c}
\toprule
Lang & Category &
$PPL_{ratio}^{\text{target}}$ &
$PPL_{ratio}^{\text{ctrl}}$ &
$p$ (ratio) &
$\Delta PPL^{\text{target}}$ &
$\Delta PPL^{\text{ctrl}}$ &
$p$ ($\Delta$) \\
\midrule
en & only-native
& 5.208 & 1.405 & $2.2{\times}10^{-65}$
& +1222.6 & +114.6 & $8.2{\times}10^{-35}$ \\

en & overlap
& 0.899 & 0.822 & $6.2{\times}10^{-96}$
& $-$28.3 & $-$50.0 & $3.6{\times}10^{-43}$ \\

hi & only-native
& 0.684 & 1.136 & $1.2{\times}10^{-54}$
& $-$55.9 & +24.1 & $2.4{\times}10^{-24}$ \\

hi & overlap
& 2.228 & 1.036 & $5.8{\times}10^{-45}$
& +228.0 & +6.5 & $7.1{\times}10^{-21}$ \\
\bottomrule
\end{tabular}
\caption{
Cross-language mean ablation results for romanization-derived neuron sets (Gemma-2-2B, raw).
Rows indicate forward-pass language; mean activations are taken from the opposite language.
Results are averaged over the first 100 FLORES+ examples.
}
\label{tab:romanization_ablation_gemma}
\end{table*}

Table~\ref{tab:shuffling_ablation_gemma} reports shuffling-based interventions for Gemma-2-2B. As in Llama, ablation of \emph{overlap neurons} produces the strongest disruptions across languages. English, French, and Chinese show large increases in perplexity relative to random controls, while Hindi exhibits weaker but directionally consistent effects. Paired tests confirm that these differences are statistically significant in nearly all cases.
Only-unshuffled neurons again yield weaker and more variable effects. In several languages, ablation produces smaller changes than random controls or even reduces perplexity, reinforcing the conclusion that these neurons encode word-order-level regularities rather than load-bearing structure, and that the robust shuffling-overlap neuron sets are more correlated with orthographic and subword-level structures.

\paragraph{Qualitative Effects Under Shuffling.}
Figure~\ref{fig:gemma-gen-ex-acl} presents representative Hindi and Chinese generations. Similar to Llama, overlap-neuron ablation induces \emph{script changes within words}, including partial Latin insertions and mixed-script morphemes. Crucially, such intra-word script violations do not appear under random or Only-unshuffled ablations, indicating that overlap neurons support low-level orthographic coordination during decoding. More importantly, fluency is not lost while ablating the overlapping neurons, indicating that these neurons are not responsible for syntactic behavior.

\subsubsection{Romanization-Based Cross-Language Mean Ablation}

Table~\ref{tab:romanization_ablation_gemma} summarizes romanization-based interventions.
Only-native neurons exhibit the largest sensitivity: English-to-Hindi replacement causes large perplexity increases, while Hindi-to-English replacement often yields perplexity reductions. As in Llama, inspection of generations reveals frequent language switching in the latter case, explaining the apparent improvement.
Overlap neurons again show smaller and more symmetric effects, consistent with a script-invariant functional role.

\subsection{Summary}

Across both models and perturbation regimes, a consistent causal pattern emerges:
\begin{itemize}[nosep, wide, labelwidth=!, labelindent=0pt]
    \item \textbf{Shuffling-Overlap neurons} -- defined by invariance to shuffling -- form a causally necessary backbone supporting stable, script-consistent generation. They are not causally related to fluency, rather script and subword-level regularity. Hence, the features tied strongly to script are more causally important for generation.
    \item \textbf{Romanization-Overlap neurons} -- defined by invariance to romanization -- are largely script insensitive. This suggests that representations that are not tied to script, are not causally important in generation.
    \item \textbf{Only-unshuffled neurons} encode order-sensitive surface regularities that are largely redundant for orthographic structure.
    \item \textbf{Only-native neurons} anchor script-specific realization, and their disruption induces language switching rather than structured degradation. Again, this reinforces the hypothesis that script-related neurons are most causally important.
\end{itemize}

The convergence of quantitative metrics and qualitative failure modes across Llama and Gemma indicates that invariance-based neuron identification isolates functionally meaningful components of multilingual language models.

\begin{table*}[!h]
\centering
\small
\setlength{\tabcolsep}{8pt}
\begin{tabular}{l | c c | c c | c c | c c}
\toprule
& \multicolumn{2}{c|}{\textbf{Llama-3.2-1B}} & \multicolumn{2}{c|}{\textbf{Llama-3-8B}} & \multicolumn{2}{c|}{\textbf{Gemma-2-2B}} & \multicolumn{2}{c}{\textbf{Gemma-2-9B}} \\
Language & Nat & Rom & Nat & Rom & Nat & Rom & Nat & Rom \\
\midrule
Bengali   & 0.40 & 0.08 & 0.67 & 0.23 & 0.60 & 0.11 & 0.72 & 0.33 \\
Bulgarian & 0.66 & 0.36 & 0.77 & 0.62 & 0.74 & 0.46 & 0.79 & 0.69 \\
Chinese   & 0.60 & 0.12 & 0.69 & 0.29 & 0.68 & 0.12 & 0.72 & 0.33 \\
Hindi     & 0.58 & 0.14 & 0.72 & 0.42 & 0.69 & 0.22 & 0.76 & 0.53 \\
Japanese  & 0.54 & 0.08 & 0.67 & 0.14 & 0.65 & 0.11 & 0.71 & 0.19 \\
Korean    & 0.53 & 0.08 & 0.68 & 0.14 & 0.64 & 0.10 & 0.71 & 0.20 \\
Marathi   & 0.46 & 0.10 & 0.66 & 0.23 & 0.58 & 0.12 & 0.72 & 0.33 \\
Russian   & 0.68 & 0.42 & 0.75 & 0.67 & 0.73 & 0.52 & 0.76 & 0.72 \\
Spanish   & 0.68 & 0.67 & 0.73 & 0.73 & 0.72 & 0.71 & 0.74 & 0.74 \\
Urdu      & 0.47 & 0.07 & 0.68 & 0.19 & 0.60 & 0.10 & 0.73 & 0.32 \\
\bottomrule
\end{tabular}
\caption{
Translation performance (BERTScore, \texttt{roberta-large}) from Native (Nat) and Romanized (Rom) inputs to English (8-shot). Large BERTScores on romanized inputs confirm genuine semantic competence, proving that low neuron overlap is not a result of data sparsity.
}
\label{tab:translation_romanized}
\end{table*}

\section{Extended Scaling and Semantic Competence Analysis}
\label{app:scaling}

This appendix provides a comprehensive evaluation of our findings on larger model architectures: Llama-3-8B and Gemma-2-9B. We analyze whether the observed representational fragmentation is a symptom of limited capacity or data sparsity, or whether it persists as a stable architectural trait at scale.

\subsection{Translation Performance on Romanized Inputs}
\label{app:translation_competence}

To rule out the hypothesis that low representational overlap is an artifact of undertraining or data sparsity, we evaluate translation performance from native and romanized inputs to English.

\paragraph{Experimental Setup.}
We use the \texttt{dev} split of FLORES+ in an 8-shot setting. Translation quality is measured using BERTScore (\texttt{roberta-large}) against gold English references across 10 diverse languages.

\paragraph{Results and Discussion.}
As shown in Table~\ref{tab:translation_romanized}, larger models achieve robust translation performance on romanized inputs. For instance, Llama-3-8B obtains BERTScores of 0.67 for Romanized Russian and 0.42 for Hindi, confirming the models have learned the semantics of romanized text and are not treating it as out-of-distribution noise.

Critically, despite this functional competence, representational overlap remains comparatively low (${\sim}0.25$ for Llama-3-8B vs.\ ${\sim}0.11$ for Llama-3.2-1B), which is significantly lower than the ${\sim}0.60$ overlap observed under word-order shuffling. This dissociation between competence and alignment confirms that models process different scripts through disjoint subspaces as a representational choice. This pattern persists even for high-resource languages like Russian and Spanish, where romanized performance nearly matches native-script performance, effectively ruling out undertraining as the sole explanation for representational fragmentation.

\subsection{Script Fragmentation at Scale}
\label{app:romanization_scaling}

We extend the representational overlap analysis to larger models to verify if increased parameter counts facilitate script unification.

\begin{figure}[!t]
    \centering
    \includegraphics[width=\linewidth]{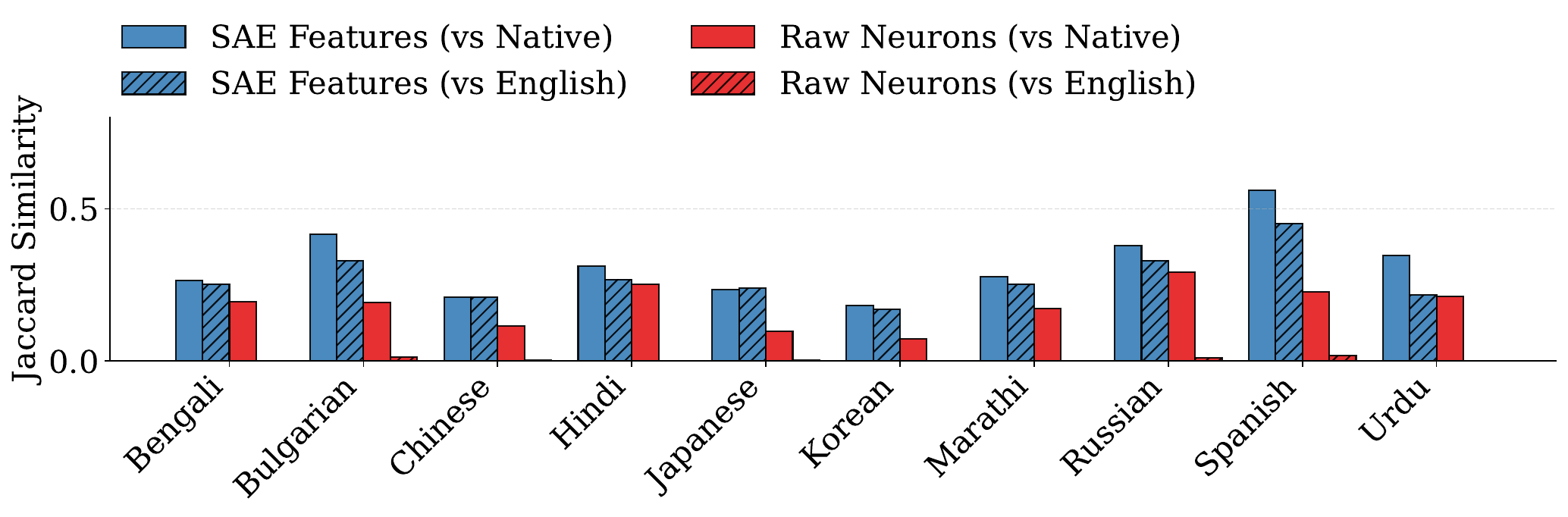}
    \caption{
    Jaccard similarity between romanized and native-script or English units in Llama-3-8B. Representational isolation persists at the 8B scale.
    }
    \label{fig:shared_isolation_8b}
\end{figure}

\begin{figure}[!t]
    \centering
    \includegraphics[width=\linewidth]{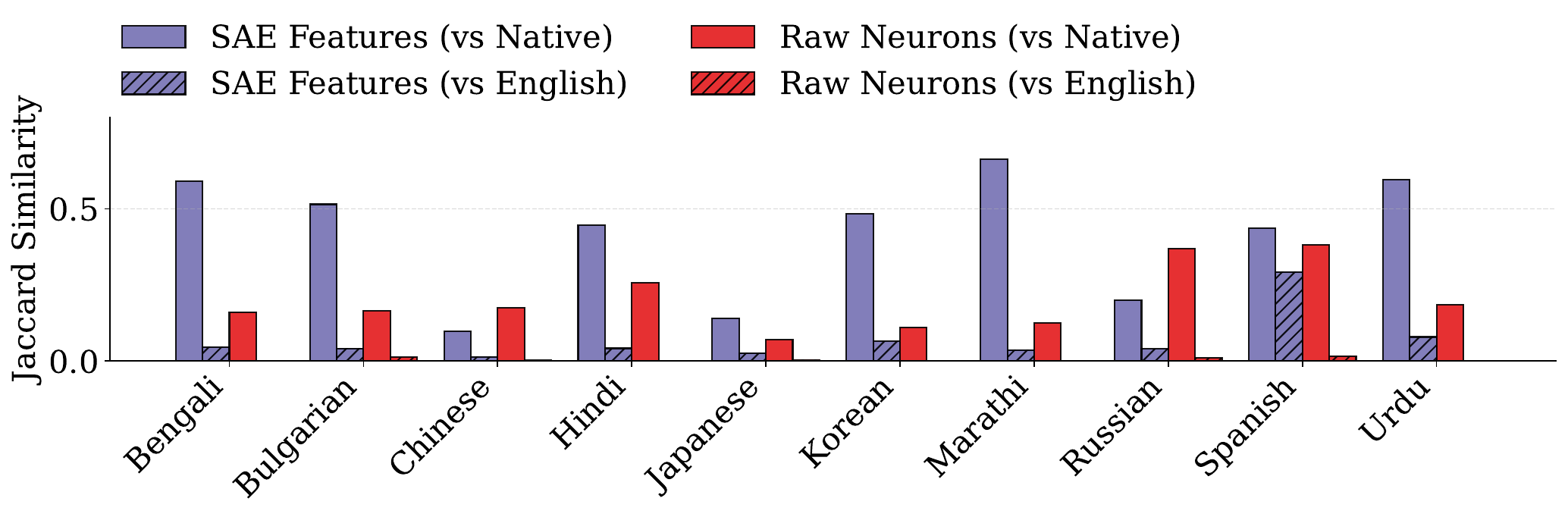}
    \caption{
    Jaccard similarity between romanized and native-script or English units in Gemma-2-9B. Fragmentation remains a dominant feature despite increased model capacity.
    }
    \label{fig:jaccard_sae_raw_native_english_9b}
\end{figure}

\paragraph{Global Overlap Trends.}
Figures~\ref{fig:shared_isolation_8b} and~\ref{fig:jaccard_sae_raw_native_english_9b} report Jaccard similarities for Llama-3-8B and Gemma-2-9B. Across both models, romanized inputs maintain near-zero overlap with English and consistently low overlap with their native-script counterparts. This indicates that even with increased capacity, models do not converge toward a unified, script-invariant representation.

\begin{figure}[!h]
    \centering
    \includegraphics[width=\columnwidth]{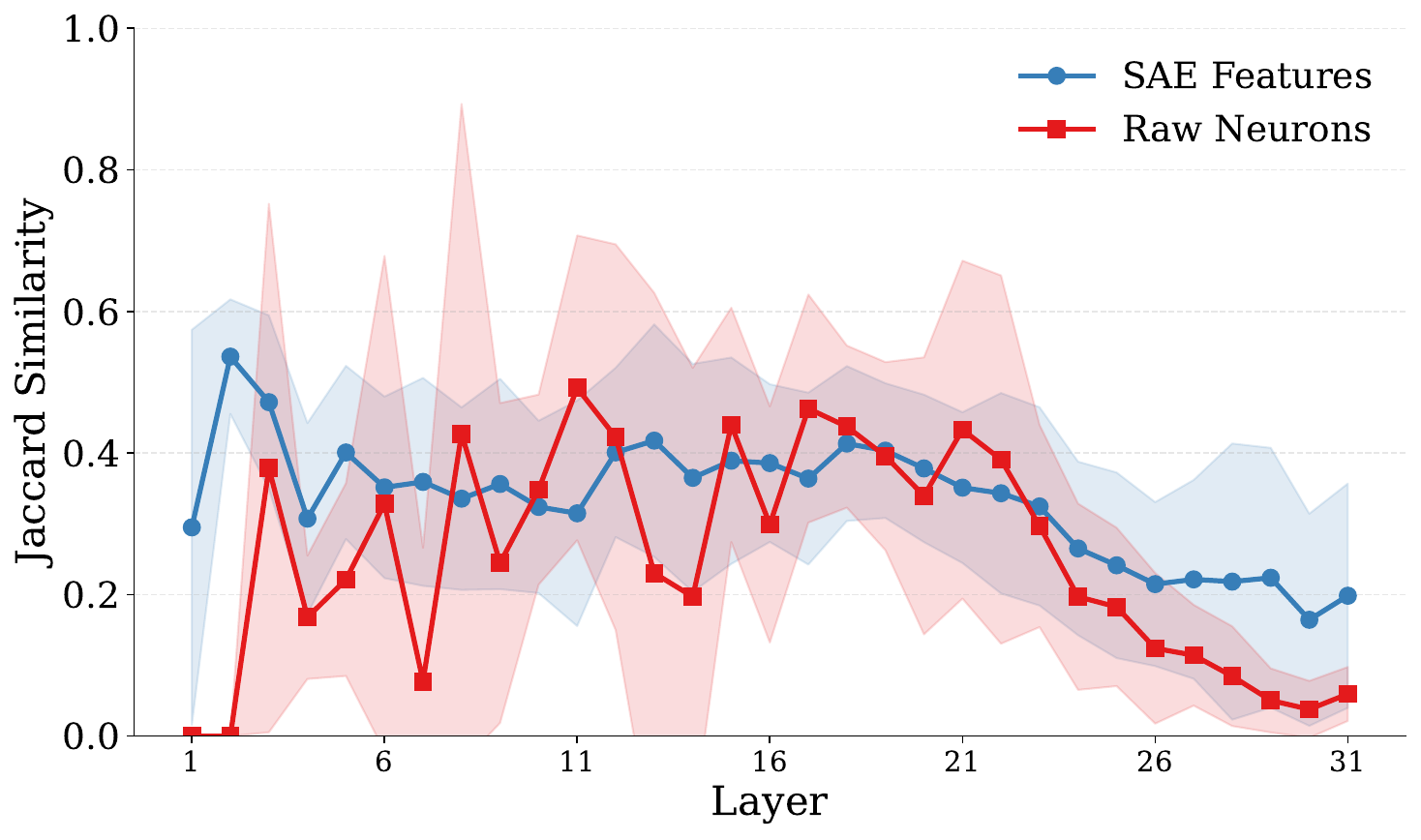}
    \caption{Layer-wise alignment in Llama-3-8B. Mid-layer increases do not lead to cross-script convergence.}
    \label{fig:layerwise_trend_8b}
\end{figure}

\begin{figure}[!h]
    \centering
    \includegraphics[width=\linewidth]{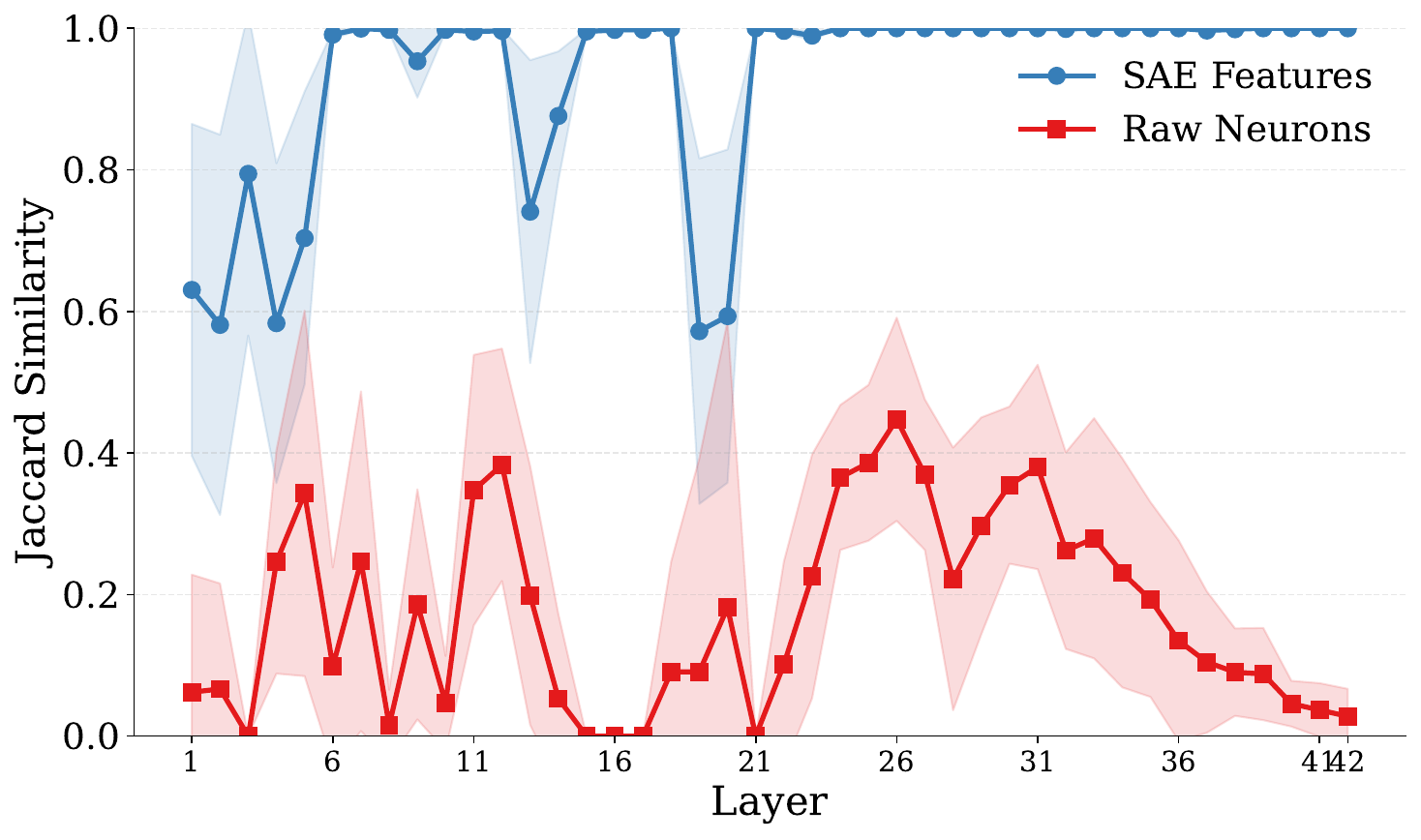}
    \caption{Layer-wise alignment in Gemma-2-9B, showing consistent representational separation in raw neurons across depth.}
    \label{fig:layerwise_trend_9b}
\end{figure}

\paragraph{Layer-wise Persistence.}
Figures~\ref{fig:layerwise_trend_8b} and~\ref{fig:layerwise_trend_9b} illustrate layer-wise alignment. While a slight increase in overlap is observed in middle layers, alignment remains far from convergence across the entire depth of the network. This confirms that representational separation is a fundamental architectural trait that persists even as models become larger and more competent.

\subsection{Structural Robustness at Scale}
\label{app:shuffling_scaling}

Finally, we examine whether larger models maintain the high robustness to structural (word-order) perturbations observed in 1B and 2B models.

\begin{figure}[!t]
    \centering
    \begin{subfigure}{\linewidth}
        \centering
        \includegraphics[width=\linewidth]{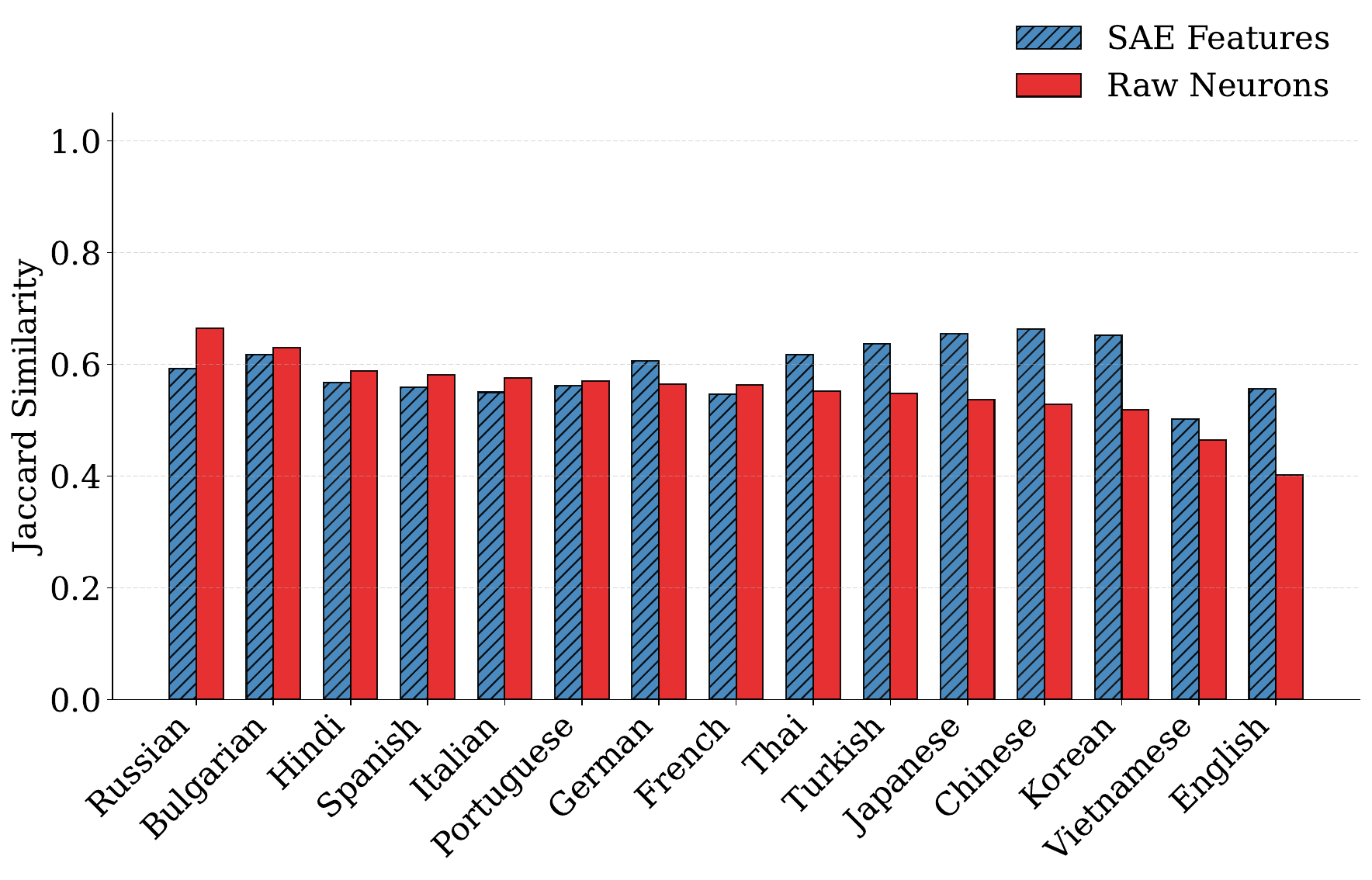}
        \caption{Llama-3-8B}
        \label{fig:shuffle_jaccard_8b}
    \end{subfigure}
    \vspace{0.5em}
    \begin{subfigure}{\linewidth}
        \centering
        \includegraphics[width=\linewidth]{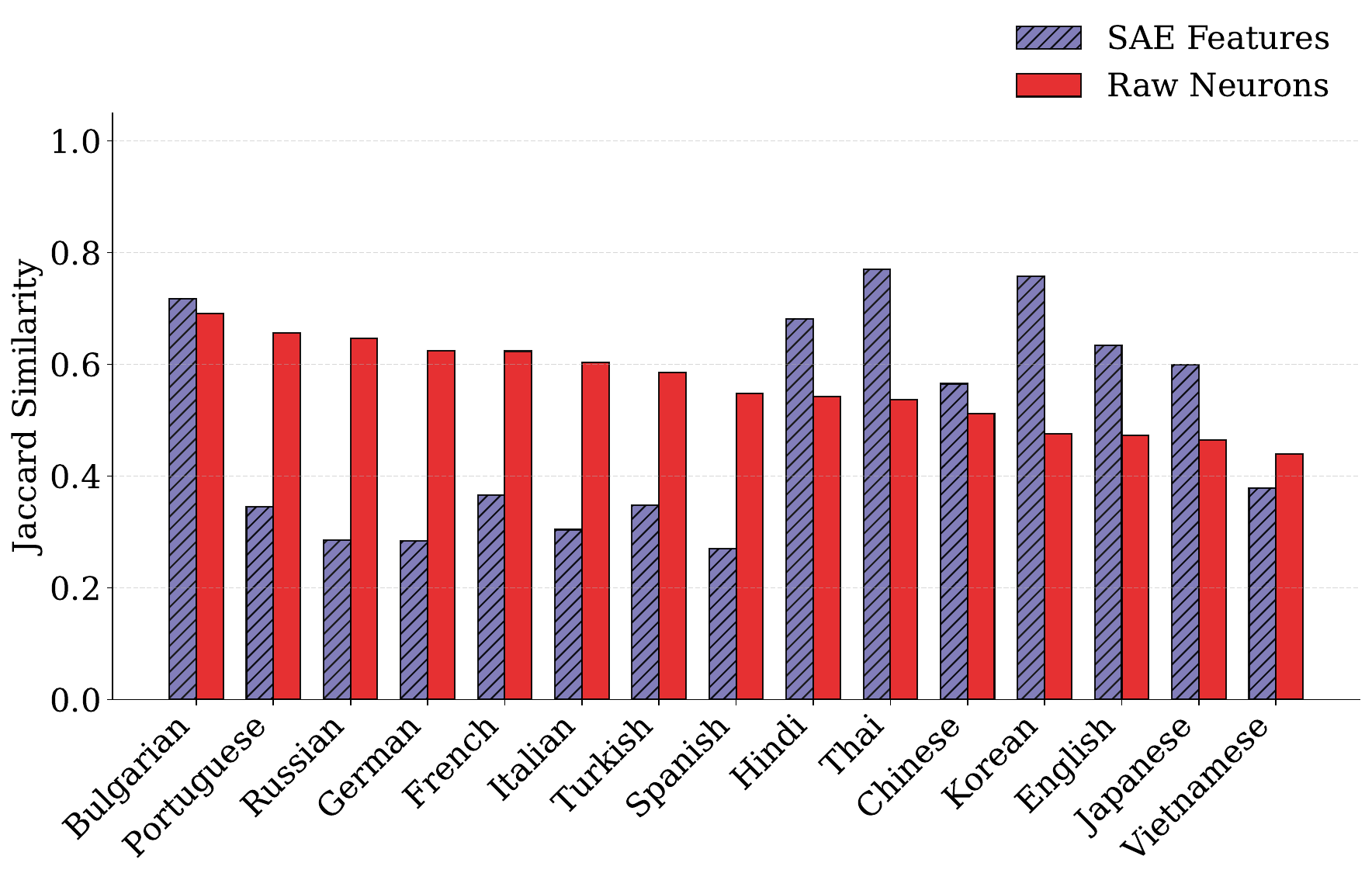}
        \caption{Gemma-2-9B}
        \label{fig:shuffle_jaccard_9b}
    \end{subfigure}
    \caption{
    Jaccard similarity between units from original and shuffled text at scale. Robustness to word-order perturbation remains consistently high across larger architectures.
    }
    \label{fig:shuffle_jaccard_scale}
\end{figure}

\paragraph{High Overlap Under Shuffling.}
As shown in Figure~\ref{fig:shuffle_jaccard_scale}, both Llama-3-8B and Gemma-2-9B exhibit consistently high Jaccard overlap between units identified from original and shuffled inputs. This confirms that the models' reliance on token-level and distributional cues (rather than strict syntactic order) is a scale-invariant property.

\end{document}